\documentclass[10pt,journal,cspaper,compsoc]{IEEEtran}
\usepackage{bm}
\usepackage{tabularx}

\usepackage{makecell}

\ifCLASSOPTIONcompsoc
  \usepackage[nocompress]{cite}
\else
\fi
\ifCLASSINFOpdf
    \usepackage[pdftex]{graphicx}
    \DeclareGraphicsExtensions{.pdf,.jpeg,.png}
\else
    \usepackage[dvips]{graphicx}
    \DeclareGraphicsExtensions{.eps}
\fi
\usepackage[cmex10]{amsmath}
\usepackage{algorithmic}
\usepackage{url}

\usepackage{multirow}
\usepackage{algorithm}

\usepackage{color}
\usepackage{amsmath}
\usepackage{amssymb}
\usepackage{mathrsfs}
\usepackage{booktabs}
\usepackage{comment}
\usepackage{multirow}
\usepackage{makecell,rotating}
\usepackage{array}
\usepackage{setspace}
\usepackage{framed}
\usepackage{bbm}
\newtheorem{proposition}{Proposition}
\newtheorem{theorem}{Theorem}
\usepackage[dvipsnames, svgnames, x11names]{xcolor}
\usepackage{extarrows}
\usepackage{soul}

\begin{document}
%
\title{Understanding and Unifying Fourteen Attribution Methods with Taylor Interactions}

%
%

\author{Huiqi Deng, Na Zou, Mengnan Du, Weifu Chen, Guocan Feng,  
Ziwei Yang, Zheyang Li, 
and Quanshi Zhang\IEEEauthorrefmark{2}
\thanks{ \IEEEauthorrefmark{2} Correspondence.}
\IEEEcompsocitemizethanks{\IEEEcompsocthanksitem Huiqi Deng, Quanshi Zhang are with the Department of Computer Science and Engineering, the John Hopcroft Center, at the Shanghai Jiao Tong University, China.}
}
\IEEEcompsoctitleabstractindextext{%

\begin{abstract}
Various attribution methods have been developed to explain deep neural networks (DNNs) by inferring the attribution/importance/contribution score of each input variable to the final output. 
However, existing attribution methods are often built upon different heuristics. 
There remains a lack of a unified theoretical understanding of why these methods are effective and how they are related. 
To this end, for the first time, we formulate core mechanisms of fourteen attribution methods, which were designed on different heuristics, into the same mathematical system, \textit{i.e.}, the system of Taylor interactions. 
Specifically, we prove that attribution scores estimated by fourteen attribution methods can all be reformulated as the weighted sum of two types of effects, \textit{i.e.}, 
independent effects of each individual input variable and interaction effects between input variables. 
The essential difference among the fourteen attribution methods mainly lies in the weights of allocating different effects.  
Based on the above findings, we propose three principles for a fair allocation of effects to evaluate the faithfulness of the fourteen attribution methods. 
\end{abstract}


}

\maketitle
\IEEEdisplaynotcompsoctitleabstractindextext
\IEEEpeerreviewmaketitle

\section{Introduction}
Despite its widespread success in  a variety of real-world applications, DNNs are typically regarded as "black boxes",  because it is difficult to interpret how a DNN makes a decision. 
The lack of interpretability hurts the reliability of DNNs, thereby hampering their wide applications on high-stake tasks, such as automatic driving \cite{dikmen2016autonomous} and AI healthcare \cite{9193963}. 
Therefore, interpreting DNNs has drawn increasing attentions recently. 

As a typical perspective of interpreting DNNs, 
attribution methods aim to calculate the attribution/im-portance/contribution score of each input variable to the network output \cite{du2019techniques, samek2020toward, lundberg2017unified}. 
For example, given a pre-trained DNN for image classification and an input image, 
the attribution score of each input variable refers to the numerical effect of each pixel on the confidence score of classification. 

Although many attribution methods have been proposed in recent years \cite{du2019techniques, samek2020toward,deng2021mutual}, 
most of them are built upon different heuristics. 
For example, some methods \cite{shrikumar2016not,sundararajan2017axiomatic} consider that the gradient of the output \textit{w.r.t.} the input can reflect the importance of input variables. In addition, some methods \cite{zeiler2014visualizing,zintgraf2017visualizing} use the output change when the input variable $x_i$ is occluded to measure the importance.

There is \textit{a lack of unified theoretical perspective} to examine the correctness of these attribution methods, or at least to mathematically clarify  their  core mechanisms, 
\textit{e.g.}, explaining their essential similarity and difference, 
and comparing their advantages and disadvantages. 

A few researchers have attempted to unify different attribution methods \cite{Marco2018Towards}\cite{samek2020toward}\cite{lundberg2017unified}, 
but these studies cover only a few methods (please see Table 1 for details). 
In this paper, we propose the Taylor interaction as a new unified perspective, which first allows us to mathematically formulate mechanisms of up to fourteen attribution methods into the same system. 
We believe that a mathematical system that unifies more methods is more likely to reflect essential factors in generating attributions,  and enable an impartial comparison between different attribution methods. 

\begin{table}[]
\centering
\begin{tabular}{c|c|c} \hline
 \textbf{Work} & \textbf{Unification} & \textbf{\# methods}  \\ \hline
\cite{lundberg2017unified} &  Additive feature attribution & 6 \\ \hline
\cite{Marco2018Towards} &  Modified gradient $\times$ input & 5 \\ \hline
\cite{samek2020toward} & First-order Taylor framework &  4 \\ \hline
Ours & Taylor interaction perspective &  14 \\ \hline
\end{tabular}
\caption{A summary of works unifying attribution methods.}
\label{tab:comparison to existing unified work}
\vspace{-3.0em}
\end{table}

\begin{figure*}[t]\centering
\includegraphics[width = 0.95 \textwidth]{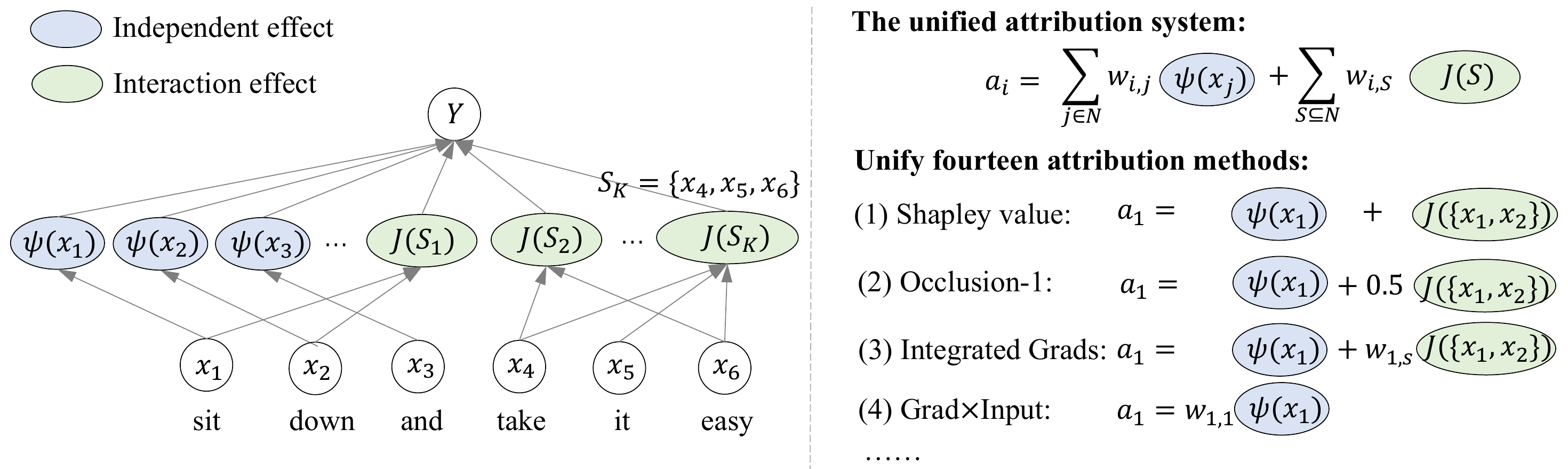}
\vspace{-1mm}
\caption{(left) We mathematically decompose the output of a DNN $Y$ as the sum of two types of effects, \textit{i.e.}, the independent effect $\psi(x_i)$ of each individual input variable $x_i$ and the interaction effect  $J(S)$ caused by collaborations between input variables  in the set $S$. 
(right) Accordingly,  we propose a unified attribution system, which formulates the attribution score $a_i$ of each specific input variable $x_i$ into a weighted sum of  independent effects  and interaction effects.  
We prove that fourteen existing attribution methods, which are designed on different heuristics, \textbf{all} can be unified into the above attribution system. 
}
\vspace{-4mm}
\label{modeling attribution problem}
\end{figure*}

The proposed Taylor interaction is a new metric to represent the two types of effects on the network output  caused by input variables. 
First, an input variable may make a direct effect on the network output, which is not influenced by other input variables.   
Such an effect is termed an \textit{independent effect}.
Second, an input variable may also collaborate with other input variables to affect the network output.
Such an effect is termed an \textit{interaction effect}. 
Both types of effects can be quantified as specific Taylor interactions. 

As a toy example, let us consider a DNN for a scene classification task that is trained to fit the target function $f_{\textrm{study room}}(\bm{x}) = 3x_{\textrm{book}} + 2 x_{\textrm{desk}}x_{\textrm{lamp}}x_{\textrm{book}}$. 
Here, the binary input variables $x_{\textrm{book}}, x_{\textrm{desk}}, x_{\textrm{lamp}} \in \{0,1\}$ denote the binary present/absent state  of these objects in the scene. 
The book variable $x_{\textrm{book}}$ has an independent effect $3x_{\textrm{book}}$ on the output.
The collaboration between variables $x_{\textrm{desk}}, x_{\textrm{lamp}}, x_{\textrm{book}}$ makes an interaction effect  on the classification of the study room scene.

In this paper, we prove that attributions scores generated by fourteen different attribution methods can \textbf{\textit{all}} be explained by the above two types of effects. 
The essential task of each attribution method can be represented as allocating a specific ratio of each independent effect and a specific ratio of each interaction effect to the input variable $x_i$, so as to compute the attribution score of $x_i$.

Furthermore, the essential difference between these attribution methods is that they compute attribution scores by allocating different ratios of independent effects and interaction effects to input variables.
For example, let us consider the previous example $f_{\textrm{study room}}(\bm{x}) = 3x_{\textrm{book}} + 2 x_{\textrm{desk}}x_{\textrm{lamp}}x_{\textrm{book}}$.  
There are an independent effect $3x_{\textrm{book}}$ and an interaction effect $2 x_{\textrm{desk}}x_{\textrm{lamp}}x_{\textrm{book}}$. 
Then, the Shapley value \cite{lundberg2017unified}  allocates the entire independent effect to the variable $x_{\textrm{book}}$, and allocate 1/3 of the interaction effect to $x_{\textrm{book}}$. 
In this way, the attribution is computed as $a_{\textrm{book}} = 3x_{\textrm{book}} + 2/3 x_{\textrm{desk}}x_{\textrm{lamp}}x_{\textrm{book}}$.
In comparison, the Occlusion-1 \cite{zeiler2014visualizing}  allocates the entire independent effect and the entire interaction effect to the variable $x_{\textrm{book}}$. 
That is, the attribution is computed as $a_{\textrm{book}} = 3x_{\textrm{book}} + 2 x_{\textrm{desk}}x_{\textrm{lamp}}x_{\textrm{book}}$. 

%
\textbf{Principles of faithful attribution.} 
The above unified perspective enables us to fairly compare different attribution methods.
To this end, we propose three principles to examine whether an attribution method faithfully allocates the two types of effects to input variables.
Let us use the attribution in the previous example of \textit{study room} scene classification to explain the three principles. 

(i) The independent effect of a variable ($3x_{\textrm{book}}$) is directly caused by the variable ($x_{\textrm{book}}$), which is not influenced by other variables. 
Therefore, the independent effect ($3x_{\textrm{book}}$) is supposed to be allocated entirely to the variable ($x_{\textrm{book}}$). 
Other variables should not be allocated such an effect. 

(ii) The interaction effect ($2 x_{\textrm{desk}}x_{\textrm{lamp}}x_{\textrm{book}}$)  is caused by the collaboration between its own set of variables ($S = \{x_{\textrm{desk}}, x_{\textrm{lamp}}, x_{\textrm{book}}\}$). 
Therefore, such an interaction effect is supposed to be allocated to the involved variables, not to variables without participating in the collaboration. 

(iii) The interaction effect ($2 x_{\textrm{desk}}x_{\textrm{lamp}}x_{\textrm{book}}$) should be \textit{all} allocated to the involved variables. 
In other words, when we sum up the numerical effects allocated from the interaction effect to the involved variables, we  obtain the exact value of the overall interaction effect ($2 x_{\textrm{desk}}x_{\textrm{lamp}}x_{\textrm{book}}$).

Subsequently, we apply the three principles to evaluate the faithfulness of the fourteen attribution methods. 
We find that attribution methods such as Shapley value  \cite{lundberg2017unified}, Integrated Gradients \cite{sundararajan2017axiomatic}, and DeepLIFT Rescale \cite{shrikumar2017learning} satisfy all principles.

In summary, this paper has three contributions:
\begin{itemize}
\item We propose the Taylor interaction as a new unified perspective to theoretically explain the core mechanism of fourteen attribution methods. 
\item For each specific attribution method, the Taylor interaction enables us to clarify its distinctive property of computing attributions.
\item We propose three principles to evaluate the faithfulness of an attribution method, which evaluate whether the attribution method faithfully allocates independent effects and interaction effects to input variables.

\end{itemize}

The preliminary version of this paper, which unifies and explains only seven attribution methods, has been published in \cite{deng2021unified}.


\section{Related work}

\subsection{Existing attribution methods}
Various attribution methods have been developed to interpret machine learning models, especially DNNs, which infer  the contribution score of each input variable to the final output. 
In general, existing attribution methods can be roughly categorized into three types, 
i) gradient-based attribution, ii) back-propagation attribution, and iii) perturbation-based attribution.

\textbf{Gradient-based attribution methods.} 
The \textit{Gradient} method \cite{baehrens2010explain} considers the gradient of the network output \textit{w.r.t.} each input variable as the attribution of the input variable.
The \textit{Gradient $\times$ Input} method \cite{shrikumar2016not} formulates attributions as the element-wise product of gradients and input features.
The \textit{Integrated Gradients} method \cite{sundararajan2017axiomatic} estimates attributions as the element-wise product of input features and the average gradient of output \textit{w.r.t.} input, where gradients are averaged when the input varies along a linear path from the input sample to a baseline point. 
The \textit{Expected Gradients} method \cite{erion2019learning} averages attribution results estimated by the \textit{Integrated Gradients} method over multiple baseline points. 
In addition, to obtain the attribution score, the \textit{Grad-CAM} method \cite{selvaraju2017grad} uses the average gradient of the loss \textit{w.r.t.} all features in a channel as the weight for the channel, and uses such a 
channel-wise weight to compute the attribution score over different locations. 

\textbf{Back-propagation attribution methods} estimate attributions of intermediate features at a layer and then back-propagate these attributions to the previous layer, to obtain the attribution scores of input variables.
This type of method includes \textit{LRP-$\epsilon$}  \cite{bach2015pixel}, \textit{LRP-$\alpha\beta$}  \cite{bach2015pixel}, \textit{Deep Taylor} \cite{montavon2017explaining}, \textit{DeepLIFT Rescale} \cite{shrikumar2017learning}, \textit{Deep SHAP} \cite{lundberg2017unified}, and \textit{DeepLIFT RevealCancel} \cite{shrikumar2017learning}.
The essential difference between different back-propagation methods is that they employ different recursive rules for back-propagating attributions between two adjacent layers, which will be detailedly introduced in Section \ref{Unified reformulation}. 

\textbf{Perturbation-based attribution methods} infer the attribution of an input variable according to the effect of masking the variable on the model output. 
The \textit{Occlusion-1} method \cite{zeiler2014visualizing} and the \textit{Occlusion-patch} method \cite{zintgraf2017visualizing} formulate the attribution of a pixel (patch) as the output change when the pixel (patch) is unmasked \textit{w.r.t.} the case when the pixel (patch) is masked. 
Moreover, the \textit{Shapley value} method \cite{lundberg2017unified} estimates the attribution by averaging such output changes when masking states of other variables vary. 
It has been proved that the Shapley value is the unique attribution method that satisfies \textit{linearity, dummy, symmetry}, and \textit{efficiency} axioms.
In addition, several methods \cite{fong2017interpretable, fong2019understanding, deng2021mutual,fu2021differentiated} identify input variables contributing the most to the network output, by seeking a minimal subset of masked variables that significantly changes the network output. 

In this paper, we explain and unify the mechanisms of as many as fourteen existing attribution methods, which cover most mainstream attribution methods. 

\subsection{Understand and  unify attribution methods}
There are a few works on theoretically understanding the mechanisms of existing heuristic attribution methods. 
For example,  the \textit{Deconvnet} method \cite{zeiler2014visualizing}  and the \textit{GBP} method \cite{springenberg2014striving} have been theoretically proved to essentially construct (partial) recovery to the input \cite{nie2018theoretical}, which is unrelated to decision-making.
Furthermore, some efforts have also been devoted to unifying various attribution methods. 
For example, \textit{LIME} \cite{ribeiro2016should}, \textit{LRP-$\epsilon$} \cite{bach2015pixel}, \textit{DeepLIFT} \cite{shrikumar2017learning}, and \textit{Shapley value}  \cite{lundberg2017unified} are unified under the framework of additive feature attribution \cite{lundberg2017unified}.
Some attribution methods including \textit{Gradient$\times$Input} \cite{shrikumar2016not}, \textit{LRP-$\epsilon$} \cite{bach2015pixel}, \textit{DeepLIFT} \cite{shrikumar2017learning} and \textit{Integrated Gradients}  \cite{sundararajan2017axiomatic}, are unified as multiplying a modified gradient with the input \cite{Marco2018Towards}. 
In addition, \cite{montavon2018methods, samek2020toward} have shown that the attributions generated by the \textit{LRP-$\epsilon$} method \cite{bach2015pixel} and the \textit{LRP-$\alpha\beta$} method \cite{bach2015pixel}  could be reformulated as a first-order Taylor decomposition. 

To the best of our knowledge, our research is the first work to leverage Taylor interaction effects to formally define the attribution problem and unify as many as fourteen existing attribution methods.

\begin{table}[]\small
\centering
\renewcommand\arraystretch{1.25}
\begin{tabular}{c|l}
Notation & Description  \\ \hline
$f$ &  pre-trained DNN \\ \hline
$\bm{x}$ & input sample $[x_1, \cdots, x_n]^T$ \\ \hline
$\bm{b}$ &  baseline point $[b_1, \cdots, b_n]^T$  \\ \hline
$\bm{a}$ & attribution vector  $[a_1, \cdots, a_n]^T$ \\ \hline
$N$ & index set of input variables $\{1, \cdots, n\}$ \\ \hline
$S$ & subset of $N$,  $S \subseteq N$ \\ \hline
$\bm{\kappa}$ & degree vector in a Taylor expansion term \\ \hline 
$I(\bm{\kappa})$ & Taylor interaction effect \\ \hline 
$\phi(\bm{\kappa})$ & Taylor independent effect   \\ \hline 
$S_{\bm{\kappa}}$ & variables involving in the interaction $I(\bm{\kappa})$ \\ \hline 
$\Omega_i$ &  set of degree vectors $\bm{\kappa}$, \textit{s.t.} $S_{\bm{\kappa}} = \{i\}$  \\ \hline 
$\Omega_S$& set of degree vectors $\bm{\kappa}$, \textit{s.t.} $S_{\bm{\kappa}} = S$  \\ \hline 
$J(S)$ & generic interaction effect of variables in $S$ \\ \hline 
$\psi(i)$ & generic independent effect of the variable $i$ \\ \hline 
\end{tabular}
\caption{Notation in this paper.}
\label{tab:my_label}
\end{table}

\begin{figure*}[t]\centering
\includegraphics[width=0.99\textwidth]{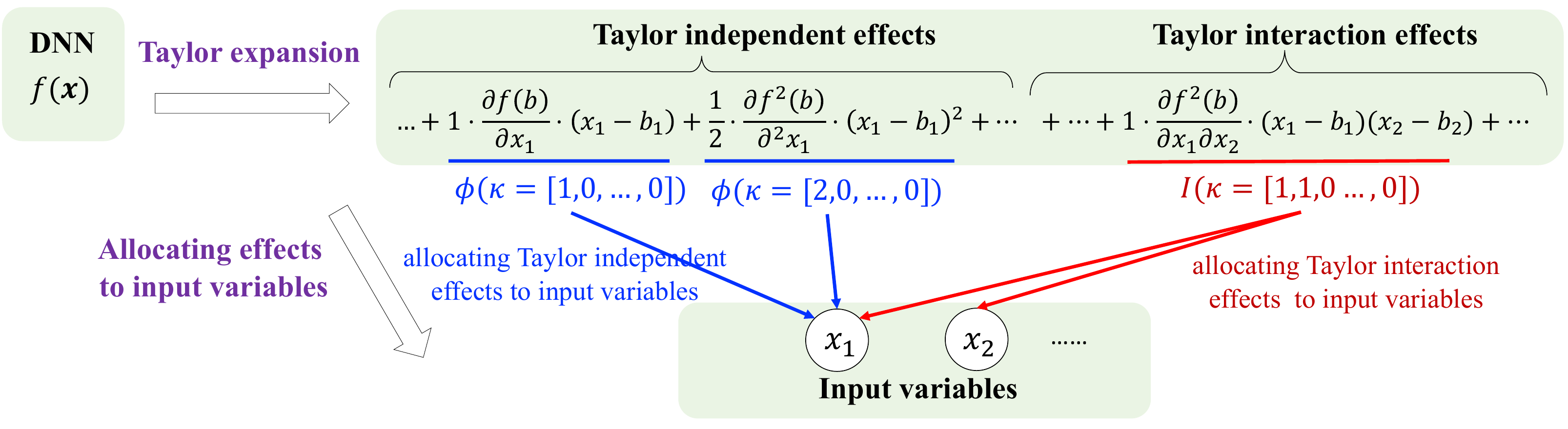}
\caption{Understanding attribution methods via Taylor interactions. 
We have proven that each attribution method is mathematically equivalent to the following flowchart, \textit{i.e.}, each attribution method first explains the network output $f(\bm{x})$ as a Taylor expansion model, 
thereby decomposing the network output $f(\bm{x})$ into the Taylor independent effect $\phi(\bm{\kappa})$ of each input variable $x_i$ and the Taylor interaction effect $I(\bm{\kappa})$ between each set of input variables (introduced in Section \ref{sec: Taylor interaction}). 
Then, this method accordingly re-allocates the two types of effects to each input variable $x_i$, so as to compute the attribution score $a_i$ (introduced in Section \ref{Unified reformulation}). 
Here, we show Taylor expansion terms of only the first and second orders for simplicity.
}
\label{Taylor attribution diagram}
\end{figure*}

\section{Unifying attribution methods}
\label{Taylor Attribution Framework}
Attribution methods have been developed as a typical perspective of explaining DNNs \cite{du2019techniques} \cite{lundberg2017unified}, which infer the attribution/importance/contribution score of each input variable (\textit{e.g.}, an image pixel, a word) to the final output. 
Specifically, given a pre-trained DNN $f$ and an input sample $\bm{x}  = [x_1,\dots x_n]^T \in \mathbb{R}^n$, an attribution method estimates an attribution vector $\bm{a} = [a_1,\dots, a_n ]^T
\in \mathbb{R}^n$, 
where $a_i$ denotes the numerical effect of the input variable $i$ on a scalar output of the DNN $f(\bm{x}) \in \mathbb{R}$.
For example, in the classification task, $f(\bm{x})$ can be set as the classification probability of the target category.

Although various attribution methods have been proposed recently, most of them are built upon different heuristics.
There still lacks a unified perspective to explain why these attribution methods are effective and how they are related.
Therefore, in this paper, we propose the Taylor interaction as a new unified perspective, which allows us to explain the mechanisms of up to fourteen attribution methods.

\subsection{Explaining a DNN by Taylor interaction effects}
\label{sec: Taylor interaction}
In this subsection, we propose the Taylor interaction as a new perspective, which mathematically proves that the output of a DNN can be decomposed into two typical types of effects,  including the Taylor independent effect of each input variable and the Taylor interaction effect between input variables.
In the following subsections, we will use the two effects to explain and compare the core mechanisms of different attribution methods. 

\textbf{Preliminaries: Taylor expansion of a DNN.} 
Given a pre-trained DNN $f$ and an input sample $\bm{x} =  [x_1, \dots, x_n]^T$ with $n$ input variables (indexed by $N= \{1, \cdots, n\}$),  let us consider the $K$-order Taylor expansion\footnote{Note that although deep networks with ReLU activation are not differentiable such that the Taylor expansion is not applicable, we can use networks with softplus activation (the approximation of ReLU) to provide insight into the rationale behind ReLU networks.} of the DNN $f$, which is expanded at a baseline point $\bm{b}  = [b_1, \dots, b_n]^T$. 
\begin{equation} \label{eqn:preliminary}
\begin{small}
\begin{aligned}
\!\! f(\bm{x})   &  =     f(\bm{b})   + \sum_{i = 1}^n  \frac{1}{1!} \cdot \frac{\partial f(\bm{b})}{\partial x_i} \cdot  (x_i - b_i) \\
  +  & \sum_{i=1}^n \sum_{j = 1}^n \frac{1}{2!} \cdot  \frac{\partial^2 f(\bm{b})}{\partial x_i\partial x_j} \cdot (x_i - b_i)(x_j - b_j) + \cdots + \epsilon_K \\
&=   f(\bm{b}) + \sum_{k=1}^{K}   \sum_{\substack{ \bm{\kappa} \in O_k}}   \underbrace{ C(\bm{\kappa}) \cdot \bigtriangledown f(\bm{\kappa}) \cdot  \pi(\bm{\kappa})}_{\text{defined as } I(\bm{\kappa})} + \ \epsilon_K 
 \end{aligned}
\end{small}
\end{equation}
where $\epsilon_K$ denotes the approximation error of the $K$-order expansion. 
Each expansion term {\small $I(\bm{\kappa})$} is defined as follows, which consists of  the coefficient {\small $C(\bm{\kappa})$}, the partial derivative {\small $\bigtriangledown f(\bm{\kappa})$}, and the product {\small $\pi(\bm{\kappa})$}. 
\begin{equation}\label{eqn:definition}
\begin{small}
\begin{aligned}
 I(\bm{\kappa})  &\overset{\text{def}}{=} C(\bm{\kappa}) \cdot \bigtriangledown f(\bm{\kappa}) \cdot \pi(\bm{\kappa}) \\[3pt]
s.t. \quad C(\bm{\kappa}) & = \frac{1}{(\kappa_1+\cdots+\kappa_n)!} \tbinom{\kappa_1+\cdots+\kappa_n}{\kappa_1, \cdots, \kappa_n}  \\[1mm]
 \bigtriangledown f(\bm{\kappa})   &= \frac{\partial^{\kappa_1+\cdots+\kappa_n} f(\bm{b})}{\partial^{\kappa_1}  x_1\cdots \partial^{\kappa_n} x_n } \\[2mm]
 \pi(\bm{\kappa}) & =  (x_1 - b_1)^{\kappa_1} \dots (x_n - b_n)^{\kappa_n} \\
  \end{aligned}
\end{small}
\end{equation}
Here, {\small $\bm{\kappa} =  [\kappa_1, \cdots,  \kappa_n] \in \mathbb{N}^n$} denotes the degree vector of the expansion term {\small $I(\bm{\kappa})$},  and {\small $\kappa_i \in \mathbb{N}$} denotes the non-negative integral degree of the variable $i$.

Moreover, we classify all expansion terms in Eq.~(\ref{eqn:preliminary}) into different orders. 
The order of each expansion term {\small $I(\bm{\kappa})$} is defined as its overall degree, 
\textit{i.e.}, {\small $\text{order}(I(\bm{\kappa}))=\kappa_1+\cdots+\kappa_n$}.
In this way, we can use the set of degree vectors {\small $O_k = \{\bm{\kappa} \in \mathbb{N}^n|  \kappa_1+\cdots+\kappa_n = k \}$} to represent all expansion terms of the $k$-th order. 
\\

\textbf{Taylor interaction effects.}
In Eq.~(\ref{eqn:preliminary}), each Taylor expansion term {\small $I(\bm{\kappa})$} represents an interaction between input variables in the set {\small $S_{\bm{\kappa}}$}. 
Here, {\small $S_{\bm{\kappa}}$} denotes the receptive  field of the interaction {\small $I(\bm{\kappa})$}, \textit{i.e.}, the set of all variables involved in the interaction. 
\begin{equation}
\begin{small}
\begin{aligned}
S_{\bm{\kappa}} \overset{\text{def}}{=}  \{i| \kappa_i >0 \} 
  \end{aligned}
\end{small}
\end{equation}
Let us take the Taylor expansion term {\small $I(\bm{\kappa}) =  c \cdot (x_{\rm eye}-b_{\rm eye})^2(x_{\rm nose} - b_{\rm nose})(x_{\rm mouth} - b_{\rm mouth})$} of the DNN for face recognition as an example.
This expansion term {\small $I(\bm{\kappa})$} indicates that  variables in {\small $S_{\bm{\kappa}} = \{\text{eye}, \text{nose}, \text{mouth}\}$} interact with each other to form an AND pattern. 
Only when all variables in {\small $S_{\bm{\kappa}}$} co-appear, the AND pattern is formed and makes an interaction effect {\small $I(\bm{\kappa})$} on the output {\small $f(\bm{x})$} of the DNN. 
Instead, masking any of variables of $x_{\rm eye}, x_{\rm nose},$ and $x_{\rm mouth}$ using their baseline value $b_i$ will deactivate the AND pattern and remove the numerical effect from the network output, \textit{i.e.}, making {\small $I(\bm{\kappa}) = 0$}. 
Therefore, {\small $I(\bm{\kappa})$} quantifies the effect of the interaction (AND pattern) on the network output,  which is termed the \textbf{\textit{Taylor interaction effect}}.

\textbf{Taylor independent effects.} 
We further define a specific type of Taylor interaction effect {\small $I(\bm{\kappa})$}, where only a single variable is involved in the interaction ({\small $|S_{\bm{\kappa}}|=1$}), as the \textbf{\textit{Taylor independent effect}}. 
We denote the Taylor independent effect by a new notation {\small $\phi(\bm{\kappa})$} to differentiate it from other Taylor interaction effects. 
\begin{equation}\label{eqn:def_independent}
\begin{small}
\begin{aligned}
\phi(\bm{\kappa}) \overset{\text{def}}{=} I(\bm{\kappa}), \quad \forall \ \bm{\kappa} \in \{\bm{\kappa} \ | \ |S_{\bm{\kappa}}| = 1\}.
\end{aligned}
\end{small}
\end{equation}
The Taylor independent effect represents the effect of a single variable without depending on (interacting with) other variables. 
For example, when the degree vector {\small $\bm{\kappa} = [0, \cdots, 0, \kappa_i, 0, \cdots, 0]$} satisfying {\small $S_{\bm{\kappa}} = \{i\}$}, the Taylor independent effect is computed as
\begin{equation}\label{eqn:independent}
\begin{small}
\begin{aligned}
\phi(\bm{\kappa}) = \frac{1}{\kappa_i !}\frac{\partial^{\kappa_i} f(\bm{b})}{\partial^{\kappa_i} x_i}(x_i - b_i)^{\kappa_i},
\end{aligned}
\end{small}
\end{equation}
which is influenced only by the single variable $i$. 

To avoid ambiguity, in the following manuscript, we use the Taylor independent effect {\small $\phi(\bm{\kappa})$} to represent the effect of a single variable without depending on (interacting with) other variables 
({\small $|S_{\bm{\kappa}}|=1$}), and use the Taylor interaction effect {\small $I(\bm{\kappa})$} to represent the interaction effect of multiple variables ({\small $|S_{\bm{\kappa}}|>1$}).  
\\

\textbf{Decomposing the network output into the generic independent effect of each input variable and the generic interaction effect of each set of input variables.} 
For a specific set of input variables {\small $S  (S \subseteq N, |S| > 1)$}, let {\small $J(S)$} denote the overall effect caused by interactions between variables in {\small $S$},
which sums up all Taylor interaction effects \textit{w.r.t.} the receptive field {\small $S$}. 
We term {\small $J(S)$} the \textbf{\textit{generic interaction effect}} for {\small $S$}.
\begin{equation}\label{eqn:genericeffects1}
\begin{small}
\begin{aligned}
 J(S)  \overset{\text{def}}{=} & \sum\nolimits_{\bm{\kappa} \in \Omega_S} I(\bm{\kappa}), \ s.t. \ \Omega_S = \{\bm{\kappa}|S_{\bm{\kappa}} = S\},  \\
\end{aligned}
\end{small}
\end{equation}
where {\small $\Omega_S$} is a set of degree vectors $\bm{\kappa}$ corresponding to all Taylor interaction effects {\small $I(\bm{\kappa})$} with the receptive field {\small $S$}. 
In the following Section \ref{sec:Taylor interactions vs Harsanyi}, we prove that the generic interaction effect {\small $J(S)$} just represents  the Harsanyi dividend \cite{harsanyi1963simplified}, which is a typical game-theoretic interaction metric.

Similarly, we define the \textbf{\textit{generic independent effect}} {\small $\psi(i)$} of the variable $i$ as follows,
to measure the overall effect of the variable $i$ without depending on (interacting with) other variables. 
\begin{equation}\label{eqn:genericeffects2}
\begin{small}
\begin{aligned}
\psi(i)  \overset{\text{def}}{=}&  \sum\nolimits_{\bm{\kappa}\in \Omega_i}\phi(\bm{\kappa}), 
\ s.t. \ \Omega_i = \{\bm{\kappa}|S_{\bm{\kappa}} = \{i\}\}. \\
\end{aligned}
\end{small}
\end{equation}
where {\small $\Omega_i$} is a set of degree vectors $\bm{\kappa}$ corresponding to all Taylor independent effects of the variable $i$. 
\vspace{2pt}

\begin{proposition}\label{proposition:output decomposition} 
\textit{(Proof in Appendix A) The network output {\small $f(\bm{x})$} can be decomposed as the sum of generic independent effects {\small $\psi(i)$} of different input variables $i$ and generic  interaction effects {\small $J(S)$} w.r.t. different subsets {\small $S$} of input variables.}
\begin{equation}\label{eqn:decomposition}
\begin{small}
\begin{aligned}
 f(\bm{x})   &= f(\bm{b})  + \sum\limits_{i \in N} \sum_{\bm{\kappa} \in \Omega_i} \phi(\bm{\kappa})  + \sum\limits_{\substack{S \subseteq N, |S| > 1}} \sum_{\bm{\kappa} \in \Omega_S} I(\bm{\kappa}) \\
 &=  f(\bm{b})  + \sum_{i \in N}  \psi(i)  + \sum_{\substack{S \subseteq N,|S| > 1}} J(S) 
\\
\end{aligned}
\end{small}
\end{equation}
\end{proposition}

\subsection{Connections between the Taylor interaction effect and the Harsanyi dividend}
\label{sec:Taylor interactions vs Harsanyi}
In this subsection, we prove theoretical connections between the Taylor interaction effect (generic interaction effect) and the Harsanyi dividend \cite{harsanyi1963simplified}.

The Harsanyi dividend {\small $H(S)$} is a typical game-theoretic interaction metric to measure the interaction effect between a specific set {\small$S$} of input variables, which is computed as follows. 
\begin{equation}
\begin{small}
\begin{aligned}
H(S) =  \sum\nolimits_{T \subseteq S} (-1)^{|T|-|S|} f(\bm{x}_T), \ \ \forall S \subseteq N, |S| > 1
\end{aligned}
\end{small}
\end{equation}
where $f(\bm{x}_T)$ denotes the network output when variables in $T$ of the input sample $\bm{x}$ remain unchanged, and variables in $N \setminus T$ are masked using baseline values, \textit{i.e.}, $\forall i \in N \setminus T$, setting $x_i = b_i$. 
Furthermore, the Harsanyi dividend is considered a general metric.
 This is because \cite{ren2021towards} has proven that the Harsanyi dividend satisfies seven desirable axioms, and can be considered an elementary interaction component of many existing game-theoretic metrics, such as the Shapley interaction index  \cite{grabisch1999axiomatic} and Shapley Taylor interaction index \cite{sundararajan2020shapley}. 
\vspace{2pt}

\begin{theorem}\label{thm:connection}
\textit{(Proof in Appendix A) The Harsanyi dividend {\small $H(S)$} is equivalent to the generic interaction effect {\small $J(S)$} between variables in {\small $S$}, which is defined in Eq.~(\ref{eqn:genericeffects1}).}
\begin{equation}
\begin{small}
\begin{aligned}
H(S) = J(S) =  \sum\nolimits_{\bm{\kappa} \in \Omega_S} I(\bm{\kappa}), \ \ \forall S \subseteq N, |S| > 1
\end{aligned}
\end{small}
\end{equation}
\end{theorem}

Theorem \ref{thm:connection} proves the equivalence between the typical Harsanyi dividend interaction metric and the generic interaction effect, which guarantees the trustworthiness of using the generic interaction effect and the Taylor interaction effect to measure the interaction effect between input variables.


\subsection{Rewriting attributions as the weighted sum of  independent effects and interaction effects}
\label{Unified reformulation}
In this subsection, we revisit the attribution problem from the interaction perspective.
We discover that all attributions generated by different attribution methods can be represented as a weighted sum of independent effects and interaction effects.

According to Eq. (\ref{eqn:decomposition}), the input variable $i$ usually has a generic independent effect $\psi(i)$ and different generic interaction effects $J(S)$ ($i \in S$) to affect the network output. 
Naturally, both types of effects are supposed to contribute attribution scores to the variable $i$.

In this paper, we \textbf{\textit{prove}} that although fourteen existing attribution methods are designed on different heuristics, the attribution score estimated by each method can all be represented as a specific re-allocation of generic independent effects and generic interaction effects.
Specifically, let $a_i$ denote the attribution score of the variable $i$.
We prove that $a_i$ estimated by fourteen attribution methods can all be reformulated into the following paradigm. 
\begin{equation}\label{eqn:unifiedattribution1}
\begin{small}
\begin{aligned}
a_i &  = \sum_{ j \in N} \underbrace{w_{i,j}\psi(j)}_{\overset{\text{def}}{=} a_{i \leftarrow \psi(j)} } +
\sum_{\substack{S \subseteq N,|S|>1}}  \underbrace{w_{i, S} J(S)}_{\overset{\text{def}}{=}  a_{i \leftarrow J(S)}}
\end{aligned}
\end{small}
\end{equation} 
 where $w_{i,j}$ denotes the ratio of $j$'s generic independent effect {\small $\psi(j)$} being allocated to the variable $i$, 
and $w_{i,S}$ denotes the ratio of the generic interaction effect {\small $J(S)$} between variables in $S$ that is allocated to the variable $i$. 
Accordingly, we can use {\small $a_{i \leftarrow \psi(j)} \overset{\text{def}}{=} w_{i,j}\psi(j)$} and {\small $a_{i \leftarrow J(S)} \overset{\text{def}}{=} w_{i,S}J(S)$} to represent the allocated effects from the generic independent effect {\small $\psi(j)$} and the generic interaction effect {\small $J(S)$}, respectively.


To be precise, we can further expand the above equation as a re-allocation of Taylor independent effects and Taylor interaction effects.
\begin{equation}\label{eqn:unifiedattribution2}
\begin{small}
\begin{aligned}
a_i & = \sum_{ j \in N}\underbrace{\sum_{\bm{\kappa} \in \Omega_j} w_{i,\bm{\kappa}}\phi(\bm{\kappa})}_{= a_{i \leftarrow \psi(j)}} +
\sum_{\substack{S \subseteq N}} \underbrace{\sum_{\bm{\kappa} \in \Omega_S}  w_{i, \bm{\kappa}} I(\bm{\kappa})}_{= a_{i \leftarrow J(S)}}
\end{aligned}
\end{small}
\end{equation} 
where $w_{i,\bm{\kappa}}$ denotes the ratio of a specific Taylor independent effect $\phi(\bm{\kappa})$ (Taylor interaction effect $I(\bm{\kappa})$) that is allocated to $a_i$. 
By combining Eq.~(\ref{eqn:unifiedattribution1}) and Eq.~(\ref{eqn:unifiedattribution2}),  we can obtain that 
{\small $w_{i,j} = \frac{\sum_{\bm{\kappa} \in \Omega_j} w_{i,\bm{\kappa}}\phi(\bm{\kappa})}{\psi(j)}$} and 
{\small $w_{i,S} = \frac{\sum_{\bm{\kappa} \in \Omega_S}  w_{i, \bm{\kappa}} I(\bm{\kappa})}{J(S)}$}.

\textbf{Essential difference between attribution methods.}
Based on the unified paradigm in Eq.~(\ref{eqn:unifiedattribution1}) and Eq.~(\ref{eqn:unifiedattribution2}), we can consider that the essential difference between different attribution methods is that each attribution method actually uses a different ratio $w_{i,j}$, $w_{i,S}$, and $w_{i,\bm{\kappa}}$ to re-allocate different effects, to compute the attribution score $a_i$. 

Furthermore, although different attribution methods can be written as the above paradigm of allocating independent effects and interaction effects, not all attribution methods allocate a reasonable ratio of each effect to the attribution score $a_i$. 
For example, we find that some attribution methods may allocate part of the generic interaction effect {\small $J(S)$} to the variable $i$ that is not involved in the interaction (\textit{i.e.}, $i \not\in S$). 
In addition, some attribution methods may fail to completely allocate all numerical values of the generic interaction effect {\small $J(S)$} to input variables, \textit{e.g.}, {\small $\sum_{i} a_{i \leftarrow J(S)} < J(S)$}. 
Therefore, in section \ref{sec:principles}, we propose three principles to examine whether an attribution method reasonably allocates independent effects and interaction effects, to evaluate the faithfulness of  attribution methods.

\subsection{Unifying fourteen attribution methods with interaction effects and independent effects}
\label{Prove unified reformulation}
In this subsection, we reformulate fourteen existing attribution methods into the unified paradigm
of allocating Taylor independent effects and Taylor interaction effects in Eq. (\ref{eqn:unifiedattribution2}) one by one.
\\

\noindent\textbf{Gradient$\times$Input.} 
Gradient$\times$Input  \cite{shrikumar2016not} estimates the attribution by roughly considering the complex DNN $f$ as a linear model, \textit{i.e.}, {\small $f(\bm{x}) \xlongequal{\text{explained}} f(\bm{0}) + \sum_i \frac{\partial f(\bm{x})}{\partial x_i}x_i$}.
Here, {\small $\frac{\partial f(\bm{x})}{\partial x_i}$} denotes the gradient of the output \textit{w.r.t.} the input variable $i$.
Therefore, Gradient$\times$Input considers that the product of the gradient and input reflects the attribution of the variable $i$. 
\begin{equation}
\begin{small}
\begin{aligned}
a_i = \frac{\partial f(\bm{x})}{\partial x_i}x_i.
\end{aligned}
\end{small}
\end{equation}

\begin{theorem}\label{thm:gradinput}
\textit{(Proof in Appendix B) In the Gradient $\times$Input method, the attribution of the input variable $i$ can be reformulated as}
\begin{equation}
\begin{small}
\begin{aligned}
a_i &= \phi(\bm{\kappa}) = \frac{\partial f(\bm{x})}{\partial x_i}x_i.
\end{aligned}
\end{small}
\end{equation}
\textit{where $\bm{\kappa} = [\kappa_1, \cdots, \kappa_n]$ is a one-hot degree vector with $\kappa_i = 1$ and  $\forall j \neq i, \kappa_j = 0$.} 
\end{theorem}
\vspace{2pt}

Theorem \ref{thm:gradinput} shows that the Gradient$\times$Input method follows the paradigm of allocating Taylor interaction effects in Eq.~(\ref{eqn:unifiedattribution2}).
Specifically, this method allocates only a specific Taylor independent effect  $\phi(\bm{\kappa})$ of the variable $i$ to the attribution of $i$.
\\

\begin{table*}[!hbtp]
\renewcommand\arraystretch{1.5}
\centering
\begin{tabular}{l|l}
\toprule
\textbf{Attribution methods} & \textbf{Unified paradigm of allocating Taylor interaction effects}   \\  \hline
Gradient$\times$Input \cite{shrikumar2016not}& $a_i = \phi(\bm{\kappa}),  \ \bm{\kappa} = [0,\cdots, \kappa_i = 1, \cdots, 0]$  \\ \hline
Occlusion-1  \cite{zeiler2014visualizing}  & $a_i  = \sum_{\bm{\kappa} \in \Omega_i} \phi(\bm{\kappa}) + \sum_{|S|>1, i \in S}\sum_{\bm{\kappa} \in \Omega_S} I(\bm{\kappa})$  \\ \hline
Occlusion-patch  \cite{zeiler2014visualizing}  & $a_i  = \sum\nolimits_{k \in S_j} \sum\nolimits_{\bm{\kappa} \in \Omega_k}\phi(\bm{\kappa}) + \sum\nolimits_{|S|>1, S \cap S_j \neq \emptyset} \sum\nolimits_{\bm{\kappa} \in \Omega_S} I(\bm{\kappa})$ \\ \hline
Prediction Difference \cite{zintgraf2017visualizing} &     $a_i  = \mathbb{E}_{\bm{b} \sim p(\bm{b})}  [ \sum_{\bm{\kappa} \in \Omega_i} \phi(\bm{\kappa}|\bm{b}) +  \sum_{i \in S} \sum_{\bm{\kappa} \in \Omega_S} I(\bm{\kappa}|\bm{b})]$ \\ \hline
Grad-CAM \cite{selvaraju2017grad} & $\tilde a_i = \phi(\bm{\kappa}), \ \bm{\kappa} = [0,\cdots, \kappa_i = 1, \cdots, 0]$  \\ \hline
Integrated Gradients \cite{sundararajan2017axiomatic} &  $a_i  =  \sum_{\bm{\kappa} \in \Omega_i} \phi(\bm{\kappa}) + \sum_{|S|>1, i \in S}\sum_{\bm{\kappa} \in \Omega_S} \frac{\kappa_i}{\sum_{i} \kappa_{i}} \cdot I(\bm{\kappa})$ \\ \hline
Expected Gradients \cite{erion2021improving} & $a_i  = \mathbb{E}_{\bm{b} \sim p(\bm{b})}  [\sum_{\bm{\kappa}\in \Omega_i} \phi(\bm{\kappa}|\bm{b}) + \sum_{\substack{ i \in S}} \sum_{\bm{\kappa} \in \Omega_S} \frac{\kappa_i}{\sum_{i} \kappa_{i}} I(\bm{\kappa}|\bm{b})]$ \\ \hline
Shapley value \cite{lundberg2017unified} & $a_i  =  \sum_{\bm{\kappa} \in \Omega_i} \phi(\bm{\kappa}) +  \sum_{|S|>1, i \in S}\sum_{\bm{\kappa} \in \Omega_S}  \frac{1}{|S|} \cdot I(\bm{\kappa})$ \\ \hline
LRP-$\epsilon$  \cite{bach2015pixel} &  $a_i = \phi(\bm{\kappa}), \ \bm{\kappa} = [0,\cdots, \kappa_i = 1, \cdots, 0]$  \\ \hline
LRP-$\alpha\beta$  \cite{bach2015pixel} &  \makecell[l]{$a_i = \alpha [\sum_{\bm{\kappa}\in \Omega_i} \phi(\bm{\kappa}) + \sum_{i \in S} \sum_{\bm{\kappa} \in \Omega_S}  c_i I(\bm{\kappa})  + \sum_{S \subseteq N^-} \sum_{\bm{\kappa} \in \Omega_S}  d_i I(\bm{\kappa})], \forall i \in N^+ $ \\[1mm]
$a_i = \beta   [\sum_{\bm{\kappa}\in \Omega_i} \phi(\bm{\kappa}) +  \sum_{ i \in S} \sum_{\bm{\kappa} \in \Omega_S}  \tilde c_i I(\bm{\kappa}) +   \sum_{S \subseteq N^+}\sum_{\bm{\kappa} \in \Omega_S} \tilde d_i I(\bm{\kappa})], \forall i \in N^-$}
\\  \hline
Deep Taylor \cite{montavon2017explaining} &   \makecell[l]{$a_i = \sum_{\bm{\kappa}\in \Omega_i} \phi(\bm{\kappa}) + \sum_{i \in S} \sum_{\bm{\kappa} \in \Omega_S} c_i  I(\bm{\kappa})  + \sum_{S \subseteq N^-} \sum_{\bm{\kappa} \in \Omega_S} d_i I(\bm{\kappa}), \forall i \in N^+ $
\\[1mm]
$a_i = 0, \forall i \in N^-$}
\\ \hline
DeepLIFT Rescale \cite{shrikumar2017learning} &  $a_i  =  \sum_{\bm{\kappa} \in \Omega_i} \phi(\bm{\kappa}) + \sum_{|S|>1, i \in S}\sum_{\bm{\kappa} \in \Omega_S} \frac{\kappa_i}{\sum_{i} \kappa_{i}} \cdot I(\bm{\kappa})$ \\ \hline
 DeepShap \cite{lundberg2017unified} & $a_i  =  \sum_{\bm{\kappa} \in \Omega_i} \phi(\bm{\kappa}) +  \sum_{|S|>1, i \in S}\sum_{\bm{\kappa} \in \Omega_S}  \frac{1}{|S|} \cdot I(\bm{\kappa})$  \\ \hline
DeepLIFT Reveal \cite{shrikumar2017learning} & \makecell[l]{$a_i = \sum_{\bm{\kappa}\in \Omega_i} \phi(\bm{\kappa}) +  \sum_{\substack{S \subseteq N^+, i \in S}} \ \sum_{\bm{\kappa} \in \Omega_S}  c_i I(\bm{\kappa})+  \sum\nolimits_{S \cap N^+ \neq \emptyset, S \cap N^- \neq \emptyset, i \in S}$ 
\\ \quad \quad  $\sum_{\bm{\kappa} \in \Omega_S}  \frac{1}{2}c_i I(\bm{\kappa})], \ \forall i \in N^+$
\\[1mm]
$a_i = \sum_{\bm{\kappa}\in \Omega_i} \phi(\bm{\kappa}) +  \sum_{\substack{S \subseteq N^-, i \in S}} \ \sum_{\bm{\kappa} \in \Omega_S}  \tilde c_i I(\bm{\kappa}) + \sum\nolimits_{S \cap N^+ \neq \emptyset, S \cap N^- \neq \emptyset, i \in S}$
\\
\quad \quad $\sum_{\bm{\kappa} \in \Omega_S} \frac{1}{2} \tilde c_i I(\bm{\kappa})], \forall i \in N^-$}
 \\ \hline
\bottomrule
\end{tabular}
\caption{Fourteen attribution methods can be unified into the same paradigm of allocating Taylor interaction effects.}
\label{tab:reformulation}
\end{table*}

\noindent 
\textbf{Occlusion-1.} 
To compute the attribution of the input variable $i$, Occlusion-1 \cite{zeiler2014visualizing} occludes the variable $i$ by the baseline value $b_i$ and obtains an occluded input $\bm{x}|_{x_i = b_i}$. 
Then, Occlusion-1 considers that the output change between the original input $\bm{x}$ and the occluded input $\bm{x}|_{x_i = b_i}$ reflects the attribution of the variable $i$.
\begin{equation} 
\begin{small}
\begin{aligned}
a_i = f(\bm{x}) - f(\bm{x}|_{x_i = b_i}).
\end{aligned}
\end{small}
\end{equation}
where $\forall i, b_i = b$ and $b$ is a constant scalar.

\begin{theorem}
\label{thm:occ1}
\textit{(Proof in Appendix B) In the Occlusion-1 method, the attribution of the input variable $i$ can be reformulated as}
\begin{equation} 
\begin{small}
\begin{aligned}
a_i & = \sum_{\bm{\kappa} \in \Omega_i} \phi(\bm{\kappa}) +  \sum_{|S| > 1, i \in S} \sum_{\bm{\kappa} \in \Omega_S} I(\bm{\kappa}).
\end{aligned}
\end{small}
\end{equation}
\end{theorem}
\vspace{2pt}

Theorem \ref{thm:occ1} shows that the Occlusion-1 method also follows the paradigm of  allocating Taylor interaction effects in Eq.~(\ref{eqn:unifiedattribution2}).
This method allocates the generic independent effect {\small $\psi(i) =  \sum\nolimits_{\bm{\kappa} \in \Omega_i} \phi(\bm{\kappa})$} of the variable $i$ to its attribution $a_i$.
In addition, this method allocates each generic interaction effect {\small $J(S) = \sum\nolimits_{\bm{\kappa} \in \Omega_S} I(\bm{\kappa})$}, which involves the variable  $i$ ($i \in $ {\small $S$}), 
to the attribution $a_i$.
In other words, the Occlusion-1 method repeatedly allocates the generic interaction effect {\small $J(S)$} to each variable in {\small $S$}.
\\

\noindent
\textbf{Occlusion-patch.}
Occlusion-patch  \cite{zeiler2014visualizing} first divides an image into $m$ patches, \textit{i.e.}, $M = \{S_1, \cdots, S_m\}$.
To compute the attribution of pixels in each patch $S_j$, Occlusion-patch occludes all pixels in the patch by the baseline value $\bm{b}$ and obtains an occluded input $\bm{x}|_{\bm{x}_{S_j} = \bm{b}}$. Then,  Occlusion-patch considers that the output change between the original input and the occluded input reflects the attribution of pixels in $S_j$. 
\begin{equation}
\begin{small}
\begin{aligned}
a_i = f(\bm{x}) - f(\bm{x}|_{\bm{x}_{S_j} = \bm{b}}), \ \forall i \in S_j
\end{aligned}
\end{small}
\end{equation}

\begin{theorem}
\label{thm:occp}
\textit{(Proof in Appendix B) In the Occlusion-patch method, the attribution of the pixel $i$ in the patch $S_j$ ($i \in S_j$) can be reformulated as}
\begin{equation}
\begin{small}
\begin{aligned}
a_i & = \sum_{m \in S_j} \sum_{\bm{\kappa} \in \Omega_m}\phi(\bm{\kappa}) + \sum_{|S|>1, S \cap S_j \neq \emptyset} \sum_{\bm{\kappa} \in \Omega_S} I(\bm{\kappa}).
\end{aligned}
\end{small}
\end{equation}
\end{theorem}

Theorem \ref{thm:occp} shows that the Occlusion-patch method follows the paradigm of  allocating Taylor interaction effects in Eq.~(\ref{eqn:unifiedattribution2}).
Specifically, for the pixel {\small $i \in S_j$},
this method allocates generic independent effects of all pixels in {\small $S_j$}  to the attribution $a_i$, \textit{i.e.}, allocating {\small $\sum_{k \in S_j} \sum_{\bm{\kappa} \in \Omega_k}\phi(\bm{\kappa}) $} to $a_i$.
In addition, this method allocates all generic interaction effects {\small $J(S) = \sum_{\bm{\kappa} \in \Omega_S} I(\bm{\kappa})$}, which involve some pixels in {\small $S_j$ ({\small $S \cap S_j \neq \emptyset$})},  to the attribution $a_i$.
Hence, the Occlusion-patch method may mistakenly assign the generic interaction effect {\small $J(S)$}, which does not involve the variable $i$ ($i\not\in S$), to the attribution of the variable $i$.
\\

\noindent
\textbf{Prediction Difference.} 
The Prediction Difference method \cite{zintgraf2017visualizing} is an extension of the Occlusion-1 method \cite{zeiler2014visualizing}.
Unlike the Occlusion-1 method  simply using a single baseline value to represent the occlusion state of $x_i$, the Prediction Difference method samples multiple baseline points from a distribution $p(b_i)$. 
For example, the distribution can be set as the conditional distribution of $x_i$ given other  variables, $p(b_i) = p(x_i|x_1, x_2, \dots, x_{i-1}, x_{i+1}, \dots, x_n)$.  
However, this method assumes $b_1 = b_2 = \dots = b_n = b$, here. 
Then, the attribution $a_i$ is computed by averaging attributions generated by the Occlusion-1 method over different baseline points. 
\begin{equation} 
\begin{small}
\begin{aligned}
a_i = \mathbb{E}_{b_i \sim p(b_i)} [f(\bm{x}) -  f(\bm{x}|_{x_i = b_i})].
\end{aligned}
\end{small}
\end{equation}

\begin{theorem}\label{thm:Prediction Difference}
\textit{(Proof in Appendix B) In the Prediction Difference method, the attribution of the input variable $i$ can be reformulated as
\begin{equation}
\begin{small}
\begin{aligned}
    a_i   &= \mathbb{E}_{\bm{b} \sim p(\bm{b})}  [ \sum_{\bm{\kappa} \in \Omega_i} \phi(\bm{\kappa}|\bm{b}) +  \sum_{|S|>1, i \in S} \sum_{\bm{\kappa} \in \Omega_S} I(\bm{\kappa}|\bm{b})],  \\
\end{aligned}
\end{small}
\end{equation}
where $\bm{b} = b \cdot \bm{1}$}.
\end{theorem}
\vspace{2pt}

Theorem \ref{thm:Prediction Difference} shows that the Prediction Difference method follows the paradigm of allocating Taylor interaction effects in Eq.~(\ref{eqn:unifiedattribution2}).
This method allocates the generic independent effect {\small $\sum_{\bm{\kappa} \in \Omega_i} \phi(\bm{\kappa}|\bm{b})$} of the variable $i$ to its attribution $a_i$.
Similarly, this method allocates the entire Taylor interaction effect  {\small $I(\bm{\kappa}|\bm{b})$} ({\small $\bm{\kappa} \in \Omega_S, i \in S$}), which involves the variable $i$, to the attribution $a_i$.
 This method then averages the attributions $a_i$ over different baseline points $\bm{b}$, to obtain the final attribution.
\\

\begin{figure*}[t]\centering
\includegraphics[width = 0.95 \textwidth]{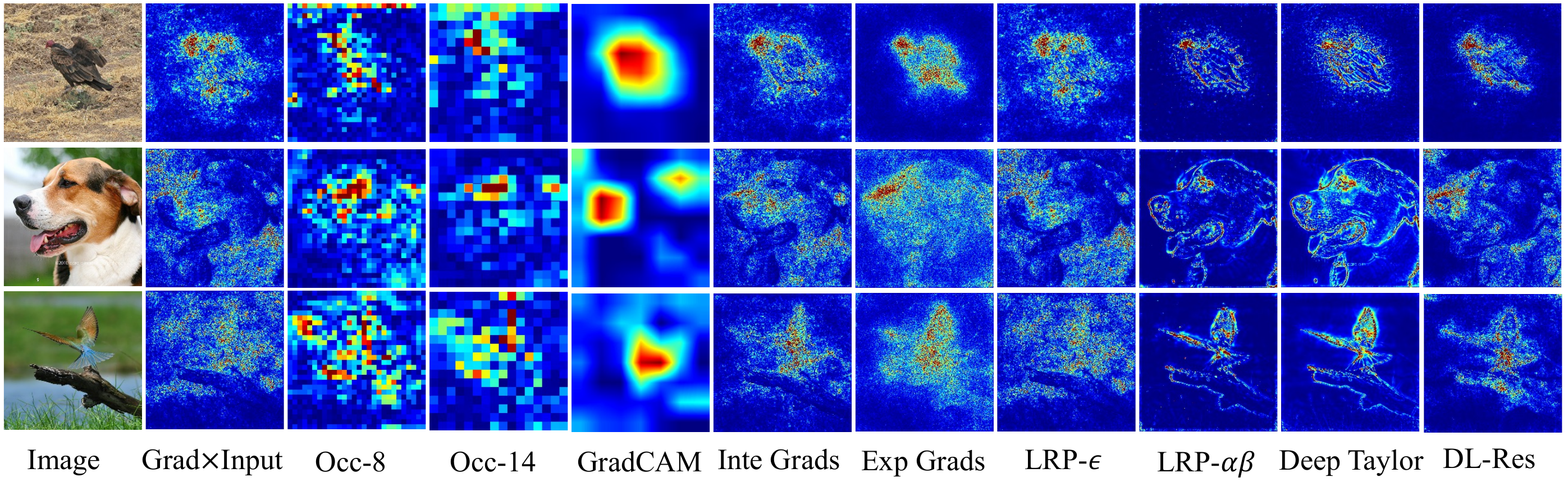}
\vspace{-1mm}
\caption{Attribution maps generated by different attribution methods.}
\label{attribution maps}
\end{figure*}

\noindent
\textbf{Grad-CAM.} 
Grad-CAM \cite{selvaraju2017grad} estimates the attribution of neural activations at each location $(i,j)$ in a convolutional layer, as follows. 
\begin{equation}\label{eqn:GradCAMformula}
\begin{small}
\begin{aligned}
a_{ij} &= {\rm ReLU}(\sum\nolimits_{k=1}^{K} \alpha_k A_{ij}^k),  \\
 \text{where}  \ \alpha_k &= \frac{1}{W \times H}\sum\nolimits_{i=1}^W\sum\nolimits_{j=1}^H \frac{\partial y}{\partial A_{ij}^k}.
\end{aligned}
\end{small}
\end{equation}
where $A^k \in \mathbb{R}^{W \times H}$ denotes the feature map of the $k$-th channel, 
and $A^k_{ij}$ denotes the neuron activation at the $(i,j)$ location in the feature map $A^k$.  

Grad-CAM can be understood as follows. 
Grad-CAM considers that the activation strength {\small $A^k_{ij}$} of different neurons $(i,j,k)$ reflects the importance of different neurons. 
For example, if an input sample activates the neuron {\small $A^k_{ij}$} strongly, then Grad-CAM considers that the neuron $(i,j,k)$ is important for the inference of the sample.
Moreover, feature maps in different channels have different importance. 
Hence, Grad-CAM re-weights feature maps of the $k$-th channel, using the average gradient of this channel $\alpha_k$. 

To simplify the analysis of Grad-CAM, we just explain the following attribution $\tilde a_{ij}$ before the ReLU operation in Eq.(\ref{eqn:GradCAMformula}), subject to $a_{ij} = ReLU(\tilde a_{ij})$.
\begin{equation}\label{eqn:Grad-CAMformula}
\begin{small}
\begin{aligned}
\tilde a_{ij} = & \sum\nolimits_{k=1}^{K} \alpha_k A_{ij}^k.  
\end{aligned}
\end{small}
\end{equation}
In fact, previous work \cite{selvaraju2017grad} has proven that Grad-CAM actually explains a DNN as the following linear model of global average pooled feature maps {\small $F$}. Here, {\small $F = [F_1, \dots, F^K]$} and {\small $F^k = \frac{1}{W\times H}\sum_{i=1}^W \sum_{j=1}^H A_{ij}^k$.}
\begin{equation}\label{eqn:revisitGradCAM}
\begin{small}
\begin{aligned}
y \xlongequal{\text{explained}} g(F) = \sum\nolimits_{k=1}^K \frac{\partial y}{\partial F^k} \cdot F^k. 
\end{aligned}
\end{small}
\end{equation}
Based on the above conclusion, we prove Theorem \ref{thm:gradcam}. 
\vspace{2pt}

\begin{theorem}
\label{thm:gradcam}
\textit{(Proof in Appendix B) In the Grad-CAM method, let us consider each neuron {\small $A_{ij}^k$} in the convolutional layer as an input variable. Then, the attribution of each input variable can be reformulated as follows.}
\begin{equation}
\begin{small}
\begin{aligned}
    \tilde a_{ij}^k = \phi(\bm{\kappa}) = \frac{\partial g(F)}{\partial A_{ij}^k} A_{ij}^k.
    \end{aligned}
    \end{small}
    \end{equation}
\textit{where $g(F)$ is the explanatory model of the DNN in the Grad-CAM method. 
In addition, $\bm{\kappa} = [\kappa_1, \cdots, \kappa_n]$ is a one-hot degree vector with $\kappa_i = 1$ and  $\forall j \neq i, \kappa_j = 0$.}
\end{theorem}
\vspace{2pt}

Theorem \ref{thm:gradcam} shows that the Grad-CAM method follows the paradigm of  allocating  Taylor interaction effects in Eq.~(\ref{eqn:unifiedattribution2}).
Specifically, Grad-CAM only allocates a specific Taylor independent effect {\small $\phi(\bm{\kappa})$} of the neuron {\small $A^k_{ij}$} to its attribution.

By comparing Theorem \ref{thm:gradinput} and Theorem \ref{thm:gradcam}, we find that Grad-CAM and Gradient$\times$Input share similar mechanisms, \textit{i.e.}, both methods can be explained as exclusively using the Taylor independent effect to compute the attribution.
However, the main difference between the two methods is that Grad-CAM explains the attribution  of features in the convolutional layer, whereas Gradient$\times$Input explains the attribution of input variables.
\\

\noindent
\textbf{Integrated Gradients.}
The Integrated Gradients method \cite{sundararajan2017axiomatic} estimates the attribution of each input variable as follows.
\begin{equation} \label{eqn:IG}
\begin{small}
\begin{aligned}
a_{i} & = ( x_i - b_i) \cdot \int_{\alpha = 0}^1 \frac{\partial f(c)}{\partial {c_i}}\bigg|_{\bm{c} = (\bm{b} + \alpha( \bm{x} - \bm{b}))} d\alpha \\
\end{aligned}
\end{small}
\end{equation}
The Integrated Gradients method estimates the attribution as the product of the input feature and the \textit{average} gradient of the output \textit{w.r.t.} the input feature, where the \textit{average} gradient is computed over numerous input points along a linear path from the baseline point $\bm{b}$ to the given input $\bm{x}$.
\vspace{2pt}

\begin{theorem}
\label{thm:ig}
\textit{
(Proof in Appendix B) In the Integrated Gradients method, the attribution of the input variable $i$ can be reformulated as}
\begin{equation}\label{eqn:IGreformulation}
\begin{small}
\begin{aligned}
\!\!\! a_i & = \sum_{\bm{\kappa} \in \Omega_i} \!\! \phi(\bm{\kappa}) \! + \sum_{\substack{|S|>1, i \in S}}\sum_{\bm{\kappa} \in \Omega_S} \frac{\kappa_i}{\sum_{i'} \kappa_{i'}}  I(\bm{\kappa}) \\
\end{aligned}
\end{small}
\end{equation}
\end{theorem}

Theorem \ref{thm:ig} shows that the Integrated Gradients method follows the paradigm of  allocating Taylor interaction effects in Eq.~(\ref{eqn:unifiedattribution2}).
This method allocates the generic independent effect {\small $\psi(i) = \sum_{\bm{\kappa}\in \Omega_i}\phi(\bm{\kappa})$} of the variable $i$ to the attribution $a_i$. 
In addition, this method allocates each Taylor interaction effect 
{\small $I(\bm{\kappa})$} ({\small $\bm{\kappa} \in \Omega_S, i \in S$}), which involves the variable $i$,  to the attribution $a_i$. 
The weight of allocating {\small $I(\bm{\kappa})$} is proportional to the degree $\kappa_i$ of the variable $i$.
\\


\noindent
\textbf{Expected Gradients.} 
The Expected Gradients method \cite{erion2021improving} is an extension of the Integrated Gradients method \cite{sundararajan2017axiomatic}. 
To estimate the attribution, the Expected Gradients method samples baseline points from a prior distribution {\small $p(\bm{b})$} (\textit{e.g.}, {\small $\bm{b} \sim N(\bm{x}, \sigma^2\bm{I})$}), instead of specifying a certain baseline point in the Integrated Gradients method. 
Then, the attribution $a_i$ of the variable $i$ is computed by integrating attributions generated by the Integrated Gradients method over different baselines.
\begin{equation}
\begin{small}
\begin{aligned}
a_i = \mathbb{E}_{\bm{b} \sim p(\bm{b})}\cdot [(x_i-b_i)\int_{0}^1 \frac{\partial f(c)}{\partial {c_i}}\bigg|_{\bm{c} = (\bm{b} + \alpha( \bm{x} - \bm{b}))} d\alpha] 
\end{aligned}
\end{small}
\end{equation}

\begin{theorem}\label{thm:Expected Gradients}
\textit{(Proof in Appendix B) In the Expected Gradients method, the attribution of the variable $i$ can be reformulated as:}
\begin{equation}
\begin{small}
\begin{aligned}
    a_i   &= \mathbb{E}_{\bm{b} \sim p(\bm{b})}  [\sum_{\bm{\kappa}\in \Omega_i} \phi(\bm{\kappa}|\bm{b}) + \sum_{\substack{|S|>1, i \in S}} \sum_{\bm{\kappa} \in \Omega_S} \frac{\kappa_i}{\sum_{ i'} \kappa_{i'}} I(\bm{\kappa}|\bm{b})] \\
\end{aligned}
\end{small}
\end{equation}
\end{theorem}

Theorem \ref{thm:Expected Gradients} shows that the Expected Gradients method follows the paradigm of allocating Taylor interaction effects in Eq.~(\ref{eqn:unifiedattribution2}).
This method allocates the generic independent effect {\small $\mathbb{E}_{\bm{b} \sim p(\bm{b})} [\sum_{\bm{\kappa}\in \Omega_i}\phi(\bm{\kappa}|\bm{b})]$} of the variable $i$, which is averaged over different baseline points $\bm{b}$, to the attribution $a_i$.
In addition, this method allocates the average Taylor interaction effect  {\small $\mathbb{E}_{\bm{b} \sim p(\bm{b})}I(\bm{\kappa}|\bm{b})$} ({\small $\bm{\kappa} \in \Omega_S, i \in S$}), which involves the variable $i$ ($i \in S$), to the attribution $a_i$. The weight of allocating $I(\bm{\kappa})$ to the variable $i$ is the relative degree of the variable $i$, {\small $\frac{\kappa_i}{\sum_{i'} \kappa_{i'}}$}.  
\\

\noindent
\textbf{Shapley value.}
The Shapley value method \cite{hart1989shapley,lundberg2017unified} estimates the attribution of each variable as follows.
\begin{equation}\label{eqn:Shapley}
\begin{small}
\begin{aligned}
 a_i &=  \sum\nolimits_{S \subseteq N \setminus \{i\}} p(S) \cdot [f(\bm{x}_{S \cup \{i\}}) - f(\bm{x}_S)].\\
\end{aligned}
\end{small}
\end{equation}
where {\small $p(S) = |S|!(n-1-|S|)!/n!$}. 
The Shapley value method formulates the attribution of the variable $i$ as its average marginal contribution {\small $f(\bm{x}_{S \cup \{i\}}) - f(\bm{x}_S)$}  over different contextual subsets {\small $S$}. 
Here, {\small $f(\bm{x}_S)$} is computed as the network output when variables in $N \verb|\| S$ are masked and variables in $S$ keep unchanged.  
\vspace{2pt}

\begin{theorem}\label{thm:shapley}
\textit{(Proof in Appendix B) In the Shapley value method, the attribution of the input variable $i$ can be reformulated as
\begin{equation}\label{eqn:shapleyreformulation}
\begin{small}
\begin{aligned}
a_i & =  \sum_{\bm{\kappa} \in \Omega_i} \phi(\bm{\kappa}) +  \sum_{\substack{|S|>1, i \in S}}\sum_{\bm{\kappa} \in \Omega_S}  \frac{1}{|S|} I(\bm{\kappa})
 \end{aligned}
 \end{small}
\end{equation}
}
\end{theorem}

Theorem \ref{thm:shapley} shows that the Shapley value method follows the paradigm of  allocating Taylor interaction effects in Eq.~(\ref{eqn:unifiedattribution2}).
This method allocates the generic independent effect {\small $\psi(i) =  \sum_{\bm{\kappa}\in \Omega_i} \phi(\bm{\kappa})$} of the variable $i$ to the attribution $a_i$. 
Furthermore, this method allocates each Taylor interaction effect {\small $I(\bm{\kappa})$} ({\small $\bm{\kappa} \in \Omega_S, i \in S$}), which involves the variable $i$, to the attribution $a_i$.
The effect {\small $I(\bm{\kappa})$}  is uniformly allocated to all the  {\small $s = |S|$} variables involved in the interaction {\small $S$}, \textit{i.e.}, each input variable receives {\small $\frac{1}{|S|}I(\bm{\kappa})$}. 
\\

\noindent
\textbf{Back-propagation attribution methods.} 
Among various attribution methods, a typical type of method is designed to estimate the attribution of each feature dimension at an intermediate layer, and then back-propagate these attributions to previous layers until the input layer. 
That is, {\small $\bm{a}^{(L)}\rightarrow\bm{a}^{(L-1)}\rightarrow \cdots \rightarrow \bm{a}^{(1)} \rightarrow \bm{a}^{(0)}$}, 
where {\small $\bm{a}^{(l)} \in \mathbb{R}^{n_l}$} denotes attributions of all feature dimensions in the $l$-th layer. 
In particular, {\small $\bm{a}^{(0)} \in \mathbb{R}^n$} corresponds to attributions in the input layer. 
This type of method is known as \textbf{\textit{back-propagation attribution methods}}, including LRP-$\epsilon$  \cite{bach2015pixel}, LRP-$\alpha\beta$  \cite{bach2015pixel}, Deep Taylor \cite{montavon2017explaining}, DeepLIFT Rescale \cite{shrikumar2017learning}, Deep SHAP \cite{lundberg2017unified}, DeepLIFT RevealCancel \cite{shrikumar2017learning},  and so on.

The \textbf{essential difference} between different back-propagation attribution methods is that they employ different recursive rules for back-propagating attributions through adjacent layers, \textit{i.e.},
{\small $\bm{a}^{(l)}\rightarrow\bm{a}^{(l-1)}$}. 
In particular, these methods usually simplify various layer-wise operations in different DNNs as the combination of linear operations and nonlinear activations.
These methods mainly define the rule of back-propagating attributions through the typical module {\small $\bm{x}^{(l)} = \sigma(W\bm{x}^{(l-1)} + \bm{s})$} as a representative. 
Here, {\small $\bm{x}^{(l)}$} denotes the feature in the $l$-th layer.  
{\small $W$} and {\small $\bm{s}$} denote the weight and the additive bias, respectively. $\sigma$ is the activation function.
\\

\noindent
\textbf{LRP-$\epsilon$.} 
LRP-$\epsilon$ \cite{bach2015pixel}  is a typical back-propagation attribution method, which back-propagates attributions in a layer-wise manner.
Specifically, for the typical module {\small $\bm{x}^{(l)} = \sigma(W\bm{x}^{(l-1)} + \bm{s})$}, 
 LRP-$\epsilon$ propagates the following numerical value {\small $a_{i \leftarrow j}^{(l)}$} from the attribution {\small $a_j^{(l)}$} in the $l$-th layer to the attribution {\small $a_i^{(l-1)}$} in the $(l-1)$-th layer. 
\begin{equation}
\begin{small}
\begin{aligned}
 a_{i \leftarrow j}^{(l)} =
 \left \{
\begin{array}{rcl}
 \frac{z_{ij}}{(\sum_{i'}z_{i'j} + s_j)  + \epsilon} \cdot a_j^{(l)}, & \sum_{i'}z_{i'j} + s_j \geq 0 \\
  \frac{z_{ij}}{(\sum_{i'}z_{i'j} + s_j)  - \epsilon} \cdot a_j^{(l)}, & \sum_{i'}z_{i'j} + s_j < 0 \\
 \end{array}\right. 
 \\[1mm]
\end{aligned}
\end{small}
\end{equation}
Here, {\small $z_{ij} =  W_{ij} x_i^{(l-1)}$} and  {\small $x_j^{(l)} = \sigma(\sum_{i'} z_{i'j} + s_j)$}, so LRP-$\epsilon$ considers that {\small $z_{ij}$} can reflect the contribution of {\small $x_i^{(l-1)}$} to {\small $x_j^{(l)}$},
to some extent. 
To avoid dividing 0, LRP-$\epsilon$ introduces a small quantity $\epsilon>0$ in the denominator.
Then, LRP-$\epsilon$ formulates the attribution {\small $a_i^{(l-1)}$} as the sum of these propagated values from all feature dimensions in the $l$-th layer, \textit{i.e.}, {\small $a_i^{(l-1)} = \sum\nolimits_j a_{i \leftarrow j}^{(l)}$}.
\vspace{2pt}

\begin{theorem}\label{thm:lrp}
\textit{(Proof in Appendix B) When  ReLU  is used as the activation function, the attribution of the variable $i$ estimated by the LRP-$\epsilon$ method can be reformulated as}
\begin{equation}
\begin{small}
\begin{aligned}
a_i = \phi(\bm{\kappa}) = \frac{\partial f(\bm{x})}{\partial x_i}x_i
\end{aligned}
\end{small}
\end{equation}
    \textit{where $\bm{\kappa}= [\kappa_1, \cdots, \kappa_n]$ is a one-hot degree vector with $\kappa_i = 1$ and  $\forall j \neq i, \kappa_j = 0$.}
\end{theorem}
\vspace{2pt}

Theorem \ref{thm:lrp} shows that the LRP-$\epsilon$ method  follows the paradigm of  allocating Taylor interaction effects in Eq.~(\ref{eqn:unifiedattribution2}). 
Specifically,  this method allocates only a specific Taylor independent effect {\small $\phi(\bm{\kappa})$} of the input variable $i$ to the attribution of the input variable $i$.

By comparing Theorem \ref{thm:gradinput}  and Theorem \ref{thm:lrp}, 
it is easy to find that \textit{the LRP-$\epsilon$ method and the Gradient$\times$Input method are essentially the same}, because the two methods allocate the Taylor interaction effects in the same way when ReLU is adopted as the activation function.
Furthermore, Figure \ref{attribution maps} also verifies that the two methods produce the same attribution results. \\

\noindent
\textbf{LRP-$\alpha\beta$.} 
LRP-$\alpha\beta$ \cite{bach2015pixel} is also a typical back-propagation attribution method. It slightly modifies the recursive propagation rule of LRP-$\epsilon$ as follows, 
\begin{equation}\label{eqn:LRPalphabeta}
\begin{small}
\begin{aligned}
a_{i \leftarrow j}^{(l)} = 
\left \{
\begin{array}{rcl}
 \frac{\alpha \cdot z_{ij}}{\sum_{i' \in N^+} z_{i'j} 
 } \cdot a_j^{(l)},  & i \in N^+  \\[2mm]
\frac{\beta \cdot z_{ij}}{\sum_{i' \in N^-} z_{i'j} 
}   \cdot a_j^{(l)},    & i \in N^-
\end{array}\right. 
\end{aligned}
\end{small}
\end{equation}
where {\small $a_{i \leftarrow j}^{(l)}$} denotes the propagated attribution from the attribution {\small $a_j^{(l)}$} in the $l$-th layer to {\small $a_i^{(l-1)}$} in the $(l-1)$-th layer. 
In addition, {\small $z_{ij} = W_{ij} x_i^{(l-1)}$}, {\small $N^+ = \{i|z_{ij}>0\}$}, and {\small $N^- = \{i|z_{ij} \leq 0\}$}. 
Unlike LRP-$\epsilon$, LRP-$\alpha\beta$ divides all contribution scores {\small $z_{ij}$} into two groups, \textit{i.e.},  the group {\small $N^+$} subject to {\small $z_{ij}>0$} and the group {\small $N^-$} subject to {\small $z_{ij} \leq 0$}.
Then, LRP-$\alpha\beta$ computes the attribution in each group separately. 
Here, $\alpha$ and $\beta$ are the pre-defined weights for two groups. 
Finally, the attribution {\small $a_i^{(l-1)}$} is computed as {\small $a_i^{(l-1)} = \sum\nolimits_j a_{i \leftarrow j}^{(l)}$}.
\vspace{2pt}

\noindent
\begin{theorem}\label{thm:LRPab}
\textit{(Proof in Appendix B) Let us consider the feature dimension {\small $x_j^{(l)}$} as the target output and all feature dimensions in the $(l-1)$-th layer as input variables (i.e., {\small $y = x_j^{(l)}, \bm{x} = \bm{x}^{(l-1)}$}), so as to analyze the layer-wise propagation of attributions. 
Then,  for the input variable {\small $i \in N^+$}, the attribution estimated by the  LRP-$\alpha\beta$ method can be reformulated as follows. 
\begin{equation}\label{eqn:LRPab+}
\begin{small}
\begin{aligned}
\!\!\!\! a_i \! = \!
\alpha [\sum_{\bm{\kappa}\in \Omega_i} \phi(\bm{\kappa}) + \! \sum_{|S| > 1, i \in S} \! \sum_{\bm{\kappa} \in \Omega_S}  c_i I(\bm{\kappa})  + \!\!\! \sum_{S \subseteq N^-} \sum_{\bm{\kappa} \in \Omega_S} d_i I(\bm{\kappa})] 
\end{aligned}
\end{small}
\end{equation}
where $c_i = \frac{\kappa_i}{\sum_{i' \in N^+} \kappa_{i'}}$, and $d_i =  \frac{z_{ij}}{\sum_{i' \in N^+} z_{i'j}}$. Similarly, for the input variable  {\small $i \in N^-$}, the attribution estimated by the  LRP-$\alpha\beta$ method  can be reformulated as follow. }
\begin{equation}\label{eqn:LRPab-}
\begin{small}
\begin{aligned}
\!\!\!\! a_i \! = \!
\beta   [\sum_{\bm{\kappa}\in \Omega_i} \phi(\bm{\kappa}) + \! \sum_{|S|> 1, i \in S} \sum_{\bm{\kappa} \in \Omega_S}  \tilde c_i I(\bm{\kappa}) + \!\!\!  \sum_{S \subseteq N^+}\sum_{\bm{\kappa} \in \Omega_S} \tilde d_i I(\bm{\kappa})] 
\end{aligned}
\end{small}
\end{equation}
where $\tilde c_i = \frac{\kappa_i}{\sum_{i' \in N^-} \kappa_{i'}}$, and $\tilde d_i =  \frac{z_{ij}}{\sum_{i' \in N^-} z_{i'j}}$.
\end{theorem}
\vspace{3pt}

Theorem \ref{thm:LRPab} shows that the LRP-$\alpha\beta$ method follows the paradigm of  allocating Taylor interaction effects in Eq.~(\ref{eqn:unifiedattribution2}). 
As Eq.~(\ref{eqn:LRPab+}) shows, for the input variable $i \in$ {\small $N^+$},  this method allocates part of $i$'s  Taylor independent effect {\small $\phi(\bm{\kappa})$} ({\small $\bm{\kappa} \in \Omega_i$}) to the attribution $a_i$. 
In addition, this method allocates part of the Taylor interaction effect {\small $I(\bm{\kappa})$} ({\small $\bm{\kappa} \in \Omega_S, i \in S$}), which involves the variable $i$, to the attribution $a_i$. 
However, this method mistakenly allocates part of the Taylor interaction effect {\small $I(\bm{\kappa})$} ({\small $\bm{\kappa} \in \Omega_S, S \subseteq N^-$}) between some variables in {\small $N^-$}  to the attribution of the variable $i$ {\small $\in N^+$}. 
Similarly, according to Eq.~(\ref{eqn:LRPab-}), for the input variable {\small $i \in N^-$}, this method mistakenly allocates part of the Taylor interaction effect {\small $I(\bm{\kappa})$} ({\small $\bm{\kappa} \in \Omega_S, S \subseteq N^+$}) between some variables in {\small $N^+$} to the attribution of the variable $i$ {\small $\in N^-$}. 
 \\

\noindent
\textbf{Deep Taylor.} 
Deep Taylor \cite{montavon2017explaining} is a typical back-propagation attribution method.
For the typical module {\small $\bm{x}^{(l)} = \sigma(W\bm{x}^{(l-1)} + \bm{s})$}, it designs the recursive back-propagation rule as follows. 
\begin{equation}
\begin{small}
\begin{aligned}
a_{i \leftarrow j}^{(l)} = \left \{
\begin{array}{ccl}
\frac{z_{ij}}{\sum_{i' \in N^+} z_{i'j}} \cdot a_j^{(l)}, & i \in N^+  \\[1mm]
0,  & i \in N^-
\end{array}\right. 
\end{aligned}
\end{small}
\end{equation}
where {\small $z_{ij} = W_{ij} x_i^{(l-1)}$}, {\small $N^+ = \{i|z_{ij}>0\}$}, and {\small $N^- = \{i|z_{ij} \leq 0\}$}. 
Then, the attribution {\small $a_i^{(l-1)}$} in the $(l-1)$-th layer is computed as {\small $a_i^{(l-1)} = \sum\nolimits_j a_{i \leftarrow j}^{(l)}$}.
In particular, Deep Taylor can be regarded as a special case of LRP-$\alpha\beta$ \cite{bach2015pixel}  with $\alpha = 1, \beta = 0$ in Eq. (\ref{eqn:LRPalphabeta}). 
\vspace{2pt}

\noindent
\begin{theorem}\label{thm:DeepTaylor}
\textit{(Proof in Appendix B) Let us consider the feature dimension {\small $x_j^{(l)}$} as the target output and all feature dimensions in the $(l-1)$-th layer as input variables (i.e., {\small $y = x_j^{(l)}, \bm{x} = \bm{x}^{(l-1)}$}), so as to analyze the layer-wise propagation of attributions. 
Then, in the Deep Taylor method, the attribution of  the variable {\small $i \in N^+$} is reformulated as follows.}
\begin{equation}\label{eqn:deepTaylor}
\begin{small}
\begin{aligned}
\!\!\!\! a_i \! = \!
\sum_{\bm{\kappa}\in \Omega_i} \phi(\bm{\kappa}) + \! \sum_{|S| > 1, i \in S}  \sum_{\bm{\kappa} \in \Omega_S}  c_iI(\bm{\kappa})  + \!\!\! \sum_{S \subseteq N^-} \sum_{\bm{\kappa} \in \Omega_S} d_i I(\bm{\kappa}) 
\end{aligned}
\end{small}
\end{equation} 
where $c_i = \frac{\kappa_i}{\sum_{i' \in N^+} \kappa_{i'}}$, and $d_i =  \frac{z_{ij}}{\sum_{i' \in N^+} z_{i'j}}$. \textit{Moreover, for the variable {\small $i \in N^-$}, $a_i= 0$.}
\end{theorem}
\vspace{2pt}

Theorem \ref{thm:DeepTaylor} shows that the Deep Taylor method follows the paradigm of  allocating Taylor interaction effects in Eq.~(\ref{eqn:unifiedattribution2}).
The weight of allocation is almost the same as the weight of LRP-$\alpha\beta$ in Eq.~(\ref{eqn:LRPab+}) and Eq.~(\ref{eqn:LRPab-}), and differs only by a constant. 
\\

\noindent
\textbf{DeepLIFT Rescale.}
DeepLIFT Rescale \cite{shrikumar2017learning} is also a typical back-propagation attribution method, which propagates the attribution from {\small $a_j^{(l)}$} in the $l$-th layer to the attribution {\small $a_i^{(l-1)}$} in the $(l-1)$-th layer as follows. 
\begin{equation}\label{eqn:deeplift-1}
\begin{small}
\begin{aligned}
a_{i \leftarrow j}^{(l)}   &= \frac{\Delta z_{ij}}{\sum_{i'}(\Delta z_{i'j})} \cdot a_j^{(l)}
\end{aligned}
\end{small}
\end{equation}
where {\small $ \Delta z_{ij} = z_{ij} - \tilde z_{ij}$}, {\small $z_{ij} = W_{ij}x_i^{(l-1)}$}, {\small $\tilde z_{ij} = W_{ij} \tilde x_i^{(l-1)}$}. 
Here, {\small $\tilde x_i^{(l-1)}$} is the selected baseline value to represent the state when {\small $x_i^{(l-1)}$} does not receive any information. 
Thus, {\small $\Delta z_{ij}$} reflects the contribution of {\small $x_i^{(l-1)}$} on changing {\small $x_j^{(l)}$} from the state of the baseline value to the current activation value.
Then, the attribution {\small $a_i^{(l-1)}$} is computed as {\small $a_i^{(l-1)} = \sum\nolimits_j a_{i \leftarrow j}^{(l)}$}.
\vspace{2pt}

\begin{theorem}\label{thm:deepliftrescale}
\textit{(Proof in Appendix B) Let us consider  the feature dimension {\small $x_j^{(l)}$} as the target output and features in the $(l-1)$-th layer as input variables (i.e., {\small $y = x_j^{(l)}, \bm{x} = \bm{x}^{(l-1)}$}),  so as to analyze the layer-wise propagation of attributions. 
Then,  the attribution of the input variable $i$ estimated by  the DeepLIFT Rescale method can be reformulated as}
\begin{equation}
\begin{small}
\begin{aligned}
a_i & = \sum_{\bm{\kappa} \in \Omega_i} \phi(\bm{\kappa}) + \sum_{\substack{|S|>1, i \in S}}\sum_{\bm{\kappa} \in \Omega_S} \frac{\kappa_i}{\sum_{i'} \kappa_{i'}}  I(\bm{\kappa}) \\
\end{aligned}
\end{small}
\end{equation}
\end{theorem}

Theorem \ref{thm:deepliftrescale} shows that the DeepLIFT Rescale method  follows the paradigm of  allocating Taylor interaction effects in Eq.~(\ref{eqn:unifiedattribution2}). 
This method allocates the generic independent effect {\small $\psi(i) = \sum_{\bm{\kappa} \in \Omega_i} \phi(\bm{\kappa})$} of the variable $i$ to the attribution $a_i$. 
In addition, this method allocates part of each Taylor interaction effect {\small $I(\bm{\kappa})$ ($\bm{\kappa} \in \Omega_S, i \in S$)}, which involves the variable $i$,  to the attribution $a_i$.
The weight of allocating {\small $I(\bm{\kappa})$} to the variable $i$ is  proportional to  the degree $\kappa_i$.
\\

\noindent
\textbf{Deep SHAP.}
Deep SHAP \cite{lundberg2017unified} is a typical back-propagation attribution method, which combines the Shapley value method \cite{hart1989shapley,lundberg2017unified} to propagate the attribution from {\small $a_j^{(l)}$} in the $l$-th layer to {\small $a_i^{(l-1)}$} in the $(l-1)$-th layer. 
\begin{equation}\label{eqn:deeplift-1}
\begin{small}
\begin{aligned}
a_{i \leftarrow j}^{(l)}   &= \frac{\phi_i(x_j^{(l)})}{\sum_{i'} \phi_{i'}(x_j^{(l)})} \cdot a_j^{(l)}
\end{aligned}
\end{small}
\end{equation}
where  {\small $\phi_i(x_j^{(l)})$} denotes the Shapley value of {\small $x_i^{(l-1)}$} \textit{w.r.t.} {\small $x_j^{(l)}$} when we consider  {\small $x_j^{(l)}$} as the output and consider features {\small $x_i^{(l-1)}$} as input variables. 
Finally, the attribution {\small $a_i^{(l-1)}$} is computed as {\small $a_i^{(l-1)} = \sum\nolimits_j a_{i \leftarrow j}^{(l)}$}.
\vspace{2pt}

\begin{theorem}\label{thm:deepshap}
\textit{(Proof in Appendix B) Let us consider  the feature dimension {\small $x_j^{(l)}$} as the target output and features in the $(l-1)$-th layer as input variables (i.e., {\small $y = x_j^{(l)}, \bm{x} = \bm{x}^{(l-1)}$}), to analyze the layer-wise propagation of attributions. 
Then, in the Deep SHAP method,  the attribution of the input variable $i$ can be reformulated as follows.}
\begin{equation}
\begin{small}
\begin{aligned}
a_i & =  \sum_{\bm{\kappa} \in \Omega_i} \phi(\bm{\kappa}) +  \sum_{\substack{|S|>1, i \in S}}\sum_{\bm{\kappa} \in \Omega_S}  \frac{1}{|S|} I(\bm{\kappa}) \\
\end{aligned}
\end{small}
\end{equation}
\end{theorem}

Theorem \ref{thm:deepshap} shows that the Deep SHAP method  follows the paradigm of  allocating Taylor interaction effects in Eq.~(\ref{eqn:unifiedattribution2}). 
Specifically, this method allocates the generic independent effect {\small $\psi(i) =  \sum_{\bm{\kappa}\in \Omega_i} \phi(\bm{\kappa})$} of the variable $i$ to the attribution $a_i$. 
Furthermore, this method allocates each Taylor interaction effect {\small $I(\bm{\kappa})$} ({\small $\bm{\kappa} \in \Omega_S, i \in S$}), which involves the variable $i$, to the attribution $a_i$.
The effect {\small $I(\bm{\kappa})$}  is uniformly allocated to all the  {\small $s = |S|$} variables involved in the interaction, \textit{i.e.}, each input variable receives {\small $\frac{1}{|S|}I(\bm{\kappa})$}.
\\

\noindent
\textbf{DeepLIFT RevealCancel.} 
DeepLIFT RevealCancel \cite{shrikumar2017learning} is a typical back-propagation attribution method, 
which modifies the recursive back-propagation rule of the DeepLIFT Rescale method as follows. 
\begin{equation}\label{eqn:deeplift2}
\begin{small}
\begin{aligned}
 a_{i \leftarrow j}^{(l)}&  = 
\left\{
\begin{array}{rcl}
\frac{\Delta z_{ij}}{\sum_{i'\in N^+} \Delta z_{i'j}} \cdot \frac{\Delta y^+}{\Delta y^+ + \Delta y^-} \cdot a_j^{(l)}, & i \in N^+ \\[2mm]
\frac{\Delta z_{ij}}{\sum_{i'\in N^-} \Delta z_{i'j}} \cdot \frac{\Delta y^-}{\Delta y^+ + \Delta y^-} \cdot a_j^{(l)}, &  i \in N^- 
\end{array}\right. 
\\[1mm] 
\end{aligned}
\end{small}
\end{equation}
where {\small $z_{ij} = W_{ij}x_i^{(l-1)}$}, {\small $\tilde z_{ij} = W_{ij} \tilde x_i^{(l-1)}$}, {\small $ \Delta z_{ij} = z_{ij} - \tilde z_{ij}$}. 
Accordingly, {\small$N^+ = \{i|\Delta z_{ij} > 0\}$} and {\small $N^- = \{i|\Delta z_{ij} \leq 0\}$}.
\begin{equation}
\begin{small}
\begin{aligned}
\Delta y^+ &=1/2 \ (\sigma(\tilde{z} +\Delta z^+) - \sigma(\tilde{z})) \\
&+ 1/2 \ (\sigma(\tilde{z} + \Delta z^+ + \Delta z^-) - \sigma(\tilde{z} +\Delta z^-) \\
\Delta y^- &= 1/2 \  (\sigma(\tilde{z} + \Delta z^-) - \sigma(\tilde{z})) \\
&+ 1/2 \  (\sigma(\tilde{z} + \Delta z^+ + \Delta z^-) - \sigma(\tilde{z} + \Delta z^+) \\
s.t. \quad \tilde z & = \sum\nolimits_{i \in N} \tilde z_{ij} + s_j,  \Delta z^+  = \sum\nolimits_{i \in N^+} \Delta z_{ij}, \\
 \Delta z^- &= \sum\nolimits_{i \in N^-} \Delta z_{ij}. 
\end{aligned}
\end{small}
\end{equation}
Both the DeepLIFT Rescale method and the DeepLIFT RevealCancel method use {\small $\Delta z_{ij}$} to represent the contribution of {\small $x_i^{(l-1)}$} on {\small $x_j^{(l)}$}. 
However, Unlike DeepLIFT Rescale, DeepLIFT RevealCancel divides all contributions {\small $\Delta z_{ij}$} into the group {\small $N^+$} with positive contributions and the group {\small $N^-$} with negative contributions.
Then, DeepLIFT RevealCancel computes the attribution in each group separately.
Here, {\small $\Delta y^+/(\Delta y^+ + \Delta y^-)$} and {\small $\Delta y^-/(\Delta y^+ + \Delta y^-)$} denote the weights for two groups, respectively.
Then, the attribution {\small $a_i^{(l-1)}$} is computed as {\small $a_i^{(l-1)} = \sum\nolimits_j a_{i \leftarrow j}^{(l)}$}.
\vspace{2pt}

\noindent
\begin{theorem}\label{thm:DeepLIFT RevealCancel}
\textit{(Proof in Appendix B) Let us consider the $j$-th feature dimension in the $l$-th layer as the target output and consider features in the $(l-1)$-th layer as input variables (i.e., $y = x_j^{(l)}$ and $\bm{x} = \bm{x}^{(l-1)}$), to analyze the layer-wise propagation of attributions. 
Then, in the DeepLIFT RevealCancel method,  the attribution of the input variable {\small $i \in N^+$} can be reformulated as follows.}
\begin{equation}\label{eqn:DeepLIFT RevealCancel1}
\begin{small}
\begin{aligned}
a_i & =  \sum_{\bm{\kappa}\in \Omega_i} \phi(\bm{\kappa}) +  \sum_{\substack{S \subseteq N^+, i \in S}} \ \sum_{\bm{\kappa} \in \Omega_S}  c_i I(\bm{\kappa}) \\
& +  \sum\nolimits_{S \cap N^+ \neq \emptyset, S \cap N^- \neq \emptyset, i \in S}\sum_{\bm{\kappa} \in \Omega_S} \frac{1}{2}  c_i I(\bm{\kappa})] 
\end{aligned}
\end{small}
\end{equation}
\textit{where $c_i = \frac{\kappa_i}{\sum_{i' \in N^+} \kappa_{i'}}$. 
Besides, the attribution of the input variable {\small $i \in N^-$} can be reformulated as}
\begin{equation}\label{eqn:DeepLIFT RevealCancel2}
\begin{small}
\begin{aligned}
a_i & =  \sum_{\bm{\kappa}\in \Omega_i} \phi(\bm{\kappa}) +  \sum_{\substack{S \subseteq N^-, i \in S}} \ \sum_{\bm{\kappa} \in \Omega_S}   \tilde c_i I(\bm{\kappa}) \\
& + \sum\nolimits_{S \cap N^+ \neq \emptyset, S \cap N^- \neq \emptyset, i \in S}\sum_{\bm{\kappa} \in \Omega_S} \frac{1}{2} \tilde c_i I(\bm{\kappa})] 
\end{aligned}
\end{small}
\end{equation}
\textit{where $\tilde c_i = \frac{\kappa_i}{\sum_{i' \in N^-} \kappa_{i'}}$.}
\end{theorem}
\vspace{4pt}

Theorem \ref{thm:DeepLIFT RevealCancel} shows that the DeepLIFT RevealCancel method  follows the paradigm of  allocating Taylor interaction effects in Eq.~(\ref{eqn:unifiedattribution2}). 
As Eq.~(\ref{eqn:DeepLIFT RevealCancel1}) shows, for the variable {\small $i \in N^+$},  this method allocates the variable $i$'s  generic independent effect {\small $\psi(i) = \sum_{\bm{\kappa} \in \Omega_i}\phi(\bm{\kappa})$} to the attribution $a_i$. 
Besides, this method allocates part of the Taylor interaction effect {\small $I(\bm{\kappa})$} ({\small $\bm{\kappa} \in \Omega_S,$ $S \subseteq N^+, i \in S$}) between variables in {\small $S \subseteq N^+$}, which involves the variable $i$, to the attribution $a_i$. 
Moreover, this method allocates a different ratio of the Taylor interaction effect {\small $I(\bm{\kappa})$} ({\small $S \cap N^+ \neq \emptyset,$ $S \cap N^- \neq \emptyset, i \in S$}) between variables in {\small $N^+$} and variables in {\small $N^-$}, to the attribution $a_i$. 
\\

\subsection{Experimental verification}
In this section, we conduct experiments to check the correctness of Theorems \ref{thm:gradinput}-\ref{thm:DeepLIFT RevealCancel},  \textit{i.e.}, whether the reformulated attributions really reflect true attributions estimated by different attribution methods.

Let us use a specific attribution method to explain the inference of a DNN on a given input sample $\bm{x}$.
We use the following metric to measure the average fitting error between the theoretically derived attribution values  $\bm{a}(\bm{x}) \in \mathbb{R}^n$ and the true attribution values $\bm{a}^{*}(\bm{x})\in \mathbb{R}^n$  estimated in real applications.
\begin{equation}\label{eqn:fittingerror}
\begin{small}
\begin{aligned}
E =\mathbb{E}_{\bm{x}} \frac{\Vert\bm{a}(\bm{x}) - \bm{a}^{*}(\bm{x})\Vert_2}{\Vert \bm{a}^{*}(\bm{x}) \Vert_2} \times 100 \%,
\end{aligned}
\end{small}
\end{equation}
where $\bm{a}^*(\bm{x}) \in \mathbb{R}^n$ denotes the true attribution values estimated by the attribution method, 
and $\bm{a}(\bm{x}) \in \mathbb{R}^n$ represents the reformulated attribution values of the attribution method. 
For example, according to Theorem \ref{thm:occ1}, the $i$-th dimension of the reformulated attribution values  $\bm{a}(\bm{x})$ in the Occlusion-1 method is computed as 
$a_i(\bm{x}) = \sum\nolimits_{\bm{\kappa} \in \Omega_i} \phi(\bm{\kappa}) +  \sum\nolimits_{|S| > 1, i \in S} \sum\nolimits_{\bm{\kappa} \in \Omega_S} I(\bm{\kappa})$. 

\begin{table}[t]\small
\centering
\renewcommand\arraystretch{1.25}
\begin{tabular}{l|c|c} \hline
Methods & \makecell[c]{Polynomial} & \makecell[c]{Sigmoid-MLP}   \\  \hline
Grad$\times$Input & 0  & 0  \\ \hline
Occlusion-1 & 0 &  2.46\%   \\ \hline
Occlusion-2$\times$2 & 0 & 2.36\%  \\ \hline
Prediction Difference & 0    & 2.69\%     \\ \hline
Integrated Grads & 0.12\%  & 0.82\% \\ \hline
Expected Grads & 0.16\%  &     0.90\% \\ \hline
Shapley value &  0  &  1.18\%  \\ \hline
\end{tabular}  
\caption{Average fitting errors between the reformulated attribution values and actual attribution values computed by different methods. }
\label{Differences between formula and reformula}
\end{table}

Note that it is impossible for us to enumerate all Taylor interaction effects $I(\bm{\kappa})$ in Eq. (\ref{eqn:preliminary}).
Thus, it is difficult for us to precisely compute the theoretically derived attribution values  $\bm{a}(\bm{x})$. 
Instead, given a DNN and an input sample $\bm{x}$, we compute only the first-order and the second-order Taylor interaction effects $I(\bm{\kappa})$, which subject to $\bm{\kappa} \in O = \{\bm{\kappa} \in \mathbb{N}^n| \kappa_1 + \dots + \kappa_n = 1 \textrm{ or } 2 \}$.
Then, we estimate $\bm{a}(\bm{x})$  by ignoring all Taylor interaction effects of greater than two orders for implementation. 
For example, in the Occlusion-1 method, the $i$-th dimension of the theoretically derived attribution values $a(\bm{x})$ is approximated by $a_i(\bm{x}) \approx \sum_{\bm{\kappa} \in \Omega_i, \bm{\kappa} \in O} \phi(\bm{\kappa}) +  \sum_{|S| > 1, i \in S} \sum_{\bm{\kappa} \in \Omega_S, \bm{\kappa} \in O} I(\bm{\kappa})$.
In addition, we do not conduct the Taylor expansion at the input sample $\bm{x}$ as in the preliminary version of this paper \cite{deng2021unified}. 
Instead, we expand the neural network at a pre-defined baseline point $\bm{b}$, which is more standard than the previous version. 
The baseline point  $\bm{b}$ is generated by adding a random Gaussian perturbation on the input sample.
Furthermore, we notice that the gating states of ReLU networks do not have continuous gradients, which may introduce a large measurement error. 
Therefore, we train only DNNs with sigmoid activation functions, rather than DNNs with ReLU activation functions, for testing.

We test  $\bm{a}(\bm{x})$  and $\bm{a}^{*}(\bm{x})$  on two types of models. 
The first type of model is the second-order polynomial model, \textit{i.e.}, $f(\bm{x}) = \sum_{i \in N} c_i x_i + \sum_{i \in N} \sum_{j \in N} c_{ij} x_ix_j$, where $c_i$ and $c_{ij}$ denote model weights. 
We term this type of model   \textbf{\textit{Polynomial model}}. 
The second type of model is the three-layer multi-layer perceptron network, which applies the sigmoid activation function. 
We term this type of  model  \textbf{\textit{Sigmoid-MLP}}. 
We train these models on the MNIST dataset \cite{deng2012mnist} and compute the average fitting errors $E$ according to Eq. (\ref{eqn:fittingerror}). 

\begin{figure*}[t]\centering
\includegraphics[width=0.8\textwidth]{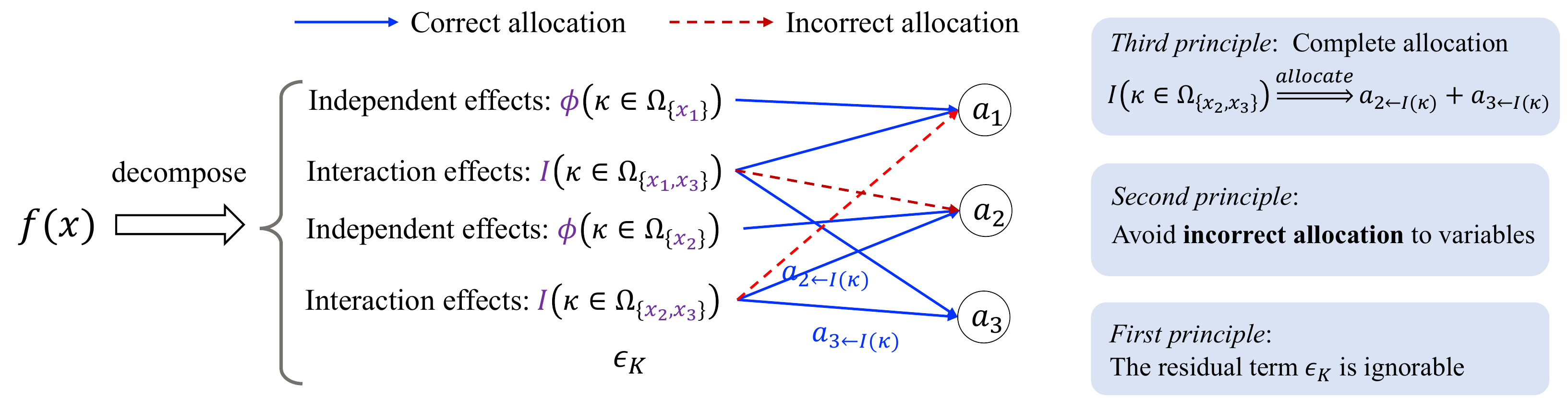}
\caption{An illustration of three principles of faithfulness.}
\label{principle of faithfulness}
\end{figure*}

Moreover, we evaluate the fitting errors $E$  of seven attribution methods, including the Gradient$\times$Input, Occlusion-1, Occlusion-patch, Prediction Difference, Integrated Gradients, Expected Gradients, and Shapley value methods. 
We do not test back-propagation attribution methods because theoretically derived attribution values for these methods mainly explain layer-wise propagation rules. 

Table \ref{Differences between formula and reformula} lists the average fitting errors $E$  of the seven attribution methods, which are evaluated on the above two types of models. 
We find that on different types of models, fitting errors $E$ of different attribution methods are all close to $0$. 
Theoretically, there should not be any fitting errors when we test on Polynomial models, but tiny errors of the \textit{Integrated Gradients} and \textit{Expected Gradients} methods come from the unavoidable error of the integral computation.
The above result indicates that for these attribution methods, the theoretically derived attribution value  $\bm{a}(\bm{x})$ well fits the actual attribution value $\bm{a}^{*}(\bm{x})$ computed by these methods in real applications.

\section{Evaluating attribution methods}
\label{sec:principles}
In the last section, we have proven that various attribution scores estimated by fourteen attribution methods can all be reformulated into the \textbf{\textit{unified paradigm}} of allocating Taylor independent effects $\phi(\bm{\kappa})$ and Taylor interaction effects $I(\bm{\kappa})$ in Eq.~(\ref{eqn:unifiedattribution2}).

The unified paradigm shared by different attribution methods enables us to fairly evaluate and compare different attribution methods in a theoretical manner. 
Therefore, in this subsection, we propose three principles to evaluate the faithfulness of fourteen attribution methods. 


\begin{table*}[t]
\renewcommand\arraystretch{1.1}
\centering
\resizebox{1.0\linewidth}{!}{
\begin{tabular}{l | c | c | c | l | c | c | c}
\toprule
\makecell[l]{Attribution  \\ methods} & \makecell[c]{low approxi- \\ mation error}& \makecell[c]{no unrelated\\ allocation}  & \makecell[c]{complete \\ allocation} & \makecell[l]{Attribution  \\ methods}& \makecell[c]{low approxi- \\ mation error}& \makecell[c]{no unrelated\\ allocation}   & \makecell[c]{complete \\ allocation}   \\  \hline
Grad$\times$Input & $\times$  &  \checkmark &  \checkmark  & Shapley value  & \checkmark & \checkmark & \checkmark   \\ \hline
Occ-1 &  \checkmark & \checkmark  & $\times$ & LRP-$\epsilon$ & $\times$ & \checkmark &  \checkmark \\ \hline
Occ-patch  &  \checkmark & $\times$ & $\times$ & LRP-$\alpha\beta$ & \checkmark  & $\times$  & \checkmark  \\  \hline
Prediction Diff &  \checkmark & \checkmark &  $\times$  & Deep Taylor & \checkmark  & $\times$  & \checkmark   \\ \hline
Grad-CAM  & $\times$ & \checkmark & \checkmark  & DeepLIFT Rescale  & \checkmark & \checkmark & \checkmark   \\ \hline
Integrated Grads & \checkmark & \checkmark & \checkmark   & DeepShap & \checkmark & \checkmark & \checkmark  \\ \hline
Expected Grads & \checkmark & \checkmark & \checkmark & DeepLIFT Reveal  & \checkmark & \checkmark & \checkmark \\ \hline 
\bottomrule
\end{tabular}
}
\caption{A summary of principles followed by each attribution method.}
\label{tab:evaluation}
\end{table*}

\subsection{Principles for a faithful attribution method}
\label{sec:proposeprinciple}
As shown in Figure \ref{principle of faithfulness}, the unified paradigm shared by different attribution methods indicates that each attribution method can all be considered as a flowchart, which firstly represents the DNN as a Taylor expansion model, and then accordingly re-allocates the Taylor independent effects {\small $\phi(\bm{\kappa})$} and the Taylor interaction effects {\small $I(\bm{\kappa})$} to compute the attribution score $a_i$.  

To this end, we find that the faithfulness of an attribution method depends on two key factors:\\
(i) whether the residual term $\epsilon_K$ in the Taylor expansion of the DNN is small enough; \\
(ii) whether the Taylor independent effect and the Taylor interaction effect are allocated to input variables in a reasonable manner.\\
Accordingly, we propose three principles that faithful attributions are supposed to follow. 

\textbf{First principle: low approximation error.}
The unified paradigm of attribution methods proves that each attribution method actually explains different Taylor expansion terms of the DNN, including Taylor independent effects and Taylor interaction effects.
Therefore, faithful attributions are expected to cover almost all Taylor expansion terms of the DNN, 
and leave an ignorable residual term $\epsilon_K$ not been explained.

\textbf{Second principle: avoiding allocation to unrelated variables.}
The unified paradigm of attribution methods shows that each attribution method actually re-allocates different Taylor independent effects and different Taylor interaction effects to each input variable, so as to compute attribution scores. Then, 

\textit{(i) Each Taylor independent effect of the variable $i$, {\small $\phi(\bm{\kappa})$} subject to {\small $\bm{\kappa} \in \Omega_i$}, is supposed to be allocated only to the variable $i$}.
More specifically, we can decompose a term {\small $a_{i \leftarrow \phi(\bm{\kappa})} = w_{i,\bm{\kappa}}\phi(\bm{\kappa})$} from the attribution $a_i$ according to Eq.~(\ref{eqn:unifiedattribution2}), to represent the numerical effect assigned from the Taylor independent effect {\small $\phi(\bm{\kappa})$} to the variable $i$. 
Then, we should avoid allocating the independent effect {\small $\phi(\bm{\kappa})$} \textit{s.t.} {\small $\bm{\kappa} \in \Omega_i$} to other unrelated variables $j \neq i$.
\begin{equation}
\begin{small}
\begin{aligned}
\forall \bm{\kappa} \in \Omega_i, \ \forall j \neq i,  \ \  a_{j \leftarrow \phi(\bm{\kappa})} = 0.
\end{aligned}
\end{small}
\end{equation}

(ii) \textit{Each Taylor interaction effect between variables in {\small $S$},  {\small $I(\bm{\kappa})$} subject to {\small $\bm{\kappa} \in \Omega_S$}, is supposed to exclusively be allocated to variables in {\small $S$}, without being allocated to any other unrelated variables {\small $j \not\in S$}}. 
\begin{equation}
\begin{small}
\begin{aligned}
\forall \bm{\kappa} \in \Omega_S, \  \forall j \not\in S, \  \ a_{j \leftarrow I(\bm{\kappa})} = 0.
\end{aligned}
\end{small}
\end{equation}
where {\small $a_{j \leftarrow I(\bm{\kappa})} = w_{j,\bm{\kappa}}I(\bm{\kappa})$} in Eq.~(\ref{eqn:unifiedattribution2}) denotes the numerical effect 
assigned from the Taylor interaction effect  {\small $I(\bm{\kappa})$}  to the input variable $j$.

\textbf{Third principle: complete allocation.}
\textit{Each Taylor independent effect {\small $\phi(\bm{\kappa})$} is supposed to \textbf{completely} be allocated to different input variables.}
In other words, if we accumulate all numerical effects  allocated from {\small $\phi(\bm{\kappa})$} to different variables, we can obtain the exact value of {\small $\phi(\bm{\kappa})$}.
\begin{equation}
\begin{small}
\begin{aligned}
\forall i \in N, \forall \bm{\kappa} \in \Omega_i, \ \sum\nolimits_{j \in N} a_{j \leftarrow \phi(\bm{\kappa})} =   \phi(\bm{\kappa}). 
\end{aligned}
\end{small}
\end{equation}
Similarly, \textit{each Taylor interaction effect {\small $I(\bm{\kappa})$} is supposed to \textbf{completely} be allocated to different input variables.}
\begin{equation}\label{eqn:completeness}
\begin{small}
\begin{aligned}
\forall S \subseteq N, \  \forall \bm{\kappa} \in \Omega_S, \  \sum\nolimits_{i \in N} a_{i \leftarrow I(\bm{\kappa})} =  I(\bm{\kappa}). 
\end{aligned}
\end{small}
\end{equation}

\subsection{Evaluating attribution methods}
In this subsection, we use the proposed principles to evaluate the aforementioned fourteen attribution methods. 
\textit{Note that the proposed principles only provide a new perspective to evaluate the faithfulness of attribution methods. 
This does not imply that attribution methods satisfying these principles are ideal attributions. 
There are many other perspectives to evaluate attribution methods \cite{fong2017interpretable,adebayo2018sanity,Marco2018Towards,yeh2019fidelity,hooker2019benchmark}.}
Thus, the main contribution of this work is to unify fourteen different attribution methods into the same mathematical system, thereby enabling us to evaluate different attribution methods in the same theoretical system.
\vspace{4pt}

\noindent
$\bullet$ Gradient $\times$ Input, LRP-$\epsilon$, and Grad-CAM do not satisfy the \textbf{\textit{low-approximation-error} principle}. 
According to Theorems \ref{thm:gradinput},  \ref{thm:gradcam}, and \ref{thm:lrp}, 
these methods consider only the first-order Taylor expansion terms of the DNN to compute attributions, and ignore expansion terms of higher orders. 
\vspace{4pt}

\noindent
$\bullet$ Deep Taylor, LRP-$\alpha\beta$, and Occlusion-patch all violate the principle of \textbf{\textit{avoiding allocation to unrelated variables}}. 
According to Theorems \ref{thm:LRPab} and \ref{thm:DeepTaylor}, Deep Taylor and LRP-$\alpha\beta$  mistakenly allocate the Taylor interaction effects between variables in {\small $N^-$}, {\small $I(\bm{\kappa}) (\bm{\kappa} \in \Omega_S, S \subseteq N^-)$}, to variables {\small $i \in N^+$}  that are unrelated to this interaction $I(\bm{\kappa})$.
In addition, according to Theorem \ref{thm:occp}, the Occlusion-patch method may mistakenly allocate the Taylor interaction effect, which does not involve the variable $i$, to the unrelated variable $i$. 
\vspace{4pt}

\noindent
$\bullet$ Occlusion-1, Occlusion-patch, and Prediction Difference all violate the \textbf{\textit{complete-allocation} principle}.
Specifically, according to Theorems \ref{thm:occ1}, \ref{thm:occp}, and \ref{thm:Prediction Difference}, the three methods \textbf{repeatedly} allocate all numerical values of each Taylor interaction effect {\small $I(\bm{\kappa}) (\bm{\kappa} \in \Omega_S)$} between variables in {\small $S$} to each variable in {\small $S$}, 
\textit{i.e.}, {\small $\forall i \in S, \ a_{i \leftarrow I(\bm{\kappa})} = I(\bm{\kappa})$}.
In this way, the sum of numerical effects allocated from {\small $I\bm{\kappa})$} to different variables is given as {\small $\sum_{i \in S} a_{i \leftarrow I(\bm{\kappa})} = |S| \cdot I(\bm{\kappa})$}, which is {\small $|S|$} times  greater than the Taylor interaction effect {\small $I(\bm{\kappa})$}. 
This violates the \textit{complete-allocation} principle in Eq.~(\ref{eqn:completeness}). 
\vspace{4pt}

\noindent
$\bullet$ According to Theorems \ref{thm:ig}, \ref{thm:Expected Gradients}, \ref{thm:shapley}, \ref{thm:deepliftrescale}, and \ref{thm:DeepLIFT RevealCancel}, 
the Integrated Gradients,  Expected Gradients, Shapley value, Deep Shap,  DeepLIFT Rescale, and DeepLIFT RevealCancel methods satisfy all principles. 
The main difference among the four methods is that each method allocates a different numerical effect {\small $a_{i \leftarrow I(\bm{\kappa})}$}  from the interaction effect {\small $I(\bm{\kappa}) (\bm{\kappa} \in \Omega_S)$} to each variable $i \in $ {\small $S$}.
\vspace{4pt}

As a toy example, let us consider the allocation of the Taylor interaction effect $I(\bm{\kappa} = [2,1,2,0,0,\dots]) = x_1^2x_2x_3^2$ in a polynomial function, where $I(\bm{\kappa})$ quantifies the interaction effect between variables in {\small $S = \{x_1, x_2, x_3\}$}. 
Both the Integrated Gradients and the Expected Gradients methods allocate a numerical effect {\small $a_{i \leftarrow I(\bm{\kappa})}$} from {\small $I(\bm{\kappa})$} to each $i$-th input variable, where the allocated numerical effect is proportional to the relative degree of this variable, \textit{i.e.}, {\small $a_{i \leftarrow I(\bm{\kappa})} \propto {\kappa_i}/{\sum_{i'} \kappa_{i'}}$}. 
In this way, the two methods allocate $a_{1 \leftarrow I(\bm{\kappa})} = 2/5I(\bm{\kappa})$, $a_{2 \leftarrow I(\bm{\kappa})} = 1/5 I(\bm{\kappa})$, $a_{3 \leftarrow I(\bm{\kappa})} = 2/5 I(\bm{\kappa})$ to the variables {\small $x_1, x_2, x_3$}, respectively.
Then, different attributions generated by these methods are caused by the fact that these methods use different baseline points.
In addition, the Shapley value method \textbf{uniformly} allocates the Taylor interaction effect {\small $I(\bm{\kappa})$} to each variable $i \in $ {\small $S$}, \textit{i.e.}, {\small $a_{1 \leftarrow I(\bm{\kappa})} = a_{2 \leftarrow I(\bm{\kappa})}  = a_{3 \leftarrow I(\bm{\kappa})} = \frac{1}{3}I(\bm{\kappa})$}.

The suitability of an attribution method depends on the specific task for the DNN. 
For example, in the image classification task, 
the attribution of each pixel generated by Integrated Gradients may be biased.
This is because according to Theorems \ref{thm:ig}, the Integrated Gradients method usually allocates a greater Taylor interaction effect to the pixel with a more significant pixel value (\textit{e.g.}, white pixels).
In comparison, the Shapley value, which uniformly allocates Taylor interaction effects to different pixels involved in the interaction, may be more suitable for the image classification task.

\subsection{Connections between the three principles and previous evaluation metrics}
We find that some attribution methods, which satisfy the proposed three principles, are usually also  top-ranked attribution methods evaluated by previous evaluation metrics \cite{Karl2020Restricting,yeh2019fidelity}.  
Specifically, we investigate the following two evaluation metrics for attribution scores.

\begin{figure}[t]\centering  
\includegraphics[width = 0.48 \textwidth]{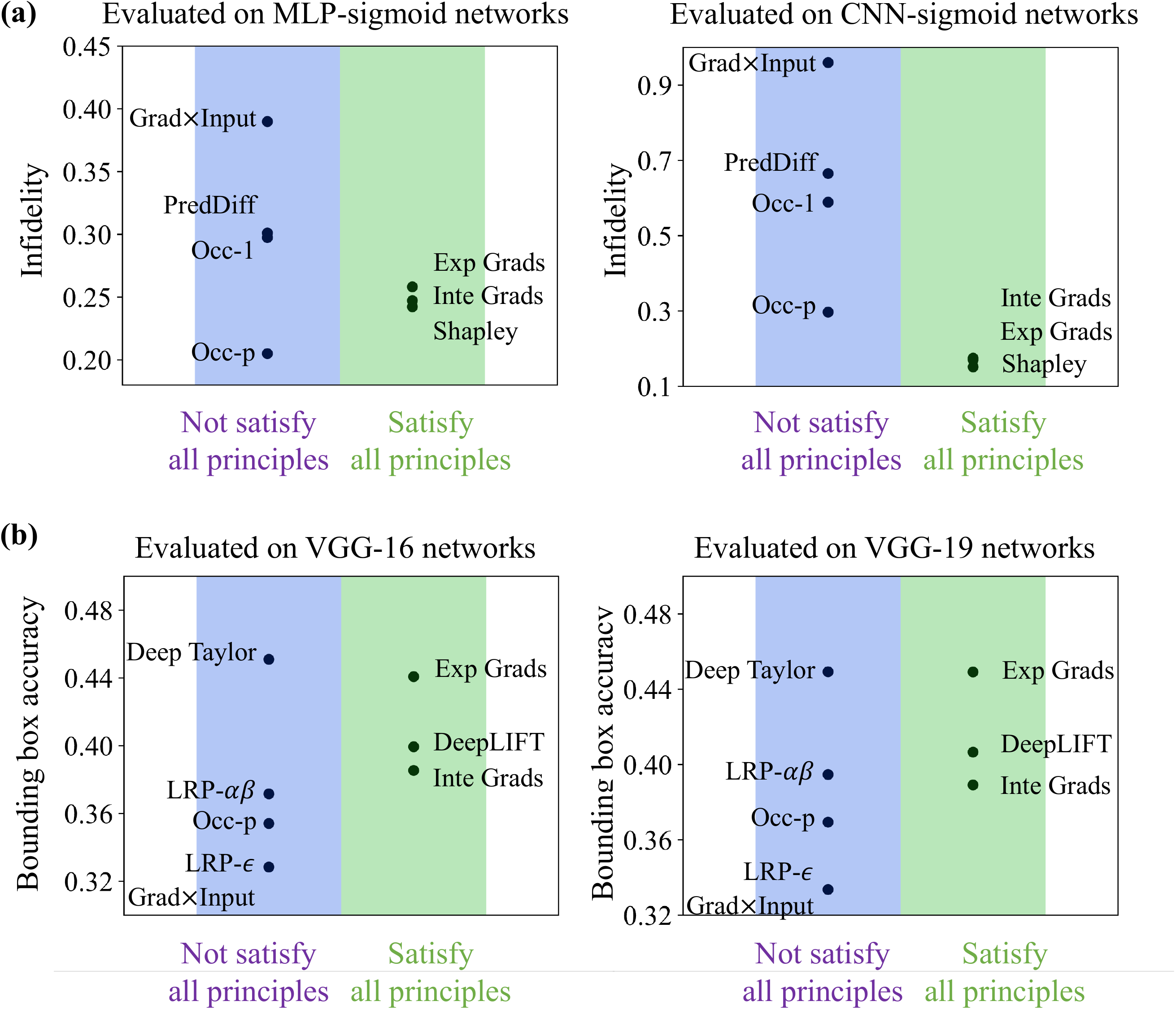}
\caption{(a) Infidelity of different attribution methods evaluated on  MLP-sigmoid networks and CNN-sigmoid networks. 
(b) Bounding box accuracy of different attribution methods evaluated on VGG-16 networks and VGG-19 networks. 
}
\label{IFtrend and bbox}
\end{figure}

$\bullet$ \textbf{Infidelity metric \cite{yeh2019fidelity}.} For a specific attribution method, the infidelity metric  is proposed to evaluate whether attribution scores generated by the attribution method can well predict the output changes when we add certain perturbations to the input.
Specifically,  given a DNN $f$ and an input sample $\bm{x} \in \mathbb{R}^n$, the attribution method estimates attribution scores $\bm{a} \in \mathbb{R}^n$ of different input variables. Then,  the infidelity metric is defined as 
	\begin{equation}
		\begin{small}
			\begin{aligned}
				\textrm{INFD}(\bm{a}, f, \bm{x}) = \mathbb{E}_{\bm{p}_{\bm{x}}} [\bm{p}_{\bm{x}}^T \bm{a} - (f(\bm{x}) - f(\bm{x} - \bm{p}_{\bm{x}})) ]^2
 			\end{aligned}
		\end{small}
	\end{equation}
where $\bm{p}_{\bm{x}} \in \mathbb{R}^n$ denotes the perturbation added on the sample $\bm{x}$. 
Thus,  the infidelity metric quantifies the average error of using attribution scores to predict the output change $f(\bm{x}) - f(\bm{x} - \bm{p}_{\bm{x}})$ \textit{w.r.t.} input perturbations. 
Low infidelity indicates that the attribution method can well reflect the output change of the DNN \textit{w.r.t.} input perturbations.
In implementation, we adopt the square removal perturbation in  \cite{yeh2019fidelity} for evaluation.
We evaluate the infidelity on a three-layer MLP network with sigmoid activation functions and a three-layer CNN with sigmoid functions, respectively. 
These networks are trained on the MNIST dataset.

Figure \ref{IFtrend and bbox}(a) illustrates the relationship between the proposed three principles and the infidelity metric. 
The $x$-axis denotes the number of principles that a specific attribution method satisfies, and the $y$-axis denotes the corresponding infidelity metric of the attribution method. 
Figure \ref{IFtrend and bbox}(a) 
indicates that attribution methods that satisfy all the three principles usually show a lower infidelity.
However, attribution methods that perform well on the infidelity metrics are not necessarily equivalent to faithful methods that satisfy all the three principles.

$\bullet$ \textbf{Bounding box accuracy metric \cite{Karl2020Restricting}.} For a specific attribution method, the bounding box accuracy metric is proposed to evaluate whether pixels estimated with top-ranked attribution scores can well localize the object of the target category. 
Specifically, given an input image with $n$ pixels $N = \{1, \dots, n\}$, let us assume that the annotated bounding box $B \subseteq N$ of the target object contains $m$ pixels, \textit{i.e.}, $|B| = m$. 
Then, we select a subset $M \subseteq N$ of $m$ pixels with top-ranked attribution scores estimated by the attribution method.
The bounding box accuracy metric is defined as $\frac{|B \cap M|}{|B|}$.
A  high bounding box accuracy indicates that the attribution method can well localize the target object. 
For evaluation, we only use testing images whose bounding box covers less than 33\% pixels of the whole input image, \textit{i.e.}, $m < 33\% n$. 
We evaluate the bounding box accuracies of each attribution method on VGG16 \cite{Simonyan15} and VGG19 \cite{Simonyan15} networks and test images on the ImageNet dataset \cite{deng2009imagenet}.

Figure \ref{IFtrend and bbox}(b) illustrates the relationship between the three principles proposed in Section \ref{sec:proposeprinciple} and  the bounding box accuracy metric. The $x$-axis denotes the number of principles that a specific attribution method satisfies, and the $y$-axis denotes  the corresponding bounding box accuracy metric of the attribution method. 
Figure \ref{IFtrend and bbox}(b) indicates that  attribution methods, which satisfy all the three principles, perform a bit better on average in the localization of target objects.

The experimental results show a relative consistency between the previous evaluation metrics and our proposed principles.
However, the above two metrics and our proposed principles evaluate attribution methods from different perspectives.  
The infidelity metric evaluates the ability of an attribution method to predict output changes under input perturbations.
The bounding box metric measures the consistency between the attribution result and human intuition of localizing the target object, but human intuitions are not necessarily equivalent to the true inference logic of a neural network.
Thus, attribution methods that perform well in terms of infidelity and bounding box accuracy do not always allocated interaction effects in a faithful manner.

\section{Conclusion}
In this study, we propose the Taylor interaction effect as a unified perspective to explain the mechanisms of fourteen attribution methods. 
Specifically, we prove that the attribution score estimated by each method can all be reformulated as a specific re-allocation of the Taylor independent effects and the Taylor interaction effects.
Furthermore, from the unified perspective, we propose three principles for faithful attributions and then use them to evaluate the fourteen attribution methods.

%

\ifCLASSOPTIONcaptionsoff
  \newpage
\fi

\bibliographystyle{IEEE}
\bibliography{PAMI_semantics}

\vspace{-30pt}
\begin{IEEEbiography}[{\includegraphics[width=1in,height=1.5in,clip,keepaspectratio]{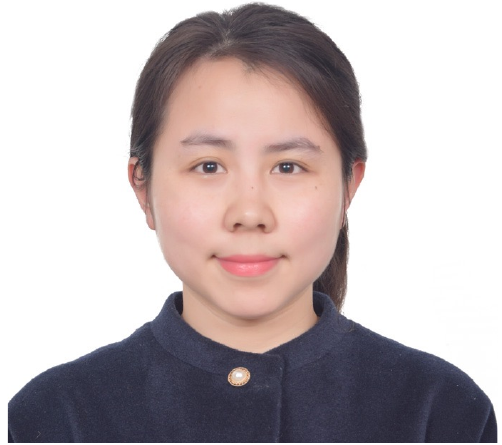}}]{Dr. Huiqi Deng} is currently a postdoctoral researcher at Shanghai Jiao Tong University, China. 
She received her Ph.D. degree in applied mathematics at Sun Yat-sen University, China, in 2021.
Her research interests cover a wide range of explainable machine learning and adversarial robustness. 
\end{IEEEbiography}

\begin{IEEEbiography}[{\includegraphics[width=1in,height=1.4in,clip,keepaspectratio]{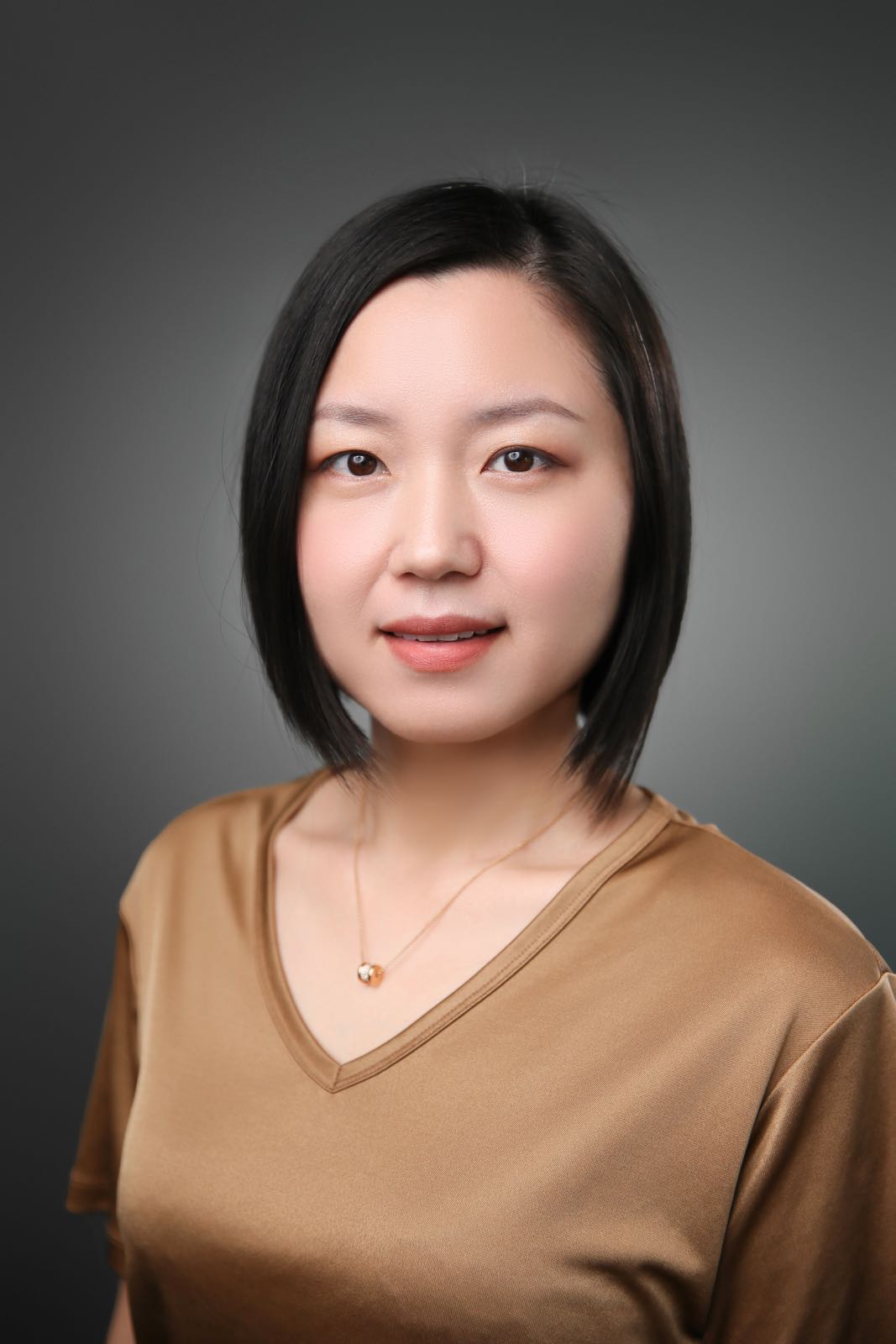}}]{Dr. Na Zou} is an assistant professor and Corrie \& Jim Furber'64 Faculty Fellow in Engineering Technology and Industrial Distribution at Texas A\&M University. Her research focuses on fair and interpretable machine learning, transfer learning, network modeling and inference, supported by NSF and industrial sponsors. The research projects have resulted in publications at prestigious journals such as Technometrics, IISE Transactions and ACM Transactions, including one Best Paper Finalist and one Best Student Paper Finalist at INFORMS QSR section and two featured articles at ISE Magazine. She was the recipient of IEEE Irv Kaufman Award and Texas A\&M Institute of Data Science Career Initiation Fellow.
\end{IEEEbiography}

\begin{IEEEbiography}[{\includegraphics[width=1in,height=1.5in,clip,keepaspectratio]{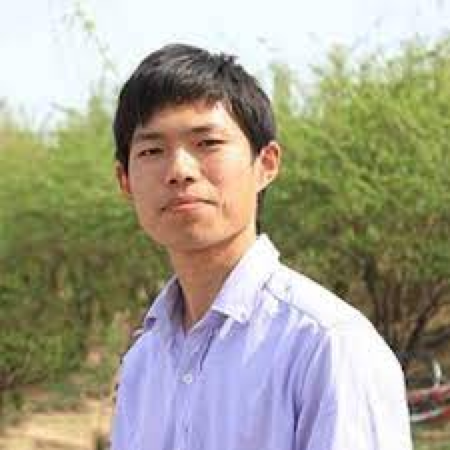}}]{Dr. Mengnan Du}  is an Assistant Professor in the Department of Data Science, New Jersey Institute of Technology (NJIT). He earned his Ph.D. in Computer Science from Texas A\&M University. He has previously worked/interned with Microsoft Research (MSR), Adobe Research, Intel, Baidu Research, Baidu Search Science and JD Explore Academy. His research covers a wide range of trustworthy machine learning topics, such as model explainability, fairness, and robustness. He has had more than 40 papers published in prestigious venues such as NeurIPS, AAAI, KDD, WWW, ICLR, and ICML. He received over 2,300 citations with an H-index of 16.
\end{IEEEbiography}

\begin{IEEEbiography}[{\includegraphics[width=1in,height=1.5in,clip,keepaspectratio]{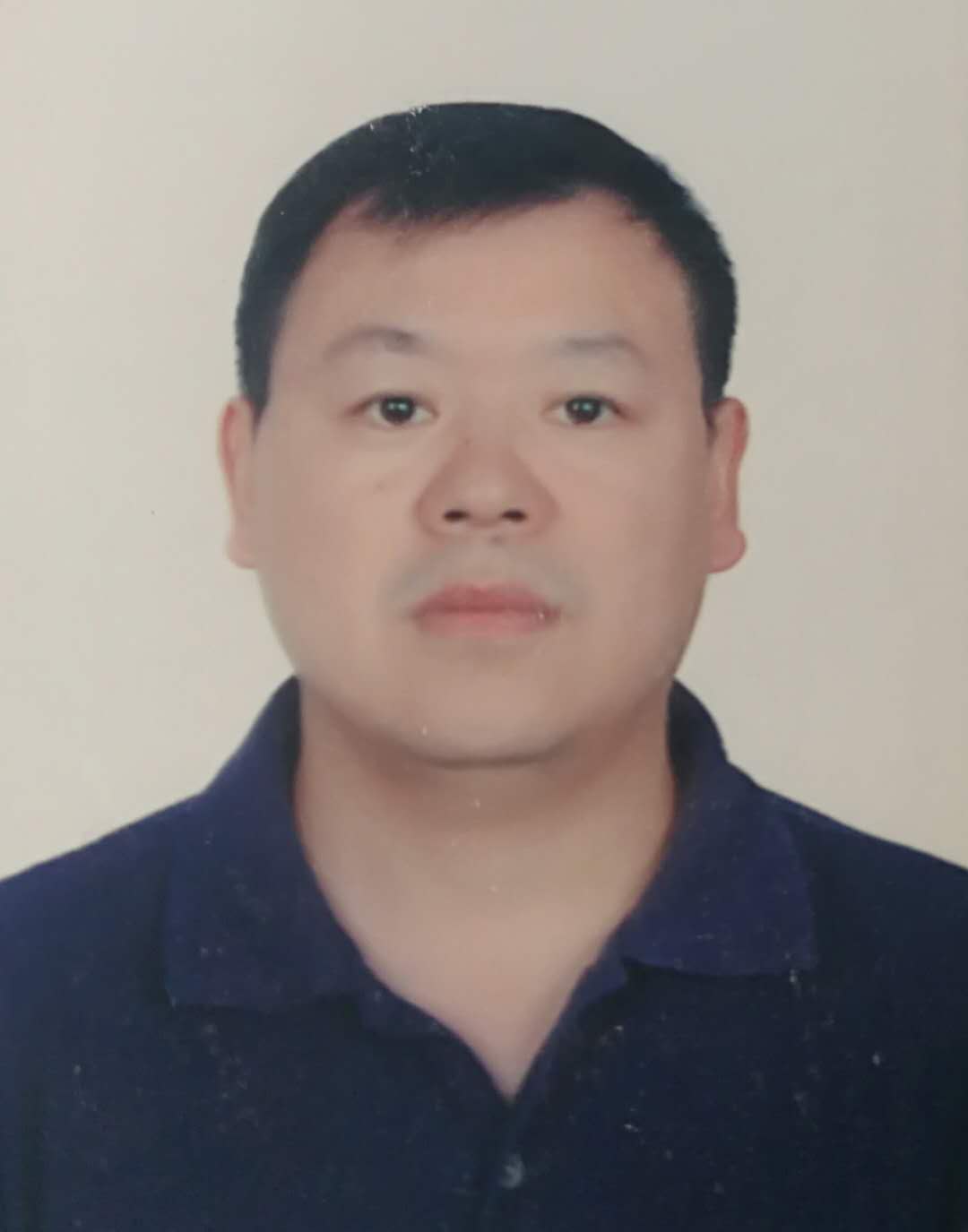}}]{Dr. Guocan Feng}  is a professor at Sun Yat-sen University, China. 
He received his Ph.D. degree in computer science from Hong Kong Baptist University in 1999. 
He was a research fellow in Digital Media Lab in University of Glamorgan and Univ. of Bradford in the UK from 2000 to 2002. His research interests include Digital Image Processing, Pattern Recognition, Computer Vision, Image Retrieval and Indexing in the compressed domain, and manifold learning. Dr. Feng has published over 80 refereed papers at conferences and journals.
\end{IEEEbiography}

\begin{IEEEbiography}[{\includegraphics[width=1in,height=1.5in,clip,keepaspectratio]{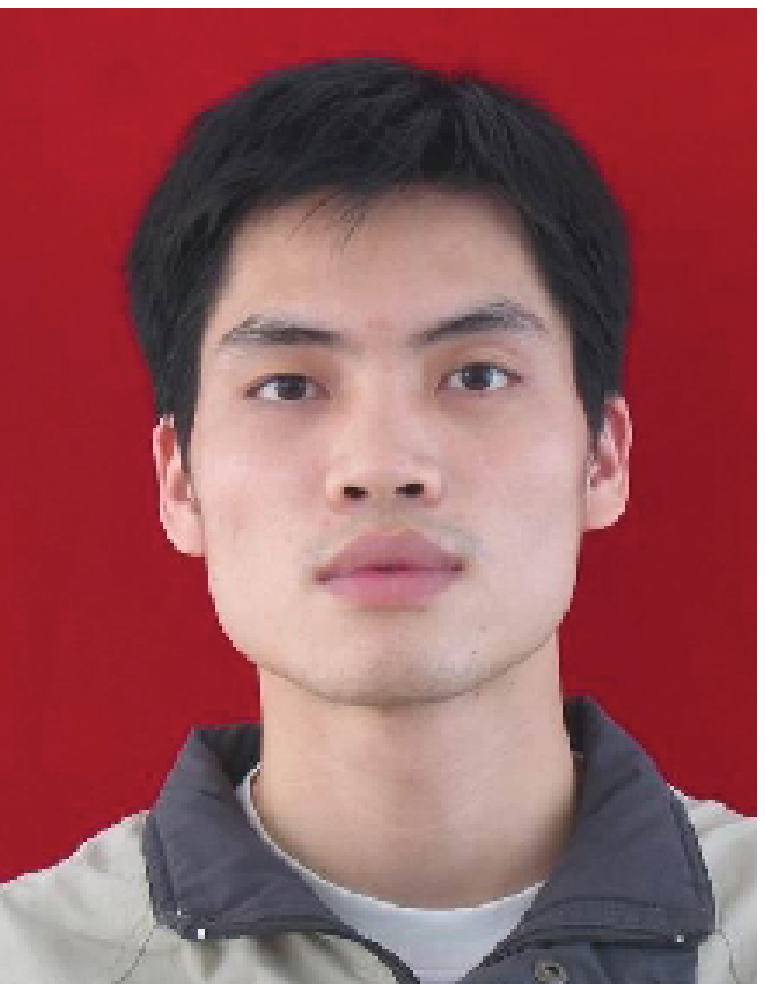}}]{Dr. Weifu Chen}  is an associate professor in Department of Computer Science, Guangzhou Jiaotong University. He received his Ph.D degree in Computing Mathematics at Sun Yat-sen University China in 2012. He was a senior research associate in Department of Electronic Engineering, City University of Hong Kong from 2012 to 2016. From 2016 to 2022, he was an associate research fellow at Sun Yat-sen University. His research interests include statistical pattern recognition and medical image processing.
\end{IEEEbiography}

\begin{IEEEbiography}[{\includegraphics[width=1in,height=1.5in,clip,keepaspectratio]{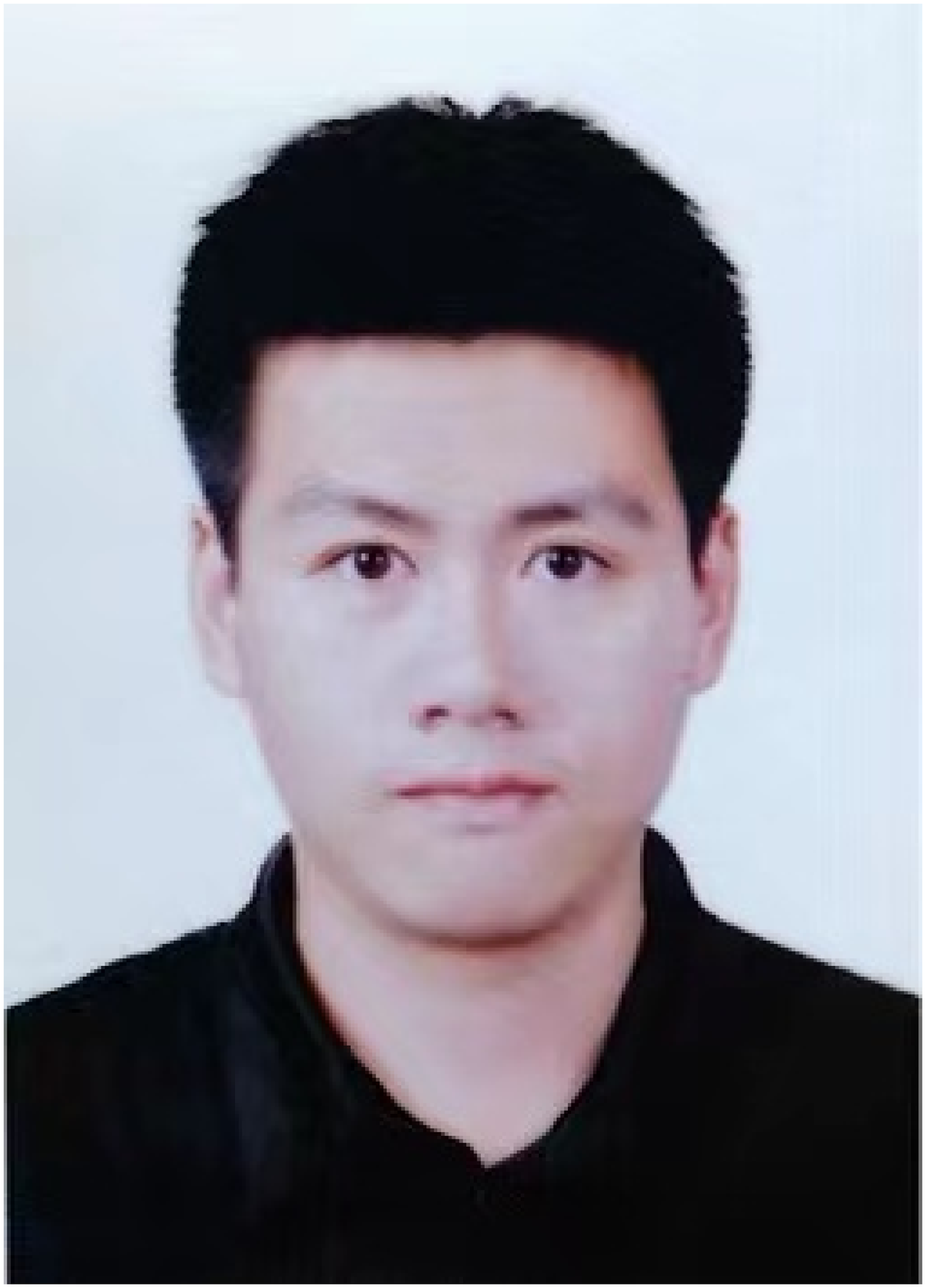}}]{Mr. Zheyang Li} is an algorithm researcher at Hikvision Research Institute. He received the MSc degree in Shanghai JiaoTong University, Shanghai, China, in 2015. His current research interests include perception algorithm, neural network acceleration, explainable AI.
\end{IEEEbiography}

\begin{IEEEbiography}[{\includegraphics[width=1in,height=1.5in,clip,keepaspectratio]{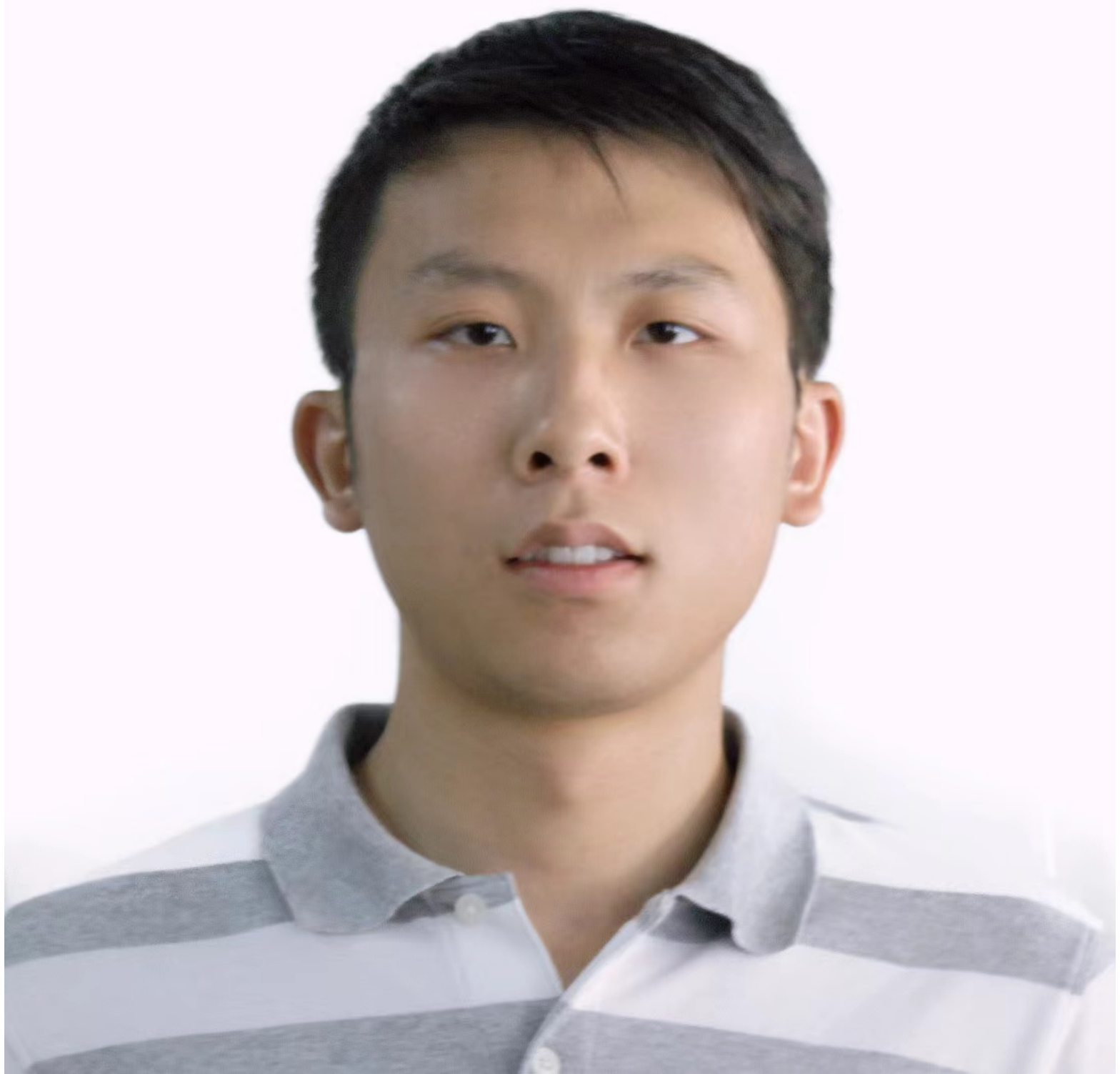}}]{Mr. Ziwei Yang} is an algorithm researcher at Hikvision Research Institute. He received the MSc degree from Tianjin University, China, in 2018. His research interests mainly include neural architecture search, transfer learning and explainable machine learning.
\end{IEEEbiography}

\begin{IEEEbiography}[{\includegraphics[width=1in,height=1.5in,clip,keepaspectratio]{./figures/zhang.pdf}}]{Dr. Quanshi Zhang}
is an associate professor at Shanghai Jiao Tong University, China. He received the Ph.D. degree from the University of Tokyo in 2014. From 2014 to 2018, he was a post-doctoral researcher at the University of California, Los Angeles. His research interests are mainly machine learning and computer vision. In particular, he has made influential research in explainable AI (XAI). He won the ACM China Rising Star Award at ACM TURC 2021. He is the speaker of the tutorials on XAI at IJCAI 2020 and IJCAI 2021. He was the co-chairs of the workshops towards XAI in ICML 2021, AAAI 2019, and CVPR 2019. 
\end{IEEEbiography}
\newpage

\section*{Appendix A}

\textbf{Proof of Proposition 1}

\textit{Proposition 1: The network output {\small $f(\bm{x})$} can be decomposed as the sum of generic independent effects {\small $\psi(i)$} of different input variables $i$ and generic  interaction effects {\small $J(S)$} w.r.t. different subsets {\small $S$} of input variables.}
\begin{equation}
\begin{small}
\begin{aligned}
 f(\bm{x})   &= f(\bm{b})  + \sum\limits_{i \in N} \sum_{\bm{\kappa} \in \Omega_i} \phi(\bm{\kappa})  + \sum\limits_{\substack{S \subseteq N, |S| > 1}} \sum_{\bm{\kappa} \in \Omega_S} I(\bm{\kappa}) \\
 &=  f(\bm{b})  + \sum_{i \in N}  \psi(i)  + \sum_{\substack{S \subseteq N,|S| > 1}} J(S) 
\end{aligned}
\end{small}
\end{equation}

\begin{proof}
According to the Taylor expansion of $f(\bm{x})$ at the baseline point $\bm{b}$, we have
\begin{equation}\label{eqn:taylor expansion}
\begin{small}
\begin{aligned}
 f(\bm{x})   &= f(\bm{b})  + \sum\nolimits_{k = 1}^{\infty} \sum\nolimits_{\bm{\kappa} \in O_k} I(\bm{\kappa}) \\
 & =  f(\bm{b})  +  \sum\nolimits_{\bm{\kappa} \in Q} I(\bm{\kappa}),   \ \ (Q = \cup_{k=1}^{\infty} O_k) 
\end{aligned}
\end{small}
\end{equation}
where each $I(\bm{\kappa})$ represents a Taylor expansion term (Taylor interaction effect). 
Here, $\bm{\kappa} =  [\kappa_1, \cdots,  \kappa_n] \in \mathbb{N}^n$ denotes the degree vector of $I(\bm{\kappa})$, where $\kappa_i \in \mathbb{N}$ denotes the non-negative integral degree of the input variable $i$. 
In addition, the set of degree vectors $O_k = \{\bm{\kappa}|\bm{\kappa} \in \mathbb{N}^n, \kappa_1 + \dots + \kappa_n = k\}$ represents all expansion terms of the $k$-th order. 

For a specific degree vector $\bm{\kappa}$, let us define $S_{\bm{\kappa}} \overset{\textrm{def}}{=} \{i | \kappa_i > 0, \kappa_i \in \mathbb{N}\}$ as its receptive field, \textit{i.e.}, the set of variables with positive integral degrees. 
In this way, we can further categorize all degree vectors $ \bm{\kappa} \in P$ into different receptive fields as follows.
\begin{equation} 
\begin{small}
\begin{aligned}
Q & = \left(\cup_{i \in N} \Omega_i\right) \cup \left(\cup_{S \subseteq N, |S|>1} \Omega_S\right) \\
s.t. \quad \Omega_i &= \{ \bm{\kappa}| S_{\bm{\kappa}} = i \} \\
      	       \Omega_S &= \{ \bm{\kappa}| S_{\bm{\kappa}} = S \} 
\end{aligned}
\end{small}
\end{equation}
Therefore, Eq. (\ref{eqn:taylor expansion}) can be rewritten as 
\begin{equation} 
\begin{small}
\begin{aligned}
 f(\bm{x})   &= f(\bm{b})  + \sum\limits_{i \in N} \sum_{\bm{\kappa} \in \Omega_i} I(\bm{\kappa})  + \sum\limits_{\substack{S \subseteq N, |S| > 1}} \sum_{\bm{\kappa} \in \Omega_S} I(\bm{\kappa}) \\
 & = f(\bm{b})  + \sum\limits_{i \in N} \sum_{\bm{\kappa} \in \Omega_i} \phi(\bm{\kappa})  + \sum\limits_{\substack{S \subseteq N, |S| > 1}} \sum_{\bm{\kappa} \in \Omega_S} I(\bm{\kappa}) 
\end{aligned}
\end{small}
\end{equation}
Furthermore, according to the definition of the generic independent effect {\small $\psi(i) = \sum_{\bm{\kappa} \in \Omega_i} \phi(\bm{\kappa}) $} and the generic  interaction effect {\small $J(S) = \sum_{\bm{\kappa} \in \Omega_S} I(\bm{\kappa})$}, it's easy to obtain 
\begin{equation} 
\begin{small}
\begin{aligned}
 f(\bm{x})   &=  f(\bm{b})  + \sum_{i \in N}  \psi(i)  + \sum_{\substack{S \subseteq N,|S| > 1}} J(S) 
\end{aligned}
\end{small}
\end{equation}
Hence, Proposition 1 holds. \\
\end{proof}


\noindent\textbf{Proof of Theorem 1}

\textit{Theorem 1: The Harsanyi dividend interaction {\small $H(S)$} is equivalent to the generic interaction effect {\small $J(S)$} between variables in the set {\small $S$}.}
\begin{equation}
\begin{small}
\begin{aligned}
H(S) = J(S) =  \sum\nolimits_{\bm{\kappa} \in \Omega_S} I(\bm{\kappa}), \ \ \forall S \subseteq N, |S| > 1
\end{aligned}
\end{small}
\end{equation}

\begin{proof}
Actually, it has been proven in \cite{grabisch1999axiomatic} that the Harsanyi dividend {\small $H(S)$} is the \textbf{unique} metric to satisfy the following faithfulness requirement, 
\begin{equation}\label{eqn:apdxfaithfulness}
\forall  \ T\subseteq N, \ f(\bm{x}_T) =    f(\bm{b})+ \sum\nolimits_{S\subseteq T, S \neq \emptyset} H(S),
\end{equation}
where {\small $\bm{x}_T$} denotes a masked sample of the sample $\bm{x}$ when variables in {\small $T$}  keep unchanged and variables in {\small $N\setminus T$} are masked using baseline values $b_i$. 
Accordingly, {\small $f(\bm{x}_T)$} denotes the network output of the masked sample {\small $\bm{x}_T$}.
Thus, as long as we can prove that the generic interaction effect {\small $J(S)$} also satisfies the above faithfulness requirement, we can obtain {\small $H(S) = J(S)$}.

To this end, we only need to prove that {\small $J(S)$} also satisfies Eq.~(\ref{eqn:apdxfaithfulness}). 
Specifically, given an input sample {\small$\bm{x} \in \mathbb{R}^n$}, 
let us consider the Taylor expansion of the network output {\small $f(\bm{x}_T)$} on the masked sample {\small $\bm{x}_T$}, which is expanded at the baseline point {\small $\bm{b} = [b_1, \dots, b_n]^T$}. Then, we have {\small $\forall \ T \subseteq N$}
	\begin{equation}\label{eqn:expansionx_T1}
		\begin{small}
			\begin{aligned}
			f(\bm{x}_T)  & =  f(\bm{b})+  \sum_{k = 1}^{\infty}\sum_{\bm{\kappa}\in O_k} I(\bm{\kappa}|\bm{x}_T) \\
			& =    f(\bm{b})+  \sum_{k = 1}^{\infty}\sum_{\bm{\kappa}\in O_k}  C(\bm{\kappa}) \cdot  \bigtriangledown f(\bm{\kappa}) \cdot \pi(\bm{\kappa}|\bm{x}_T)
			\end{aligned}
		\end{small}
	\end{equation}
where  
	\begin{equation}
		\begin{small}
			\begin{aligned}
			 C(\bm{\kappa}) & = \frac{1}{(\kappa_1+\cdots+\kappa_n)!} \tbinom{\kappa_1+\cdots+\kappa_n}{\kappa_1, \cdots, \kappa_n}  \\[1mm]
 \bigtriangledown f(\bm{\kappa})   &= \frac{\partial^{\kappa_1+\cdots+\kappa_n} f(\bm{b})}{\partial^{\kappa_1}  x_1\cdots \partial^{\kappa_n} x_n } \\[2mm]
 \pi(\bm{\kappa}|\bm{x}_T) & =  \prod_{i=1}^n  [(\bm{x}_T)_i - b_i] ^{\kappa_i} \\
			\end{aligned}
		\end{small}
	\end{equation}
According to the definition of the masked sample {\small $\bm{x}_T$}, we have that {\small $\forall \ i \in T$, $(\bm{x}_T)_i  = x_i$} and {\small $\forall \ i \not\in T$, $(\bm{x}_T)_i  = b_i$}. 
Hence, we obtain that  for any {\small $i \not\in T$}, as long as  {\small $\kappa_i \neq 0, [(\bm{x}_T)_i - b_i]^{\kappa_i} = 0$}. 
Then, among all Taylor expansion terms, only expansion terms corresponding to degrees {\small $\bm{\kappa}$} in the set  {\small $P = \{\bm{\kappa} = [\kappa_1, \dots, \kappa_n]| \forall i \in T, \kappa_i \in \mathbb{N}; \forall i \not\in T, \kappa_i =0\}$} may not be zero. 

Therefore, Eq.~(\ref{eqn:expansionx_T1}) can be re-written as 
	\begin{equation}
		\begin{small}
			\begin{aligned}
			& f(\bm{x} _T)   =   f(\bm{b})+  \sum_{\bm{\kappa} \in P}   C(\bm{\kappa}) \cdot  \bigtriangledown f(\bm{\kappa}) \cdot \pi(\bm{\kappa}|\bm{x}_T) \\
			 & = f(\bm{b})+ \sum_{\bm{\kappa} \in P}   C(\bm{\kappa}) \cdot  \bigtriangledown f(\bm{\kappa}) \cdot   \prod_{i \in T} (x_i - b_i) ^{\kappa_i} \\
			 & = f(\bm{b})+ \sum_{\bm{\kappa} \in P}   C(\bm{\kappa}) \cdot  \bigtriangledown f(\bm{\kappa}) \cdot   \prod_{i \in T} (x_i - b_i) ^{\kappa_i} \cdot \prod_{i \not\in T} (x_i - b_i) ^{0} \\
			 & = f(\bm{b})+  \sum_{\bm{\kappa} \in P}   C(\bm{\kappa}) \cdot  \bigtriangledown f(\bm{\kappa}) \cdot \pi(\bm{\kappa}|\bm{x}) =  f(\bm{b})+ \sum_{\bm{\kappa} \in P}  I(\bm{\kappa}|\bm{x})\\
			\end{aligned}
		\end{small}
	\end{equation}
The third equation holds because $(x_i - b_i)^0 = 1$.

Furthermore, we find that the set $P$ can be divided into multiple disjoint sets as follows, {\small $P = \cup_{S \subseteq T, S \neq \emptyset} \Omega_{S}$}, where {\small $\Omega_{S} = \{\bm{\kappa}| S_{\bm{\kappa}} = S\} = \{\bm{\kappa} = [\kappa_1, \dots, \kappa_n]| \forall i \in S, \kappa_i \in \mathbb{N}^+; \forall i \not\in S, \kappa_i =0\}$} is a set of degree vectors with the receptive field {\small $S$}. 
Then, 
	\begin{equation}\label{eqn:final}
		\begin{small}
			\begin{aligned}
				\quad f(\bm{x}_T)  &= f(\bm{b})+  \sum\nolimits_{S \subseteq T, S \neq \emptyset}\sum\nolimits_{\bm{\kappa} \in \Omega_{S}} I(\bm{\kappa}|\bm{x})			
				\end{aligned}
		\end{small}
	\end{equation}
According to the definition {\small $J(S) \overset{\textrm{def}}{=} \sum\nolimits_{\bm{\kappa} \in \Omega_{S}} I(\bm{\kappa}|\bm{x})$}, Eq. (\ref{eqn:final}) can be written as
	\begin{equation}
		\begin{small}
			\begin{aligned}
   f(\bm{x}_T)  = f(\bm{b})+ \sum\nolimits_{S \subseteq T, S \neq \emptyset} J(S).
			\end{aligned}
		\end{small}
	\end{equation}
That is,  the generic interaction effect {\small $J(S)$} also satisfies the faithfulness requirement in Eq.~(\ref{eqn:apdxfaithfulness}). 
Hence, $\forall S \subseteq N, S \neq \emptyset$, we have $H(S) = J(S)$.
Theorem 1 holds.
\end{proof}

{\section*{Appendix B}
\textbf{Proof of Theorem 2}

\textit{Theorem 2:  In the Gradient $\times$Input method, the attribution {\small $a_i= \frac{\partial f(\bm{x})}{\partial x_i}x_i$} of the input variable $i$ can be reformulated as follows.}
\begin{equation}
\begin{small}
\begin{aligned}
a_i &= \phi(\bm{\kappa}) 
\end{aligned}
\end{small}
\end{equation}
\textit{where $\bm{\kappa} = [\kappa_1, \cdots, \kappa_n]$ is a one-hot degree vector with $\kappa_i = 1$ and  $\forall j \neq i, \kappa_j = 0$.}

\begin{proof}
Let us consider the following first-order Taylor expansion of $f(\bm{0})$ expanded at the sample $\bm{x}$. 
\begin{equation}\label{eqn:theorem2eq1}
\begin{small}
\begin{aligned}
f(\bm{0}) & =  f(\bm{x}) +  \sum\nolimits_{\bm{\kappa} \in O_1} \tilde \phi(\bm{\kappa}) + O(\epsilon) \\
\end{aligned}
\end{small}
\end{equation}
where $\bm{\kappa} = [\kappa_1, \cdots, \kappa_n] \in O_1$ is a one-hot degree vector such that $\exists i, \kappa_i = 1$ and  $\forall j \neq i, \kappa_j = 0$.
Accordingly, the Taylor independent effect $\tilde \phi(\bm{\kappa}) = - \frac{\partial f(\bm{x})}{\partial x_i} x_i$. 

Note that the \textit{Gradient$\times$Input} method explains the output change $f(\bm{x}) -f(\bm{0})$.
Thus, we rewrite Eq. (\ref{eqn:theorem2eq1}) as $f(\bm{x}) = f(\bm{0}) +  \sum\nolimits_{\bm{\kappa} \in O_1} \phi(\bm{\kappa}) + O(\epsilon)$, where the corresponding Taylor independent effect  $\phi(\bm{\kappa}) =  - \tilde \phi(\bm{\kappa}) = \frac{\partial f(\bm{x})}{\partial x_i} x_i$. 
Therefore, we have $a_i = \frac{\partial f(\bm{x})}{\partial x_i} x_i = \phi(\bm{\kappa})$, \textit{i.e.},  the attribution of the \textit{Gradient$\times$Input} method is equivalent to the Taylor independent effect.\\
\end{proof}

\noindent
\textbf{Proof of Theorem 3}

\textit{Theorem 3: In the Occlusion-1 method, the attribution of the input variable $i$, $a_i = f(\bm{x}) - f(\bm{x}|_{x_i = b})$,  can be reformulated as}
\begin{equation} 
\begin{small}
\begin{aligned}
a_i & = \sum\nolimits_{\bm{\kappa} \in \Omega_i} \phi(\bm{\kappa}) +  \sum\nolimits_{S \subseteq N, |S| > 1, i \in S} \sum\nolimits_{\bm{\kappa} \in \Omega_S} I(\bm{\kappa})  \\[3bp]
\end{aligned}
\end{small}
\end{equation}

\begin{proof}
In the \textit{Occlusion-1} method, the attribution of the variable $i$ is formulated as $a_i = f(\bm{x}) - f(\bm{\tilde x})$, where $\bm{\tilde x} = \bm{x}|_{x_i = b}$ denotes the occluded input satisfying that 
\begin{equation}
\begin{small}
\begin{aligned}
\tilde x_i = b, \ \text{and} \ \ \forall j \neq i, \tilde x_j = x_j
\end{aligned}
\end{small}
\end{equation}
Let us consider Taylor expansions of $f(\bm{x})$ and $f(\bm{\tilde x})$ expanded at the baseline point $\bm{b} = [b, \cdots, b]^T$. 
According to Proposition 1, we have
\begin{equation}\label{eqn:expansion_of_occ}
\begin{small}
\begin{aligned}
f(\bm{x}) &= f(\bm{b}) + \sum_{j\in N}\sum_{\bm{\kappa} \in \Omega_j} \phi(\bm{\kappa}|\bm{x}) + \sum_{S \subseteq N,  |S|>1} \sum_{\bm{\kappa} \in \Omega_S} I(\bm{\kappa}|\bm{x})\\
f(\bm{\tilde x}) &= f(\bm{b}) + \sum_{j \in N}\sum_{\bm{\kappa} \in \Omega_j} \phi(\bm{\kappa}|\bm{\tilde x}) + \sum_{S \subseteq N, |S|>1} \sum_{\bm{\kappa} \in \Omega_S} I(\bm{\kappa}|\bm{\tilde x})
\end{aligned}
\end{small}
\end{equation}
where the Taylor independent effect $\phi(\bm{\kappa}|\bm{x})$ and the Taylor interaction effect $I(\bm{\kappa}|\bm{x})$ can both be written into the form $C(\bm{\kappa}) \cdot \bigtriangledown f(\bm{\kappa}) \cdot \prod_{j=1}^n (x_j - b)^{\kappa_j}$. 

Next, let us classify all degree vectors $\bm{\kappa}$ into four cases.
First, when $\bm{\kappa} \in \Omega_i$, then 
\begin{equation}
\begin{small}
\begin{aligned}
\phi(\bm{\kappa}|\bm{\tilde x}) & = C(\bm{\kappa}) \cdot \bigtriangledown f(\bm{\kappa}) \cdot  (\tilde x_i - b)^{\kappa_i} = 0
\end{aligned}
\end{small}
\end{equation}
Thus, in the first case, $\phi(\bm{\kappa}|\bm{x}) - \phi(\bm{\kappa}|\bm{\tilde x}) = \phi(\bm{\kappa}|\bm{x})$.
Second, when $\bm{\kappa} \in \Omega_j (\forall j \neq i)$, we have 
\begin{equation}
\begin{small}
\begin{aligned}
\phi(\bm{\kappa}|\bm{\tilde x}) & = C(\bm{\kappa}) \cdot \bigtriangledown f(\bm{\kappa}) \cdot  (\tilde x_j - b)^{\kappa_j} \\
& = C(\bm{\kappa}) \cdot \bigtriangledown f(\bm{\kappa}) \cdot  (x_j - b)^{\kappa_j} = \phi(\bm{\kappa}|\bm{x})
\end{aligned}
\end{small}
\end{equation}
Thus, in the second case, $\phi(\bm{\kappa}|\bm{x}) - \phi(\bm{\kappa}|\bm{\tilde x}) = 0$.
Third, when $\bm{\kappa} \in \Omega_S (\forall S \ni i)$,
\begin{equation}
\begin{small}
\begin{aligned}
I(\bm{\kappa}|\bm{\tilde x}) & = C(\bm{\kappa}) \cdot \bigtriangledown f(\bm{\kappa}) \cdot  \prod_{m \in S} (x_m - b)^{\kappa_m} \\
& = C(\bm{\kappa}) \cdot \bigtriangledown f(\bm{\kappa}) \cdot  (\tilde x_i - b)^{\kappa_i}  \prod_{m \in S, m \neq i} (\tilde x_m - b)^{\kappa_m} \\
& = 0
\end{aligned}
\end{small}
\end{equation}
Hence, in the third case, $I(\bm{\kappa}|\bm{x}) - I(\bm{\kappa}|\bm{\tilde x}) = I(\bm{\kappa}|\bm{x})$.
 Forth, when $\bm{\kappa} \in \Omega_S (\forall S \not\ni i)$,
\begin{equation}
\begin{small}
\begin{aligned}
I(\bm{\kappa}|\bm{\tilde x}) & = C(\bm{\kappa}) \cdot \bigtriangledown f(\bm{\kappa}) \cdot \prod_{m \in S} (\tilde x_m - b)^{\kappa_m} \\
& = C(\bm{\kappa}) \cdot \bigtriangledown f(\bm{\kappa}) \cdot  \prod_{m \in S} (x_m - b)^{\kappa_m} =  I(\bm{\kappa}|\bm{x})
\end{aligned}
\end{small}
\end{equation}
Hence, in the forth case, $I(\bm{\kappa}|\bm{x}) - I(\bm{\kappa}|\bm{\tilde x}) = 0$.

In summary, combing the above equations, the attribution $a_i= f(\bm{x}) - f(\bm{\tilde x})$ in the \textit{Occlusion-1} method can be reformulated as
\begin{equation}
\begin{small}
\begin{aligned}
a_i = \sum_{\bm{\kappa} \in \Omega_i}\phi(\bm{\kappa}|\bm{x}) + \sum_{S \subseteq N, S|>1, i \in S}\sum_{\bm{\kappa} \in \Omega_S} I(\bm{\kappa}|\bm{x})
\end{aligned}
\end{small}
\end{equation}
\end{proof}

\noindent 
\textbf{Proof of Theorem 4}

\textit{Theorem 4: In the Occlusion-patch method, the attribution of the pixel $i$ in the patch $S_j$ ($i \in S_j$) is $a_i = f(\bm{x}) - f(\bm{x}|_{\bm{x}_{S_j} = \bm{b}})$, which can be reformulated as}
\begin{equation} 
\begin{small}
\begin{aligned}
a_i & = \sum_{m \in S_j} \sum_{\bm{\kappa} \in \Omega_m}\phi(\bm{\kappa}) + \sum_{|S|> 1, S \cap S_j \neq \emptyset} \sum_{\bm{\kappa} \in \Omega_S} I(\bm{\kappa}) \\
\end{aligned}
\end{small}
\end{equation}

\begin{proof}
In the \textit{Occlusion-patch} method, the attribution of pixel $i$ in the patch $S_j$ is $a_i = f(\bm{x}) - f(\bm{\tilde x})$, where $\bm{\tilde x} = \bm{x}|_{\bm{x}_{S_j} = b}$ denotes the occluded input \textit{s.t.}
\begin{equation}
\begin{small}
\begin{aligned}
\forall m \in S_j, \tilde x_m = b \quad \text{and} \quad \forall m \not\in S_j, \tilde x_m = x_m 
\end{aligned}
\end{small}
\end{equation}
Then, we consider Taylor expansions of $f(\bm{x})$ and $f(\bm{\tilde x})$, which is expanded at the baseline point $\bm{b} = [b, \cdots, b]^T$. According to Proposition 1, we have
\begin{equation}\label{eqn:expansion_of_occ_p}
\begin{small}
\begin{aligned}
f(\bm{x}) &= f(\bm{b}) + \sum_{j\in N}\sum_{\bm{\kappa} \in \Omega_j} \phi(\bm{\kappa}|\bm{x}) + \sum_{S \subseteq N, |S|> 1} \sum_{\bm{\kappa} \in \Omega_S} I(\bm{\kappa}|\bm{x})\\
f(\bm{\tilde x}) &= f(\bm{b}) + \sum_{j \in N}\sum_{\bm{\kappa} \in \Omega_j} \phi(\bm{\kappa}|\bm{\tilde x}) + \sum_{S \subseteq N, |S|> 1} \sum_{\bm{\kappa} \in \Omega_S} I(\bm{\kappa}|\bm{\tilde x})
\end{aligned}
\end{small}
\end{equation}
where the Taylor independent effect $\phi(\bm{\kappa}|\bm{x})$ and the Taylor interaction effect $I(\bm{\kappa}|\bm{x})$ can both be written into the form $C(\bm{\kappa}) \cdot \bigtriangledown f(\bm{\kappa}) \cdot \prod_{j=1}^n (x_j - b)^{\kappa_j}$.

Next, let us further classify all degree vectors $\bm{\kappa}$ into four cases.
First, when $\bm{\kappa} \in \Omega_m (\forall m \in S_j)$,
\begin{equation}
\begin{small}
\begin{aligned}
\phi(\bm{\kappa}|\bm{\tilde x}) & = C(\bm{\kappa}) \cdot \bigtriangledown f(\bm{\kappa}) \cdot  (\tilde x_m - b)^{\kappa_m} = 0
\end{aligned}
\end{small}
\end{equation}
Hence, in the first case, $\phi(\bm{\kappa}|\bm{x}) - \phi(\bm{\kappa}|\bm{\tilde x}) = \phi(\bm{\kappa}|\bm{x})$. 
Second, when $\bm{\kappa} \in \Omega_m (\forall m \not\in S_j)$, we have 
\begin{equation}
\begin{small}
\begin{aligned}
\phi(\bm{\kappa}|\bm{\tilde x}) & = C(\bm{\kappa}) \cdot \bigtriangledown f(\bm{\kappa}) \cdot  (\tilde x_m - b)^{\kappa_m} \\
& = C(\bm{\kappa}) \cdot \bigtriangledown f(\bm{\kappa}) \cdot  (x_m - b)^{\kappa_m} = \phi(\bm{\kappa}|\bm{x})
\end{aligned}
\end{small}
\end{equation}
Hence, in the second case, $\phi(\bm{\kappa}|\bm{x}) - \phi(\bm{\kappa}|\bm{\tilde x}) = 0$.
Third, when $\bm{\kappa} \in \Omega_S$ and $S \cap S_j \neq \emptyset$,
\begin{equation} 
\begin{small}
\begin{aligned}
& I(\bm{\kappa}|\bm{\tilde x}) = C(\bm{\kappa}) \cdot \bigtriangledown f(\bm{\kappa}) \cdot  \prod_{m \in S} (x_m - b)^{\kappa_m} \\
 & = C(\bm{\kappa}) \cdot \bigtriangledown f(\bm{\kappa})  \!\! \prod_{m \in S \cap S_j}  (\tilde x_m - b_m)^{\kappa_m} 
 \!\! \prod_{m \in S, m \not\in S_j} (\tilde x_m - b)^{\kappa_m}\\
 &  = 0
\end{aligned}
\end{small}
\end{equation}
Thus, in the third case, $I(\bm{\kappa}|\bm{x}) - I(\bm{\kappa}|\bm{\tilde x}) = I(\bm{\kappa}|\bm{x})$.
Forth, when $\bm{\kappa} \in \Omega_S$ and $S \cap S_j = \emptyset$,
\begin{equation}
\begin{small}
\begin{aligned}
I(\bm{\kappa}|\bm{\tilde x}) & = C(\bm{\kappa}) \cdot \bigtriangledown f(\bm{\kappa}) \cdot \prod_{m \in S} (\tilde x_m - b)^{\kappa_m} \\
& = C(\bm{\kappa}) \cdot \bigtriangledown f(\bm{\kappa}) \cdot  \prod_{m \in S} (x_m - b)^{\kappa_m} 
= I(\bm{\kappa}|\bm{x})
\end{aligned}
\end{small}
\end{equation}
Thus, in the forth case, $I(\bm{\kappa}|\bm{x}) - I(\bm{\kappa}|\bm{\tilde x}) = 0$. 

In summary, combining the above equations, the attribution $a_i= f(\bm{x}) - f(\bm{\tilde x})$ of the \textit{Occlusion-patch} method can be reformulated as
\begin{equation} 
\begin{small}
\begin{aligned}
a_i = \sum_{m \in S_j} \sum_{\bm{\kappa} \in \Omega_m}\phi(\bm{\kappa}|\bm{x}) + \sum_{|S|>1, S \cap S_j \neq \emptyset} \sum_{\bm{\kappa} \in \Omega_S} I(\bm{\kappa}|\bm{x})
\end{aligned}
\end{small}
\end{equation}
\end{proof}

\noindent
\textbf{Proof of Theorem 5}

\textit{Theorem 5: In the Prediction Difference method, the attribution $a_i = \mathbb{E}_{b_i \sim p(b_i)} [f(\bm{x}) -  f(\bm{x}|_{x_i = b_i})]$ of the input variable $i$ can be reformulated as:
\begin{equation}
\begin{small}
\begin{aligned}
    a_i   &= \mathbb{E}_{\bm{b} \sim p(\bm{b})}  [ \sum_{\bm{\kappa} \in \Omega_i} \phi(\bm{\kappa}|\bm{b}) +  \sum_{|S|> 1, i \in S} \sum_{\bm{\kappa} \in \Omega_S} I(\bm{\kappa}|\bm{b})],  \\
\end{aligned}
\end{small}
\end{equation}
where $\bm{b} = b \cdot \bm{1}$.}

\begin{proof}
According to Theorem 3, the attribution $f(\bm{x}) -  f(\bm{x}|_{x_i = b_i})$ can be reformulated as follows. 
\begin{equation}
\begin{small}
\begin{aligned}
 f(\bm{x}) -  f(\bm{x}|_{x_i = b_i}) = \sum_{\kappa \in \Omega_i}\phi(\bm{\kappa}|\bm{b}) + \sum_{|S|> 1, i \in S}\sum_{\bm{\kappa} \in \Omega_S} I(\bm{\kappa}|\bm{b})
\end{aligned}
\end{small}
\end{equation}
where $\bm{b} = b \cdot \bm{1}$ is a baseline point, and $b \sim p(b)$ is sampled from the baseline distribution. Then, we can obtain that
\begin{equation}
\begin{small}
\begin{aligned}
 a_i  & = \mathbb{E}_{\bm{b} \sim p(\bm{b})} [\sum_{\kappa \in \Omega_i}\phi(\bm{\kappa}|\bm{b}) + \sum_{|S|>1, i \in S}\sum_{\bm{\kappa} \in \Omega_S} I(\bm{\kappa}|\bm{b})].
\end{aligned}
\end{small}
\end{equation}
Thus, the conclusion holds.\\ 
\end{proof}

\noindent
\textbf{Proof of Theorem 6}

\textit{Theorem 6: In the Grad-CAM method, let us consider each neuron {\small $A_{ij}^k$} in the convolutional layer as an input variable. Then, the attribution of each input variable can be reformulated as}
\begin{equation} 
\begin{small}
\begin{aligned}
    \tilde a_{ij}^k = \phi(\bm{\kappa}) = \frac{\partial g(A)}{\partial A_{ij}^k} A_{ij}^k
    \end{aligned}
    \end{small}
    \end{equation}
\textit{where $g(A)$ represents the explanatory model of the DNN in the Grad-CAM method (proved in \cite{selvaraju2017grad}). 
Besides, $\bm{\kappa} = [\kappa_1, \cdots, \kappa_n]$ is a one-hot degree vector with $\kappa_i = 1$ and  $\forall j \neq i, \kappa_j = 0$.}

\begin{proof}
On one hand, the \textit{Grad-CAM} method estimates the total attribution of all neurons at the $(i,j)$ location as 
\begin{equation} 
\begin{small}
\begin{aligned}
\tilde a_{ij} = \sum\nolimits_{k=1}^K \alpha_k A_{ij}^k, \ s.t. \ \  \tilde a_{ij}^k = \alpha_k A_{ij}^k. 
   \end{aligned}
    \end{small}
    \end{equation}
where the attribution of each neuron {\small $A_{ij}^k$} in the $k$-th channel is considered as $\tilde a_{ij}^k = \alpha_k A_{ij}^k$.

On the other hand, \cite{selvaraju2017grad} has proven that Grad-CAM actually explains the DNN as the following linear model of global average pooled features $F^k$, where $F^k = \frac{1}{W \times H} \sum_{i=1}^W \sum_{j=1}^H A_{ij}^k$. 
\begin{equation} 
\begin{small}
\begin{aligned}
y \xlongequal{\text{explained}}  g(A)  &= \sum\nolimits_{k=1}^K (W \times H  \cdot \alpha_k) \cdot F^k \  + \ \epsilon \\
& = \sum\nolimits_{k=1}^K \alpha_k \sum_{i=1}^W \sum_{j=1}^H A_{ij}^k \  + \ \epsilon
    \end{aligned}
    \end{small}
    \end{equation}
Therefore, we have
\begin{equation}
\begin{small}
\begin{aligned}
 \frac{\partial g(A)}{\partial A_{ij}^k} \cdot A_{ij}^k  =  \alpha_k A_{ij}^k  = \tilde a_{ij}^k 
\end{aligned}
\end{small}
\end{equation}
That is, the attribution {\small $\tilde a_{ij}^k$} of the \textit{Grad-CAM} method can be re-written as the product of the input variable $ A_{ij}^k$ and the gradient of the explanatory model $g(A)$ \textit{w.r.t.} the input variable.

Then, combining with Theorem 2, we obtain that {\small $\tilde a_{ij}^k = \phi(\bm{\kappa})$}.
The \textit{Grad-CAM} method only allocates a specific Taylor independent effect $\phi(\bm{\kappa})$ of the neuron $A_{ij}^k$ to its attribution.
\\
\end{proof}

\noindent
\textbf{Proof of Theorem 7}

\textit{Theorem 7: In the Integrated Gradients method, the attribution of the input variable $i$ is formulated as
\begin{equation}\label{eqn:IGformulation}
\begin{small}
\begin{aligned}
a_{i} = ( x_i - b_i) \cdot \int_{\alpha = 0}^1 \frac{\partial {f(\bm{c})}}{\partial {c_i}}|_{\bm{c} = \bm{b} + \alpha( \bm{x} - \bm{b})}  d\alpha
\end{aligned}
\end{small}
\end{equation}
The attribution $a_i$ can be reformulated as}
\begin{equation}\label{eqn:IGreformulation}
\begin{small}
\begin{aligned}
a_i & = \sum_{\bm{\kappa} \in \Omega_i}  \phi(\bm{\kappa})  + \sum_{\substack{|S|>1, i \in S}}\sum_{\bm{\kappa} \in \Omega_S} \frac{\kappa_i}{\sum_{i \in N} \kappa_{i}}  I(\bm{\kappa}) \\
\end{aligned}
\end{small}
\end{equation}

\begin{proof}
Let us first consider the $K$-order Taylor expansion of $f(\bm{x})$ at the baseline point $\bm{b}$.
\begin{equation}\label{eqn:IGTaylorexpansion}
\begin{small}
\begin{aligned}
f(\bm{x}) & = f(\bm{b})+\sum_{k=1}^K\sum_{\bm{\kappa} \in O_k} I(\bm{\kappa}) \\
& = \sum_{k=0}^K\sum_{\bm{\kappa} \in O_k} C(\bm{\kappa}) \bigtriangledown f(\bm{\kappa}) \pi(\bm{\kappa}) \\
\end{aligned}
\end{small}
\end{equation}
Specifically,  the coefficient, the partial derivative, and the product term are computed as
\begin{equation}
\begin{small}
\begin{aligned}
C(\bm{\kappa}) & = \frac{1}{(\kappa_1 + \dots + \kappa_n)!}\cdot \tbinom{\kappa_1 + \dots + \kappa_n}{\kappa_1, \cdots, \kappa_n} \\[2pt]
\bigtriangledown f(\bm{\kappa}) & = \frac{\partial^{\kappa_1 + \dots + \kappa_i + \dots   + \kappa_n} f(\bm{b})}{\partial^{\kappa_1} x_1 \cdots \partial^{\kappa_i} x_i \cdots \partial^{\kappa_n} x_n} \\[2pt]
\pi(\bm{\kappa}) &=  \prod\nolimits_{i=1}^n (x_i - b_i)^{\kappa_i} 
\end{aligned}
\end{small}
\end{equation}
where $\bm{\kappa} = [\kappa_1, \dots, \kappa_n] \in \mathbb{N}^n$ denotes the degree vector of $I(\bm{\kappa})$. 
In particular, we define $C(\bm{0}) = 1, \bigtriangledown f(\bm{0}) = f(\bm{b})$, and $\pi(\bm{0}) = 1$.

To prove Theorem 7, let us consider the first-order partial derivative $g_i(\bm{x}) = \frac{\partial f(\bm{x})}{\partial x_i}$ in Eq. (\ref{eqn:IGformulation}), whose Taylor expansion can be written as follows. 
\begin{equation}\label{eqn:gexpansion}
\begin{small}
\begin{aligned}
g_i(\bm{x}) & = \sum_{k=0}^K\sum_{\bm{\kappa} \in O_k} C(\bm{\kappa}) \bigtriangledown g(\bm{\kappa}) \pi(\bm{\kappa}) 
\end{aligned}
\end{small}
\end{equation}
where $\bigtriangledown g(\bm{\kappa}) =  \frac{\partial^{\kappa_1 + \dots + \kappa_n} g_i(\bm{x})}{\partial^{\kappa_1} x_1 \cdots \partial^{\kappa_n} x_n }|_{\bm{x} = \bm{b}}$.
In this way, 
\begin{equation}
\begin{small}
\begin{aligned}
& \frac{\partial {f(\bm{c})}}{\partial {c_i}}|_{\bm{c} = \bm{b} + \alpha( \bm{x} - \bm{b})} = \ g_i( \bm{b} + \alpha( \bm{x} - \bm{b}))   \\
= &  \sum_{k=0}^K\sum_{\bm{\kappa} \in O_k} C(\bm{\kappa}) \bigtriangledown g(\bm{\kappa}) \prod_{i=1}^n [\alpha (x_i - b_i)]^{\kappa_i} \\
=  & \sum_{k=0}^K\sum_{\bm{\kappa} \in O_k} C(\bm{\kappa}) \bigtriangledown g(\bm{\kappa}) \pi(\bm{\kappa}) \cdot \alpha^{\sum_i \kappa_i}
\end{aligned}
\end{small}
\end{equation}
Thus, the attribution in Eq. (\ref{eqn:IGformulation}) can be re-written as
\begin{equation}
\begin{small}
\begin{aligned}
 a_i &=  (x_i-b_i) \cdot \left( \int_{0}^1 \sum_{k=0}^K\sum_{\bm{\kappa} \in O_k} C(\bm{\kappa})\bigtriangledown g(\bm{\kappa}) \pi(\bm{\kappa})    \cdot\alpha^{\sum_i \kappa_i} d\alpha \right) \\
& = (x_i-b_i) \cdot \left(\sum_{k=1}^K\sum_{\bm{\kappa} \in O_k} C(\bm{\kappa})\bigtriangledown g(\bm{\kappa}) \pi(\bm{\kappa})\right)    \cdot \left(\int_{0}^1 \alpha^{\sum_i \kappa_i} d\alpha \right) \\
& = \sum_{k=0}^K\sum_{\bm{\kappa} \in O_k} \frac{1}{\sum_i \kappa_i + 1}C(\bm{\kappa}) \bigtriangledown g(\bm{\kappa})  \pi(\bm{\kappa})  (x_i-b_i)  
\end{aligned}
\end{small}
\end{equation}
At the same time, we find that
\begin{equation}
\begin{small}
\begin{aligned}
\bigtriangledown g(\bm{\kappa}) & = \frac{\partial^{\kappa_1 + \dots + (\kappa_i + 1) + \dots   + \kappa_n} f(\bm{b})}{\partial^{\kappa_1}  x_1 \cdots \partial^{\kappa_i+1} x_i \cdots \partial^{\kappa_n} x_n}  = \bigtriangledown f(\bm{\tilde \kappa}) \\
\end{aligned}
\end{small}
\end{equation}
where $\bm{\tilde \kappa} = [\tilde \kappa_1, \dots, \tilde \kappa_n] = [\kappa_1, \kappa_2, \dots, \kappa_i + 1, \dots, \kappa_n]$. 
Then, we have 
\begin{equation}\label{eqn:IGreformulation2}
\begin{small}
\begin{aligned}
 a_i  & = \sum_{k=0}^K\sum_{\bm{\kappa} \in O_k} \frac{1}{\sum_i \kappa_i + 1}C(\bm{\kappa}) \bigtriangledown f(\bm{\tilde \kappa})  \pi(\bm{\kappa})  (x_i-b_i)  \\
 & = \sum_{k=1}^K  \sum_{\bm{\tilde \kappa} \in O_k}  \frac{1}{\sum_i \kappa_i + 1}\cdot (\kappa_i + 1) C(\bm{\tilde \kappa}) \bigtriangledown f(\bm{\tilde \kappa})  \pi(\bm{\tilde \kappa}) \\
 & = \sum_{k=1}^K \sum_{\bm{\tilde \kappa} \in O_k} \frac{\tilde \kappa_i}{\sum_{i \in N} \tilde \kappa_i} I(\bm{\tilde \kappa}) \\
 \end{aligned}
\end{small}
\end{equation}
The second equation holds because $C(\bm{\kappa}) = (\kappa_i + 1) C(\bm{\tilde \kappa})$ and $\pi(\bm{\kappa}) (x_i - b_i)  = \pi(\bm{\tilde \kappa})$.

Eq. (\ref{eqn:IGreformulation2}) indicates that the \textit{Integrated Gradients} method allocates $\frac{\kappa_i}{\sum_{i \in N} \kappa_i}$ ratio of each Taylor expansion term to the attribution $a_i$. 
Specifically, when $\bm{\kappa} \in \Omega_i$ (\textit{i.e.}, $\kappa_i > 0$ and $\forall j \neq i, \kappa_j = 0$), this method allocates $100\%$ ratio of each Taylor independent effect $\phi(\bm{\kappa})$ to the attribution $a_i$. 
When $\bm{\kappa} \in \Omega_S$,  this method allocates  $\frac{\kappa_i}{\sum_{i \in N}  \kappa_i}$ ratio of each Taylor interaction effect $I(\bm{\kappa})$ to the attribution $a_i$. That is, 
\begin{equation}
\begin{small}
\begin{aligned}
a_i & = \sum_{\bm{\kappa} \in \Omega_i}  \phi(\bm{\kappa})  + \sum_{\substack{|S|>1, i \in S}}\sum_{\bm{\kappa} \in \Omega_S} \frac{\kappa_i}{\sum_{i \in N} \kappa_{i}}  I(\bm{\kappa})
  \end{aligned}
\end{small}
\end{equation}
Hence, Theorem 7 holds. \\
\end{proof}

\noindent
\textbf{Proof of Theorem 8}

\textit{Theorem 8: In the Expected Gradients method, the attribution of the input variable $i$, $a_i = \mathbb{E}_{\bm{b} \sim p(\bm{b})}\cdot [(x_i-b_i) \int_{\alpha=0}^1 \frac{\partial f(\bm{c})}{\partial c_i}|_{\bm{c} = \bm{b} + \alpha (\bm{x} - \bm{b})} d\alpha] $,  can be reformulated as:}
\begin{equation}
\begin{small}
\begin{aligned}
    a_i   &= \mathbb{E}_{\bm{b} \sim p(\bm{b})}  [\sum_{\bm{\kappa}\in \Omega_i} \phi(\bm{\kappa}|\bm{b}) + \sum_{\substack{ i \in S}} \sum_{\bm{\kappa} \in \Omega_S} \frac{\kappa_i}{\sum_{i \in S} \kappa_{i}} I(\bm{\kappa}|\bm{b})] \\
\end{aligned}
\end{small}
\end{equation}

\begin{proof}
According to Theorem 7, the attribution estimated by the \textit{Integrated Gradients} method, $a_i = (x_i-b_i) \int_{\alpha=0}^1 \frac{\partial f(\bm{c})}{\partial c_i}|_{\bm{c} = \bm{b} + \alpha (\bm{x} - \bm{b})} d\alpha$,  can be reformulated as follows.
\begin{equation}
\begin{small}
\begin{aligned}
a_i & = \sum_{\bm{\kappa} \in \Omega_i}  \phi(\bm{\kappa})  + \sum_{\substack{S \subseteq N, i \in S}}\sum_{\bm{\kappa} \in \Omega_S} \frac{\kappa_i}{\sum_{i \in S} \kappa_{i}}  I(\bm{\kappa}) \\
\end{aligned}
\end{small}
\end{equation}
It is easy to obtain that the attribution estimated by the \textit{Expected Gradients} method can be reformulated as
\begin{equation}
\begin{small}
\begin{aligned}
    a_i   &= \mathbb{E}_{\bm{b} \sim p(\bm{b})}  [\sum_{\bm{\kappa}\in \Omega_i} \phi(\bm{\kappa}|\bm{b}) + \sum_{\substack{ i \in S}} \sum_{\bm{\kappa} \in \Omega_S} \frac{\kappa_i}{\sum_{i \in S} \kappa_{i}} I(\bm{\kappa}|\bm{b})]. \\
\end{aligned}
\end{small}
\end{equation}
Hence, the conclusion holds. \\
\end{proof}

\noindent
\textbf{Proof of Theorem 9}
\vspace{3pt}

\textit{Theorem 9: In the Shapley value method, the attribution of the input variable $i$ is formulated as,
\begin{equation}\label{eqn:shapleyformulation}
\begin{small}
\begin{aligned}
a_i = \sum\nolimits_{S \subseteq N \setminus \{i\}} p(S)\cdot [f(\bm{x}_{S \cup i})  - f(\bm{x}_{S})],  
 \end{aligned}
 \end{small}
\end{equation}
The attribution can be reformulated as}
\begin{equation}\label{eqn:shapleyreformulation}
\begin{small}
\begin{aligned}
a_i & =  \sum_{\bm{\kappa} \in \Omega_i} \phi(\bm{\kappa}) +  \sum_{\substack{|S|>1, i \in S}}\sum_{\bm{\kappa} \in \Omega_S}  \frac{1}{|S|} I(\bm{\kappa})
 \end{aligned}
 \end{small}
\end{equation}

\begin{proof}
First, let us expand the network outputs $f(\bm{x}_{S \cup i})$ and $f(\bm{x}_S)$ at the baseline point $\bm{b}$, respectively. 
According to Proposition 1,
\begin{equation}
\begin{small}
\begin{aligned}
& f(\bm{x}_{S \cup i}) =   f(\bm{b}) + \sum_{j \in S \cup i}\sum_{\bm{\kappa} \in \Omega_j} \phi(\bm{\kappa}) 
+ \sum_{|T|>1, T \subseteq S \cup i}\sum_{\bm{\kappa} \in \Omega_{T}} I(\bm{\kappa}) \\
& f(\bm{x}_{S}) = f(\bm{b}) + \sum_{j \in S}\sum_{\bm{\kappa} \in \Omega_j} \phi(\bm{\kappa}) 
+ \sum_{|T|>1, T \subseteq S}\sum_{\bm{\kappa} \in \Omega_{T}} I(\bm{\kappa}) \\
\end{aligned}
\end{small}
\end{equation}
Therefore, the marginal contribution of $i$ is 
\begin{equation}
\begin{small}
\begin{aligned}
& f(\bm{x}_{S \cup i})  - f(\bm{x}_{S}) = \sum_{\bm{\kappa} \in \Omega_i} \phi(\bm{\kappa}) + 
\sum_{\substack{T \subseteq S \cup i, \\ T \not\subseteq S}} \sum_{\bm{\kappa} \in \Omega_T}  I(\bm{\kappa}) \\
& = \sum\nolimits_{\bm{\kappa} \in \Omega_i} \phi(\bm{\kappa}) + 
\sum\nolimits_{\substack{i \in T, \\ T \setminus \{i\} \subseteq S}} \sum\nolimits_{\bm{\kappa} \in \Omega_T}  I(\bm{\kappa}) \end{aligned}
\end{small}
\end{equation}
Then, the attribution of the \textit{Shapley value} method in Eq. (\ref{eqn:shapleyformulation}) can be re-written as
\begin{equation}
\begin{small}
\begin{aligned}
a_i & = \sum\nolimits_{S \subseteq N \setminus \{i\}} p(S) \cdot  \left(\sum\nolimits_{\bm{\kappa} \in \Omega_i} \phi(\bm{\kappa})\right) \\
 &+  \sum\nolimits_{S \subseteq N \setminus \{i\}} p(S) \cdot  \left(\sum\nolimits_{\substack{i \in T, \\ T \setminus \{i\} \subseteq S}}\sum\nolimits_{\bm{\kappa} \in \Omega_T}  I(\bm{\kappa})\right) \\
\end{aligned}
\end{small}
\end{equation}
Let $|S| = s$. 
On one hand, the first summation term $A_1$ \textit{w.r.t.} $\phi(\bm{\kappa})$ is 
\begin{equation}
\begin{small}
\begin{aligned}
A_1 & =  \sum\nolimits_{S \subseteq N \setminus \{i\}} p(S) \cdot  \left(\sum\nolimits_{\bm{\kappa} \in \Omega_i} \phi(\bm{\kappa})\right) \\
& = \sum\nolimits_{s=0}^{n-1}  \tbinom{n-1}{s} p(S)\cdot \left(\sum\nolimits_{\bm{\kappa} \in \Omega_i} \phi(\bm{\kappa})\right)  \\
& = \sum\nolimits_{s=0}^{n-1} \frac{1}{n} \cdot \left(\sum\nolimits_{\bm{\kappa} \in \Omega_i} \phi(\bm{\kappa})\right)  \ (\because p(S)  = \frac{s!(n-1-s)!}{n!}) \\
& = \sum\nolimits_{\bm{\kappa} \in \Omega_i} \phi(\bm{\kappa})
\end{aligned}
\end{small}
\end{equation}

In terms of the second summation term \textit{w.r.t.} $I(\bm{\kappa})$, 
\begin{equation}
\begin{small}
\begin{aligned}
A_2 &= \sum\nolimits_{S \subseteq N \setminus \{i\}} p(S) \cdot  \left(\sum\nolimits_{\substack{i \in T, \\ T \setminus \{i\} \subseteq S}}\sum\nolimits_{\bm{\kappa} \in \Omega_T}  I(\bm{\kappa})\right) \\
& = \sum\nolimits_{T\subseteq N, i \in T} C_T \sum\nolimits_{\bm{\kappa} \in \Omega_T}  I(\bm{\kappa})
\end{aligned}
\end{small}
\end{equation}
we observe that $A_2$ can be written as a weighted sum of Taylor interaction effects,  where the coefficient $C_T$ corresponds to how many times the specific subset $T$ appears in the summation. 
Given a specific subset $T$, we can obtain all summation terms by traversing all potential subsets $S$ satisfying $S \subseteq N \setminus \{i\}$ and $S \supseteq (T \setminus \{i\})$.
Furthermore, we can traverse $S$ by choosing elements in $S \setminus (T\setminus \{i\})$ from the larger set $N \setminus T$. 
Hence,  
\begin{equation}
\begin{small}
\begin{aligned}
C_T &= \sum_{s=t-1}^{n-1} p(S) \tbinom{n-t}{s-(t-1)}  = \sum_{s=t-1}^{n-1} \frac{s!}{n!} \frac{(n-t)!}{(s-t+1)!}  \\
& = \sum_{s=t-1}^{n-1} \frac{(n-t)! (t-1)!}{n!} \frac{s!}{(t-1)!(s-t+1)!}  \\
& = \frac{1}{n \tbinom{n-1}{t-1}} \sum_{s = t-1}^{n-1} \tbinom{s}{t-1} \quad \text{Hockey-stick identity}  \\
& = \frac{1}{n \tbinom{n-1}{t-1}}   \tbinom{n}{t} = \frac{1}{t}  
\end{aligned}
\end{small}
\end{equation}
By combining above equations,  the attribution of the \textit{Shapley value} method can be reformulated as
\begin{equation}
\begin{small}
\begin{aligned}
a_i &= \sum\nolimits_{\bm{\kappa} \in \Omega_i} \phi(\bm{\kappa}) +  \sum\nolimits_{T \subseteq N, i \in T}   \sum\nolimits_{\bm{\kappa} \in \Omega_T} \frac{1}{|T|} I(\bm{\kappa})
\end{aligned}
\end{small}
\end{equation}
Hence, the conclusion holds. \\
\end{proof}

\noindent
\textbf{Proof of Theorem 10}

\textit{Theorem 10: In the LRP-$\epsilon$ method, when the ReLU operation is used as the activation function, the attribution of the variable $i$ can be reformulated as}
\begin{equation}
\begin{small}
\begin{aligned}
a_i = \phi(\bm{\kappa}) = \frac{\partial f(\bm{x})}{\partial x_i}x_i
\end{aligned}
\end{small}
\end{equation}
    \textit{where $\bm{\kappa}= [\kappa_1, \cdots, \kappa_n]$ is a one-hot degree vector with $\kappa_i = 1$ and  $\forall j \neq i, \kappa_j = 0$.}

\begin{proof}
It has been proven in \cite{Marco2018Towards} that the \textit{LRP-$\epsilon$} and \textit{Gradient$\times$Input} are equivalent when ReLU is adopted as the activation function. Hence, the two methods allocates the Taylor interaction effects (independent effects) in the same way.
Combining with Theorem 2, Theorem 10 holds. 
\\
\end{proof}

\textbf{Remark 1:}
\textit{In terms of back-propagation attribution methods, we mainly analyze the layer-wise propagation of attributions, so we consider the feature dimension {\small $x_j^{(l)}$} in the $l$-th layer as the target output, and consider all feature dimensions in the {\small $(l-1)$}-th layer as input variables. 
For simplicity, in the following proofs, we rewrite the target output {\small $x_j^{(l)}$} as $y$, and rewrite each input variable $x_i^{(l-1)}$ as $x_i$. 
Similarly,  notations {\small $\tilde x_j^{(l)}, \tilde x_i^{(l-1)}, W_{ij}, z_{ij}$} are represented by 
 {\small $\tilde y, \tilde x_i,  W_i, z_i$}, respectively. \\ 
}

\noindent
\textbf{Proof of Theorem 11}

\textit{Theorem 11: In the LRP-$\alpha\beta$ method, let us analyze the layer-wise propagation of attributions. 
Then, for the input variable {\small $i \in N^+=  \{i|z_i = W_ix_i > 0\}$}, its attribution $a_i =  \frac{\alpha \cdot z_i}{\sum_{i' \in N^+} z_{i'}} (y - \tilde y)$  can be reformulated as
\begin{equation}\label{eqn:LRPab+}
\begin{small}
\begin{aligned}
\!\!\!\! a_i \! = \!
\alpha [\sum_{\bm{\kappa}\in \Omega_i} \phi(\bm{\kappa}) + \! \sum_{i \in S} \! \sum_{\bm{\kappa} \in \Omega_S}  c_i I(\bm{\kappa})  + \!\!\! \sum_{S \subseteq N^-} \sum_{\bm{\kappa} \in \Omega_S} d_i I(\bm{\kappa})] 
\end{aligned}
\end{small}
\end{equation}
where $c_i = \frac{\kappa_i}{\sum_{i \in N^+} \kappa_i}, d_i = \frac{z_i}{\sum_{i' \in N^+} z_{i'}}$. For the variable  {\small $i \in N^-=  \{i|z_i = W_ix_i \leq 0\}$}, the attribution $a_i =  \frac{\beta \cdot z_i}{\sum_{i' \in N^-} z_{i'}} (y - \tilde y)$ can be reformulated as
\begin{equation}\label{eqn:LRPab-}
\begin{small}
\begin{aligned}
\!\!\!\! a_i \! = \!
\beta   [\sum_{\bm{\kappa}\in \Omega_i} \phi(\bm{\kappa}) + \! \sum_{i \in S} \sum_{\bm{\kappa} \in \Omega_S}  \tilde c_i I(\bm{\kappa}) + \!\!\!  \sum_{S \subseteq N^+}\sum_{\bm{\kappa} \in \Omega_S} \tilde d_i I(\bm{\kappa})] 
\end{aligned}
\end{small}
\end{equation}
where $\tilde c_i = \frac{\kappa_i}{\sum_{i \in N^-} \kappa_i}, \tilde d_i =  \frac{z_i}{\sum_{i' \in N^-} z_{i'}}$.} \\

\begin{proof}
We first rewrite $y - \tilde y$ (First step) and accordingly reformulate the attribution $a_i$ (Second step). 

\noindent \textbf{First step}. Let us first rewrite $y - \tilde y$ by expanding {\small $y = \sigma(\sum_{i \in N} z_i + s)$} at the baseline point $\bm{\tilde z} = \bm{0}$ as follows. 
\begin{equation}
\begin{small}
\begin{aligned}
& y - \tilde y = \sigma(\sum_{i \in N^+} z_i + \sum_{i \in N^-} z_i + s) - \sigma(s) \\
& =  \sum_{k=1}^{\infty} \sum_{\tilde \kappa_1 + \tilde \kappa_2 = k} \frac{\tbinom{k}{\tilde \kappa_1, \tilde \kappa_2}}{k!}\cdot  \sigma^{k}(\bm{s}) (\sum_{i \in N^+} z_i)^{\tilde \kappa_1} (\sum_{i \in N^-} z_i)^{\tilde \kappa_2} \\
 & =  \sum_{k=1}^{\infty} \sum_{\substack{\tilde \kappa_1 + \tilde \kappa_2 = k, \\ \tilde \kappa_1 > 0}} \frac{\tbinom{k}{\tilde \kappa_1, \tilde \kappa_2}}{k!}\cdot  \sigma^{k}(\bm{s}) (\sum_{i \in N^+} z_i)^{\tilde \kappa_1} (\sum_{i \in N^-} z_i)^{\tilde \kappa_2} \\
 & + \sum_{k=1}^{\infty} \sum_{\substack{\tilde \kappa_1 = 0 \\ \tilde \kappa_2  = k}} \frac{1}{k!}\cdot  \sigma^{k}(\bm{s})  (\sum_{i \in N^-} z_i)^{k} \\
& \overset{\text{def}}{=} A_1 + A_2
\end{aligned}
\end{small}
\end{equation}
Without loss of generality, we assume that $N^+ = \{1, \dots, m\}$ and $N^- = \{m+1, \dots, n\}$ in the following.

(i) We unfold $(\sum_{i \in N^+} z_i)^{\tilde \kappa_1}$ and $(\sum_{i \in N^-} z_i)^{\tilde \kappa_2}$. Then,
the first summation term $A_1$ can be rewritten as
\begin{equation}\label{eqn:expandA1}
\begin{small}
\begin{aligned}
A_1 & = \sum_{k=1}^{\infty} \sum_{\substack{\tilde \kappa_1 + \tilde \kappa_2 = k \\ \tilde \kappa_1 > 0}} \frac{\tbinom{k}{\tilde \kappa_1, \tilde \kappa_2}}{k!} \cdot \sigma^{k}(\bm{s}) \cdot [\sum_{\substack{\kappa_1 + \dots + \kappa_m = \tilde \kappa_1>0 \\ \kappa_{m+1} + \dots + \kappa_n = \tilde \kappa_2}} \\
& \tbinom{\tilde \kappa_1}{\kappa_1, \dots, \kappa_m} \tbinom{\tilde \kappa_2}{\kappa_{m+1}, \dots, \kappa_n}
\left( z_1^{\kappa_1} \dots z_i^{\kappa_i} \dots z_m^{\kappa_m} \right) 
(z_{m+1}^{\kappa_{m+1}} \dots z_n^{\kappa_n})]  \\
& = \sum_{k=1}^{\infty} \sum_{\substack{\kappa_1 + \dots + \kappa_m = \tilde \kappa_1>0 \\ \kappa_{m+1} + \dots + \kappa_n = \tilde \kappa_2 \\ \tilde \kappa_1 + \tilde \kappa_2 = k}}
\underbrace{\frac{1}{k!} \tbinom{k}{\kappa_1, \dots, \kappa_n} \cdot \sigma^{k}(\bm{s}) \left( z_1^{\kappa_1} \dots z_n^{\kappa_n} \right)}_{I(\bm{\kappa})}  \\
 \end{aligned}
\end{small}
\end{equation}
 We find that $A_1$ actually corresponds to all Taylor expansion terms with degree vectors $\{\bm{\kappa} | \kappa_1 + \dots + \kappa_m = \tilde \kappa_1>0\}$, which means all  these Taylor expansion terms at least involve one variable in the subset $N^+$. 

(ii) We unfold $(\sum_{i \in N^-} z_i)^{\tilde \kappa_2}$, and rewrite the second summation term $A_2$ as
 \begin{equation}\label{eqn:expandA2}
\begin{small}
\begin{aligned}
A_2 & = \sum_{k=1}^{\infty} \sum_{\substack{\tilde \kappa_1 = 0, \\ \tilde \kappa_2 = k}} \frac{1}{k!} \cdot \sigma^{k}(\bm{s}) \cdot [\sum_{\substack{\kappa_{m+1} + \dots + \kappa_n = \tilde \kappa_2 = k}} \\
& \quad \tbinom{k}{\kappa_{m+1}, \dots, \kappa_n} (z_{m+1}^{\kappa_{m+1}} \dots z_n^{\kappa_n})]  \\
& = \sum_{k=1}^{\infty} \sum_{\substack{\kappa_{m+1} + \dots + \kappa_n \\ = \tilde \kappa_2 = k, \tilde \kappa_1 = 0}}
\underbrace{\frac{1}{k!} \tbinom{k}{\kappa_{m+1}, \dots, \kappa_n} \cdot \sigma^{k}(\bm{s}) \left( z_{m+1}^{\kappa_{m+1}} \dots z_n^{\kappa_n} \right)}_{I(\bm{\kappa})}
 \end{aligned}
\end{small}
\end{equation}
We find that the second summation term $A_2$ actually corresponds to all Taylor expansion terms with degree vectors 
$\{\bm{\kappa} | \kappa_1 + \dots + \kappa_m = 0\}$, which means that all these Taylor expansion terms do not involve any variables in the subset $N^+$ and only involve variables in the subset $N^-$.  \\

\noindent\textbf{Second step}.
Accordingly,  the attribution of the input variable $i \in N^+$ can be represented as 
\begin{equation}
\begin{small}
\begin{aligned}
a_i & =  \frac{\alpha z_i}{\sum_{i \in N^+} z_i} (y - \tilde y) \\
& = \sum_{k=1}^{\infty} \sum_{\substack{\tilde \kappa_1 + \tilde \kappa_2 = k, \\ \tilde \kappa_1 > 0}} \frac{\tbinom{k}{\tilde \kappa_1, \tilde \kappa_2}}{k!}\cdot  \sigma^{k}(\bm{s}) [\alpha z_i \cdot (\sum_{i \in N^+} z_i)^{\tilde \kappa_1 - 1}] (\sum_{i \in N^-} z_i)^{\tilde \kappa_2} \\
& + \frac{\alpha  z_i}{\sum\limits_{i \in N^+} z_i} \sum_{k=1}^{\infty} \sum_{\substack{\tilde \kappa_2  = k \\ \tilde \kappa_1 = 0}} \frac{1}{k!}\cdot  \sigma^{k}(\bm{s}) \cdot (\sum_{i \in N^-} z_i)^{k}\\
&  \overset{\text{def}}{=} B_1 + B_2
 \end{aligned}
\end{small}
\end{equation}
(i) By unfolding $(\sum_{i \in N^+} z_i)^{\tilde \kappa_1}$ and $(\sum_{i \in N^-} z_i)^{\tilde \kappa_2}$, we rewrite $B_1$ as follows. 
\begin{equation}\label{eqn:expandB1}
\begin{small}
\begin{aligned}
 & B_1 = \sum_{k=1}^{\infty} \sum_{\substack{\tilde \kappa_1 + \tilde \kappa_2 = k, \\ \tilde \kappa_1 > 0}} \frac{\tbinom{k}{\tilde \kappa_1, \tilde \kappa_2}}{k!}\cdot  \sigma^{k}(\bm{s})  \cdot \alpha z_i  \cdot [\sum_{\substack{\kappa_1 + \dots + \kappa_m = \tilde \kappa_1 - 1 \geq 0 \\ \kappa_{m+1} + \dots + \kappa_n = \tilde \kappa_2}}  \\
& \tbinom{\tilde \kappa_1-1}{\kappa_1, \dots, \kappa_i - 1, \dots, \kappa_m}  \tbinom{\tilde \kappa_2}{\kappa_{m+1}, \dots, \kappa_n}
\left( z_1^{\kappa_1} \dots z_i^{\kappa_i - 1} \dots z_m^{\kappa_m} \right) 
(z_{m+1}^{\kappa_{m+1}} \dots z_n^{\kappa_n})] \\
& =  \sum_{k=1}^{\infty} \sum_{\substack{\kappa_1 + \dots + \kappa_m = \tilde \kappa_1>0 \\ \kappa_{m+1} + \dots + \kappa_n = \tilde \kappa_2 \\ \tilde \kappa_1 + \tilde \kappa_2 = k, \kappa_i > 0}} \alpha \cdot \frac{\kappa_i}{\tilde \kappa_1} \cdot 
\underbrace{\frac{1}{k!} \tbinom{k}{\kappa_1, \dots, \kappa_n} \cdot \sigma^{k}(\bm{s}) \left( z_1^{\kappa_1} \dots z_n^{\kappa_n} \right)}_{I(\bm{\kappa})}  \\
 \end{aligned}
\end{small}
\end{equation}
We observe that $B_1$ actually corresponds to all Taylor expansion terms with degree vectors $\{\bm{\kappa}|\kappa_1 + \dots + \kappa_m = \tilde \kappa_1 > 0 \text{ and } \kappa_i > 0\}$, which means that all these Taylor expansion terms involve the variable $i \in N^+$. 

Then, by compare Eq. (\ref{eqn:expandA1}) and Eq. (\ref{eqn:expandB1}), we find that the \textit{LRP-$\alpha\beta$} method actually allocates $ \frac{\alpha \kappa_i}{\sum_{i \in N^+} \kappa_i}$ ratio of each Taylor expansion term $I(\bm{\kappa})$, which involves the variable $i \in N^+$, to the attribution $a_i$. 
In particular, when $\bm{\kappa} \in \Omega_i$ (\textit{i.e.}, $\kappa_i > 0$ and $\forall j \neq i, \kappa_j = 0$), we have $\frac{\kappa_i}{\sum_{i \in N^+} \kappa_i} = 1$. 
In other words, this method allocates $\alpha$ times of variable $i$'s Taylor independent effect $\phi(\bm{\kappa}) (\bm{\kappa} \in \Omega_i)$ to the attribution $a_i$. 
\begin{equation}
\begin{small}
\begin{aligned}
B_1 =   \sum_{\bm{\kappa}\in \Omega_i} \alpha \phi(\bm{\kappa}) + \! \sum_{i \in S} \! \sum_{\bm{\kappa} \in \Omega_S}  \frac{\alpha \kappa_i}{\sum\limits_{i \in N^+} \kappa_i} I(\bm{\kappa}) 
 \end{aligned}
\end{small}
\end{equation}
(ii) We rewrite $B_2$ by unfolding $(\sum_{i \in N^-} z_i)^{\tilde \kappa_2}$.
\begin{equation}\label{eqn:expandB2}
\begin{small}
\begin{aligned}
 & B_2 =  \sum_{k=1}^{\infty} \sum_{\substack{\kappa_{m+1} + \dots + \kappa_n \\ = \tilde \kappa_2 = k, \tilde \kappa_1 = 0}} \frac{\alpha  z_i}{\sum\limits_{i \in N^+} z_i} \cdot \\
&\underbrace{\frac{1}{k!} \tbinom{k}{\kappa_{m+1}, \dots, \kappa_n} \cdot \sigma^{k}(\bm{s}) \left( z_{m+1}^{\kappa_{m+1}} \dots z_n^{\kappa_n} \right)}_{I(\bm{\kappa})}
 \end{aligned}
\end{small}
\end{equation}
We observe that $B_2$ actually corresponds to all Taylor expansion terms with degree vectors $\{\bm{\kappa}|\kappa_1 + \dots + \kappa_m = \tilde \kappa_1 =  0\}$, \textit{i.e.}, all Taylor expansion terms only involving variables in $N^-$. 

By comparing Eq. (\ref{eqn:expandA2}) and Eq. (\ref{eqn:expandB2}), we find that in terms of the Taylor interaction effect $I(\bm{\kappa})$ only involving variables in $N^-$, the \textit{LRP-$\alpha\beta$} method allocates $\frac{\alpha  z_i}{\sum_{i \in N^+} z_i}$ ratio of $I(\bm{\kappa})$ to the attribution $a_i$.
\begin{equation}
\begin{small}
\begin{aligned}
 & B_2 =   \sum_{S \subseteq N^-} \sum_{\bm{\kappa} \in \Omega_S} \frac{\alpha  z_i}{\sum\limits_{i \in N^+} z_i}  I(\bm{\kappa})
 \end{aligned}
\end{small}
\end{equation}
Then, Eq. (\ref{eqn:LRPab+}) in Theorem 11 holds. 

Similarly, we can prove that Eq. (\ref{eqn:LRPab-}) in Theorem 11 also holds. \\
\end{proof}

\noindent
\textbf{Proof of Theorem 12}

\textit{Theorem 12: In the Deep Taylor method, let us analyze the layer-wise propagation of attributions. 
Then, for the variable {\small $i \in N^+ =  \{i|z_i = W_ix_i > 0\}$}, its attribution $a_i =  \frac{z_i}{\sum_{i' \in N^+} z_{i'}} (y - \tilde y)$ can be reformulated as}
\begin{equation}\label{eqn:deepTaylor}
\begin{small}
\begin{aligned}
  \!\!\!\!  a_i  & = \sum_{\bm{\kappa}\in \Omega_i} \phi(\bm{\kappa}) + \!  \sum_{\bm{\kappa} \in \Omega_S} c_i (\bm{\kappa})  + \!\!\!\!  \sum_{S \subseteq N^-}\! \sum_{\bm{\kappa}\in \Omega_S} c_2 I(\bm{\kappa})
\end{aligned}
\end{small}
\end{equation} 
\textit{where $c_i = \frac{\kappa_i}{\sum_{i' \in N^-} \kappa_{i'}}, d_i  = \frac{z_i}{\sum_{i' \in N^+} z_{i'}}$.
Moreover, for the variable {\small $i \in N^-$}, $a_i= 0$.}

\begin{proof}
Actually, Deep Taylor is a special case of LRP-$\alpha\beta$ when $\alpha = 1$ and $\beta = 0$. 
According to Theorem 11, we can obtain that Theorem 12 holds. \\
\end{proof}

\noindent
\textbf{Proof of Theorem 13}

\textit{Theorem 13: In the DeepLIFT Rescale method, let us analyze the layer-wise propagation of attributions. 
Then, the attribution of the input variable $i$, $a_i = \frac{W_i (x_i - \tilde x_i)}{\sum_i W_i (x_i - \tilde x_i)} \cdot (y - \tilde y)$ can be reformulated as}
\begin{equation}\label{eqn:deeplift rescale}
\begin{small}
\begin{aligned}
a_i & = \sum_{\bm{\kappa} \in \Omega_i} \phi(\bm{\kappa}) + \sum_{\substack{S \subseteq N, i \in S}}\sum_{\bm{\kappa} \in \Omega_S} \frac{\kappa_i}{\sum_{i \in N} \kappa_{i}}  I(\bm{\kappa}) \\
\end{aligned}
\end{small}
\end{equation}

\begin{proof}
First, we find that Eq. (\ref{eqn:deeplift rescale}) is exactly same as Eq. (\ref{eqn:IGreformulation}), \textit{i.e.}, the attribution estimated by the Integrated Gradients method. 
Therefore, to  prove Theorem 13, we only need to prove that when we analyze the layer-wise propagation of attributions, the attribution estimated by the \textit{DeepLIFT Rescale} method is equivalent to the attribution estimated by the \textit{Integrated Gradients} method. 
In the following proof, we follow notations in Remark 1. 

In terms of the $l$-th layer, let us write the output of the input point and the output of the baseline point as {\small $y= g(\bm{x}) =  \sigma(\sum_{i} W_{i} x_i +s)$} and  {\small $\tilde y = g(\bm{\tilde x}) = \sigma(\sum_i W_i \tilde x_i + s)$}. 
Then, the attribution estimated by the \textit{Integrated Gradients} method is 
\begin{equation}
\begin{small}
\begin{aligned}
&\!\!\!  a_i = (x_i - \tilde x_i) \int_0^1 \frac{\partial g(\bm{c})}{\partial c_i}|_{\bm{c} = \bm{\tilde x} + \alpha(\bm{x} - \bm{\tilde x}) }d\alpha \\
& = (x_i - \tilde x_i) \int_0^1 [\sigma'(\sum_i W_i c_i + s) \cdot W_{i}]|_{\bm{c} = \bm{\tilde x} + \alpha(\bm{x} - \bm{\tilde x}) } d\alpha \\
& = (x_i - \tilde x_i) \int_0^1 \sigma'(\sum_i W_i [\tilde x_i + \alpha(x_i - \tilde x_i)] + s)  d\alpha \\
\end{aligned}
\end{small}
\end{equation}
\begin{equation}\nonumber 
\begin{small}
\begin{aligned}
& = W_i (x_i - \tilde x_i)  \int_0^1 \sigma'(\sum_{i} \alpha (W_i x_i - W_i \tilde x_i) + W_i \tilde x_i +  s) d\alpha \\
\end{aligned}
\end{small}
\end{equation}
Let us multiply both the numerator and the denominator by {\small $\sum_i W_i (x_i - \tilde x_i)$}, then
\begin{equation} 
\begin{small}
\begin{aligned}
 a_i & = \frac{W_i (x_i - \tilde x_i)}{\sum_i W_i (x_i - \tilde x_i)} \sigma(\sum_{i} [\alpha (W_i x_i - W_i \tilde x_i] + W_i \tilde x_i +  s)|_{\alpha = 0}^{\alpha = 1} \\
& =  \frac{W_i (x_i - \tilde x_i)}{\sum_i W_i (x_i - \tilde x_i)}  \cdot (y - \tilde y)
\end{aligned}
\end{small}
\end{equation}
which is equivalent to the attribution estimated by the \textit{DeepLIFT Rescale} method. 
Combining with Theorem 7, we can obtain that Theorem 13 holds.  \\
\end{proof}

\noindent
\textbf{Proof of Theorem 14}

\textit{Theorem 14: In the Deep SHAP method, let us analyze the layer-wise propagation of attributions. 
Then, the attribution of the input variable $i$, $a_i = \frac{\phi_i(y)}{\sum_i \phi_i(y)} (y - \tilde y)$, is reformulated as}
\begin{equation}
\begin{small}
\begin{aligned}
a_i & =  \sum_{\bm{\kappa} \in \Omega_i} \phi(\bm{\kappa}) +  \sum_{\substack{S \subseteq N, i \in S}}\sum_{\bm{\kappa} \in \Omega_S}  \frac{1}{|S|} I(\bm{\kappa}) 
\end{aligned}
\end{small}
\end{equation}

\begin{proof}
Because the Shapley value $\phi_i(y)$ satisfies the efficiency axiom, we have $\sum_i \phi_i(y) = y - \tilde y$ and 
$a_i = \phi_i(y)$. 
The Deep SHAP method actually uses the Shapley value to compute attributions in each layer. 
Therefore, according to Theorem 8, we can obtain that Theorem 14 holds. \\
\end{proof}

\noindent
\textbf{Proof of Theorem 15}

\textit{Theorem 15: For the DeepLIFT RevealCancel method, let us analyze the layer-wise propagation of attributions. 
Then, for the input variable {\small $i \in N^+ = \{i| W_ix_i -W_i \tilde x_i>0\}$}, its attribution $a_i = \frac{W_i (x_i - \tilde x_i)}{\sum_{i \in N^+} W_i (x_i - \tilde x_i)} \cdot \frac{\Delta y^+}{\Delta y^+ + \Delta y^-} \cdot \Delta y$ can be reformulated as
\begin{equation}\label{eqn:DeepLIFT RevealCancel1}
\begin{small}
\begin{aligned}
a_i & =  \sum_{\bm{\kappa}\in \Omega_i} \phi(\bm{\kappa}) +  \sum_{\substack{S \subseteq N^+, i \in S}} \ \sum_{\bm{\kappa} \in \Omega_S} c_i I(\bm{\kappa}) \\
& +  \sum\nolimits_{S \cap N^+ \neq \emptyset, S \cap N^- \neq \emptyset, i \in S}\sum_{\bm{\kappa} \in \Omega_S} \frac{1}{2} c_i I(\bm{\kappa})
\end{aligned}
\end{small}
\end{equation}
where $c_i =  \frac{\kappa_i}{\sum_{i \in N^+} \kappa_i}$. Similarly, for the input variable {\small $i \in N^- = \{i| W_ix_i -W_i \tilde x_i \leq 0\}$}, its attribution $a_i = \frac{W_i (x_i - \tilde x_i)}{\sum_{i \in N^-} W_i (x_i - \tilde x_i)} \cdot \frac{\Delta y^-}{\Delta y^+ + \Delta y^-} \cdot \Delta y$ can be reformulated as
\begin{equation}\label{eqn:DeepLIFT RevealCancel2}
\begin{small}
\begin{aligned}
a_i & =  \sum_{\bm{\kappa}\in \Omega_i} \phi(\bm{\kappa}) +  \sum_{\substack{S \subseteq N^-, i \in S}} \ \sum_{\bm{\kappa} \in \Omega_S} \tilde c_i I(\bm{\kappa}) \\
& + \sum\nolimits_{S \cap N^+ \neq \emptyset, S \cap N^- \neq \emptyset, i \in S}\sum_{\bm{\kappa} \in \Omega_S} \frac{1}{2}  \tilde c_i  I(\bm{\kappa})] 
\end{aligned}
\end{small}
\end{equation}
where $\tilde c_i = \frac{\kappa_i}{\sum_{i \in N^-} \kappa_i}$.}\\

\begin{proof}
In the following proof, we follow notations in Remark 1.  First, since $\Delta y = \Delta y^+ + \Delta y^-$, then we have $a_i = \frac{W_i (x_i - \tilde x_i)}{\sum_{i \in N^+} W_i (x_i - \tilde x_i)} \cdot \Delta y^+, \forall i \in N^+$. 

Second, let us write the output of the input point and the output of the baseline point as {\small $y= g(\bm{x}) =  \sigma(\sum_{i \in N} W_{i} x_i +s)$} and  {\small $\tilde y = g(\bm{\tilde x}) = \sigma(\sum_{i \in N} W_i \tilde x_i + s)$}. 
Accordingly, $\Delta y^+$ represents the overall contribution of all input variables in $N^+$, which is defined as 
\begin{equation}
\begin{small}
\begin{aligned}
\Delta y^+ \overset{\text{def}}{=} \frac{1}{2}[g(\bm{x}_{N^+}) - g(\bm{\tilde x})] + \frac{1}{2}[g(\bm{x}) - g(\bm{x}_{N^-})] 
\end{aligned}
\end{small}
\end{equation}
According to Proposition 1, 
\begin{equation}
\begin{small}
\begin{aligned}
g(\bm{x}) &= g(\bm{\tilde x})  + \sum_{i \in N} \sum_{\bm{\kappa} \in \Omega_i} \phi(\bm{\kappa})  +  
\sum_{S \subseteq N} \sum_{\bm{\kappa} \in \Omega_S}I(\bm{\kappa}) \\
g(\bm{x}_{N^+}) & = g(\bm{\tilde x})  + \sum_{i \in N^+} \sum_{\bm{\kappa} \in \Omega_i} \phi(\bm{\kappa})  +  
\sum_{S \subseteq N^+} \sum_{\bm{\kappa} \in \Omega_S}I(\bm{\kappa}) \\
g(\bm{x}_{N^-}) & = g(\bm{\tilde x})  + \sum_{i \in N^-} \sum_{\bm{\kappa} \in \Omega_i} \phi(\bm{\kappa})  +  
\sum_{S \subseteq N^-} \sum_{\bm{\kappa} \in \Omega_S}I(\bm{\kappa})
\end{aligned}
\end{small}
\end{equation}
Therefore, we have
\begin{equation}
\begin{small}
\begin{aligned}
g(\bm{x}_{N^+}) - g(\bm{\tilde x}) = & \sum_{i \in N^+} \sum_{\bm{\kappa} \in \Omega_i} \phi(\bm{\kappa})  +  
\sum_{S \subseteq N^+} \sum_{\bm{\kappa} \in \Omega_S}I(\bm{\kappa}) \\
g(\bm{x})  - g(\bm{x}_{N^-})  = & \sum_{i \in N^+} \sum_{\bm{\kappa} \in \Omega_i} \phi(\bm{\kappa})  +  
[ \sum_{S \subseteq N^+} \sum_{\bm{\kappa} \in \Omega_S}I(\bm{\kappa}) \\
& + \sum_{\substack{S \cap N^+ \neq \emptyset, S \cap N^- \neq \emptyset}} \sum_{\bm{\kappa} \in \Omega_S}I(\bm{\kappa})] \\
\end{aligned}
\end{small}
\end{equation}
It is easy to obtain that 
\begin{equation}
\begin{small}
\begin{aligned}
\Delta y^+   \!\!  =  \!\!  \sum_{i \in N^+} \!\! \sum_{\bm{\kappa} \in \Omega_i} \phi(\bm{\kappa})   \!\!  +  
\sum_{S \subseteq N^+}   \!\!  \sum_{\bm{\kappa} \in \Omega_S}I(\bm{\kappa})  
+  \!\!  \sum_{\substack{S \cap N^+ \neq \emptyset \\ S \cap N^- \neq \emptyset}}  \! \sum_{\bm{\kappa} \in \Omega_S} \frac{1}{2} I(\bm{\kappa})
\end{aligned}
\end{small}
\end{equation}

Third, this method further uses the Rescale rule (analyzed in Theorem 13), so as to allocate the overall contribution of all input variables in $N^+$ to each input variable in $i \in N^+$, \textit{i.e.}, allocating $\frac{W_i (x_i - \tilde x_i)}{\sum_{i \in N^+} W_i (x_i - \tilde x_i)} \Delta y^+$ to each input variable.
According to Theorem 13, this method will allocate the Taylor independent effect $\phi(\bm{\kappa}) (\bm{\kappa} \in \Omega_i)$ of the variable $i$ to its attribution. 
Besides, this method will allocate $\frac{\kappa_i}{\sum_{i\in N^+} \kappa_i}$ ratio of each Taylor interaction effect $I(\bm{\kappa}) (\bm{\kappa} \in \Omega_S, i \in S)$, which involves the variable $i$, to the attribution $a_i$. 
Therefore, 
\begin{equation}
\begin{small}
\begin{aligned}
a_i  & = \sum_{\bm{\kappa}\in \Omega_i} \phi(\bm{\kappa}) +  \sum_{\substack{S \subseteq N^+, i \in S}} \ \sum_{\bm{\kappa} \in \Omega_S}  \frac{\kappa_i}{\sum_{i \in N^+} \kappa_i} I(\bm{\kappa}) \\
& +  \sum\nolimits_{S \cap N^+ \neq \emptyset, S \cap N^- \neq \emptyset, i \in S}\sum_{\bm{\kappa} \in \Omega_S} \frac{1}{2} \frac{\kappa_i}{\sum_{i \in N^+} \kappa_i} I(\bm{\kappa})
\end{aligned}
\end{small}
\end{equation}
Therefore, Eq. (\ref{eqn:DeepLIFT RevealCancel1}) in Theorem 15 holds. 
Similarly, we can prove that Eq. (\ref{eqn:DeepLIFT RevealCancel2}) holds. 
\end{proof}

\end{document}